\documentclass[11pt]{article}

\usepackage[preprint]{acl}

\usepackage{times}
\usepackage{algorithm}
\usepackage{algpseudocode}
\usepackage{latexsym}
\usepackage[most]{tcolorbox}
\usepackage{makecell}
\usepackage[table]{xcolor}
\usepackage{booktabs}
\usepackage{CJK}
\usepackage[utf8]{inputenc} % allow utf-8 input
\usepackage[T1]{fontenc}    % use 8-bit T1 fonts
\usepackage{hyperref}       % hyperlinks
\usepackage{url}            % simple URL typesetting
\usepackage{booktabs}       % professional-quality tables
\usepackage{amsfonts}       % blackboard math symbols
\usepackage{nicefrac}       % compact symbols for 1/2, etc.
\usepackage{microtype}      % microtypography
\usepackage{xcolor}         % colors
\usepackage{booktabs}
\usepackage{graphicx}
\usepackage{multirow}

\usepackage[T1]{fontenc}
\usepackage[utf8]{inputenc}

\usepackage{microtype}

\usepackage{inconsolata}

\usepackage{graphicx}

\title{When Does Visual Generation Help Visual Understanding in Unified Multimodal Models?}

\author{
Yubo Zhu$^{1,3}$\thanks{~Equal contribution.}, Zhehan Kan$^{2}$\footnotemark[1], Jingyi Yang$^{4}$,
Miaolin Chen$^{1}$, Jinbo Xing$^{3}$, \\ \textbf{Kai Zhu}$^{3}$\thanks{~Corresponding author.}, \textbf{Zijian Wang}$^{1}$, \textbf{Sheng Zhong}$^{1}$, \textbf{Wei Tong}$^{1}$\footnotemark[2] \\
\\
$^{1}$ Nanjing University, $^{2}$ Tsinghua University, $^{3}$ TongYi Lab, $^{4}$ Fudan University \\
}

\begin{document}

\maketitle

\begin{abstract}
Unified multimodal models (UMMs) can perform both understanding and generation, raising a central question: can visual generation improve understanding? Existing evaluations provide mixed evidence, but confound task difficulty, reasoning paradigms, and the closed-loop interaction between generation and understanding. We introduce \textit{VGAU-Diag}, a fine-grained evaluation framework for \underline{\textbf{v}}ision \underline{\textbf{g}}eneration-\underline{\textbf{a}}ssisted \underline{\textbf{u}}nderstanding. It stratifies samples by difficulty, enables unified evaluation of multiple reasoning paradigms, and uses Oracle-Assisted Reference Protocols. Our analysis shows that generated visual aids help on easier instances but become unreliable as reasoning complexity increases. Oracle-assisted diagnosis further reveals that the main bottleneck often lies on the visual-understanding side rather than the visual-generation side, as current UMMs struggle to leverage even faithful visual aids. We also show that effective visual generation should target visual-understanding bottlenecks rather than add more reasoning steps, and identify a three-stage transition from task-irrelevant noise, to misleading plausible guidance, and finally to useful assistance. These findings would be useful to guide the development of better UMMs. The code is available at \url{https://github.com/zyb1029/VGAU-Diag}.
\end{abstract}

\section{Introduction}

\begin{figure}[htbp]
    \centering
    \includegraphics[page=1,width=0.5\textwidth]{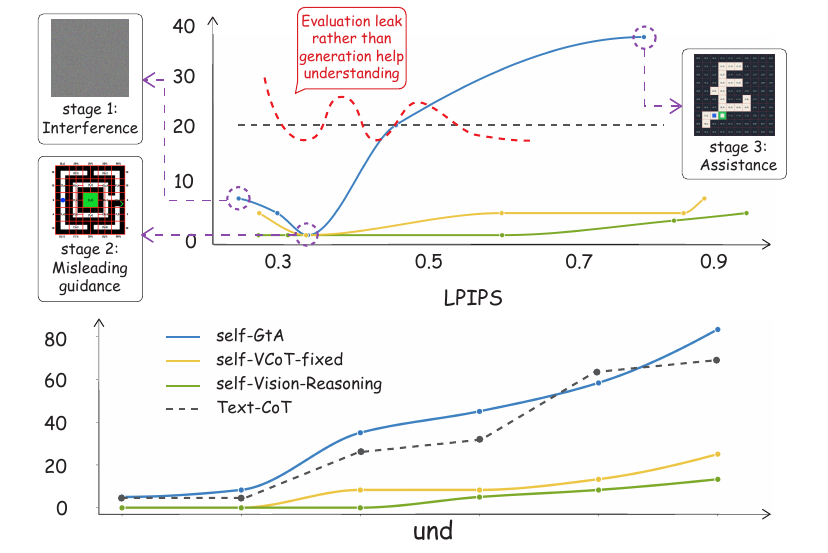}
    \vspace{-6mm}
   \caption{Solid curves show vision generation-assisted paradigms. \textbf{Top:} LPIPS serves as a proxy to distinguish interference, misleading guidance, and assistance. The dashed red trajectory denotes optimal-step prior leakage in Self-VCoT. \textbf{Bottom:} Final performance varies with visual understanding capability across different-sized Qwen-series models under the same visual aids.}
    \label{fig:intro}
    \vspace{-7mm}
\end{figure}

Recent advances in UMMs have enabled a single model to perform both visual understanding and visual generation~\cite{deng2025emergingpropertiesunifiedmultimodal, Wu_2026_CVPR, meituanlongcatteam2026longcatnextlexicalizingmodalitiesdiscrete, ai2026llada20uniunifyingmultimodalunderstanding}. This progress raises a fundamental question: can visual generation serve as an intermediate reasoning process that improves visual understanding? 
 This question has been explored through several representative vision generation-assisted understanding paradigms. In Vision-CoT (VCoT), models reason through interleaved textual and visual chains of thought. In Generate-then-Answer (GtA), models first generate auxiliary visual content and then answer through textual reasoning. In Vision-Reasoning (VR), models directly produce visual outputs to solve the problem. Recent works~\cite{Shi_2026_CVPR,zou-etal-2026-uni, wen2026unig2ubenchunifiedmodelsadvance} have tried these paradigms and suggested that visual generation can provide useful intermediate evidence for visual understanding.

However, these gains are not consistently reliable. Across different models and vision generation-assisted paradigms, the improvements brought by visual generation vary substantially. We argue that this variability arises from three key limitations in existing evaluations. \textit{First}, current evaluations are commonly formulated as multiple-choice tasks and often mix instances with different difficulty levels, making it unclear whether observed gains reflect genuine capability improvement or stochastic success on easier samples. \textit{Second}, each task is typically evaluated with only one visual reasoning paradigm, without controlled comparison against other generation-assisted paradigms. As a result, it remains unclear whether the improvement truly comes from visual generation or from paradigm-specific evaluation. \textit{Third}, existing evaluations largely treat UMM-based visual reasoning as a closed-loop black box, where visual generation and visual understanding are tightly entangled, making it difficult to diagnose whether failures mainly stem from insufficient visual understanding or unreliable visual generation.

Motivated by these limitations, we introduce a fine-grained evaluation framework for vision generation-assisted understanding. \textit{First}, rather than evaluating mixed-difficulty samples with a single aggregate score, we organize instances by difficulty, which makes it possible to examine whether visual generation helps consistently across different levels of reasoning complexity. \textit{Second}, our evaluation framework supports three representative vision generation-assisted understanding paradigms under the same setting: VCoT, GtA, and VR, enabling controlled comparison across different ways of using visual generation for reasoning. \textit{Third}, we introduce Oracle-Assisted Reference Protocols to decouple visual generation from visual understanding, enabling us to diagnose whether failures mainly stem from weak understanding or unreliable generation. To support this evaluation, we use visual planning tasks, where models must infer task-relevant spatial states from images and produce executable action sequences. This task formulation provides a verifiable and fine-grained testbed for assessing whether visual generation can support visual understanding.

Under this evaluation framework, we compare multiple vision generation-assisted reasoning paradigms and obtain several key findings as illustrated in Figure~\ref{fig:intro}. \textit{First}, difficulty-aware evaluation is crucial for revealing the effect of visual generation: when samples are stratified by difficulty, visual generation yields observable gains on easier instances, but its benefits become much less stable on harder ones. This indicates that current unified models can exploit generated visual aids when the visual state is relatively simple, yet still struggle to translate visual generation into reliable understanding gains for more complex states. \textit{Second}, visual understanding is the primary bottleneck for vision generation-assisted understanding. Although proprietary VLMs can substantially benefit from oracle visual aids, UMMs still perform poorly even when faithful aids are provided, suggesting that current UMMs often lack the visual understanding ability needed to effectively leverage such aids. \textit{Third}, vision generation-assisted understanding exhibits a three-stage transition: task-irrelevant visuals mainly act as interference, real but incorrect visual aids can provide misleading guidance, and high-fidelity task-aligned visual aids eventually offer helpful assistance. \textit{Finally}, effective visual generation should target the bottleneck of visual understanding rather than simply add more visual reasoning steps. We find a new GtA-based method offers targeted assistance by identifying decision-critical visual states and generating auxiliary visual representations for textual reasoning. By contrast, VCoT's unstable gains mainly stem from optimal-step prior leakage, after removing this prior, GtA remains more reliable, suggesting that targeted auxiliary generation is more effective than producing more intermediate visual states.

Our contributions are summarized as follows:

a) We introduce \textit{VGAU-Diag}, a fine-grained evaluation framework for vision generation-assisted understanding in UMMs, enabling evaluation across different difficulty levels, unified comparison of representative vision generation-assisted understanding paradigms, and oracle-assisted diagnosis of understanding–generation bottlenecks.

b) We argue the view that unreliable visual generation is the main bottleneck of vision generation-assisted understanding. Through oracle-assisted analysis, we show that the dominant bottleneck lies in visual understanding: current UMMs struggle to use even faithful visual aids. Based on this diagnosis, we study a bottleneck-targeted GtA method that generates task-relevant visual intermediates for reasoning. After correcting optimal-step prior leakage in previous Vision-CoT evaluations, targeted GtA proves more reliable than generic step-wise visual reasoning, showing that effective visual generation should address visual-understanding bottlenecks rather than add more visual reasoning steps.

c) We argue that the benefit of vision generation-assisted understanding does not simply increase monotonically with stronger visual generation. Instead, we reveal a three-stage transition in vision generation-assisted understanding: generated intermediates first act as task-irrelevant noise, then become misleading guidance when they are visually plausible but incorrect, and finally provide stable assistance once both visual understanding and visual generation reach a sufficient capability level.

\section{Related Work}
    
\paragraph{Visual Generation for Visual Understanding.}
Recent VLM research shows that visual generation can improve visual understanding during training through visual supervision that enhances visual perception~\cite{wang2025reconstructive, wang2025autoregressive, wei2026youtuvlunleashingvisualpotential, zhu2025llmknowsestimatingllmperceived, dai2025see, shao2025anchoring, shao2026trackaligningrewardsstates}, or during inference through external visual aids that assist grounding~\cite{cheng2025comtnovelbenchmarkchain, leng2025crosswordbench, wu2025vicbenchbenchmarkingvisualinterleavedchainofthought, fu2025refocus, li2025imagine}.

 \vspace{-2mm}
\paragraph{Unified Multimodal Models.}
    Unified multimodal models (UMMs) aim to integrate visual understanding and visual generation within a single model. Recent UMMs, including Show-o2, Janus-Pro, BAGEL, OmniGen2, LongCat-Next, LLaDA2.0-Uni, have demonstrated increasingly unified capabilities across image-text understanding and visual generation~\cite{xie2025showo, chen2025janusprounifiedmultimodalunderstanding,deng2025emergingpropertiesunifiedmultimodal, Wu_2026_CVPR, meituanlongcatteam2026longcatnextlexicalizingmodalitiesdiscrete, ai2026llada20uniunifyingmultimodalunderstanding}. This integration makes UMMs a natural setting for studying whether self-generated visual aids can improve visual understanding.
    
    \vspace{-2mm}
    \paragraph{Vision Generation-Assisted Understanding in UMMs.}
    Recent works have begun to evaluate whether vision generation can assist understanding in UMMs, such as~\cite{xie2026mmeunify, Shi_2026_CVPR, liang2026rover, zou-etal-2026-uni, zeller2026mentisoculi, wen2026unig2ubenchunifiedmodelsadvance}. MME-Unify, RealUnify, and Rover~\cite{xie2026mmeunify, Shi_2026_CVPR, liang2026rover} address this question at the unified multimodal level, Uni-MMMU~\cite{zou-etal-2026-uni} examines bidirectional understanding--generation synergy across multi-discipline reasoning tasks, MentisOculi~\cite{zeller2026mentisoculi} probes whether UMMs possess vision-reasoning capabilities, and UniG2U-Bench~\cite{wen2026unig2ubenchunifiedmodelsadvance} reports that most UMMs obtain gains under the Vision-CoT paradigm on step-level accuracy. However, existing evaluations mix task difficulties or are confined to a single evaluation paradigm, rely on weakly verifiable answers, and treat generation--understanding interactions as a closed-loop black box. In contrast, we provide a controlled and verifiable testbed for vision generation-assisted understanding, enabling difficulty-aware comparison across reasoning paradigms and fine-grained diagnosis of generation--understanding interactions, thereby revealing when visual generation helps and what limits its effectiveness in current UMMs.

\begin{figure*}[htbp]
    \centering
    \includegraphics[page=2,width=\textwidth]{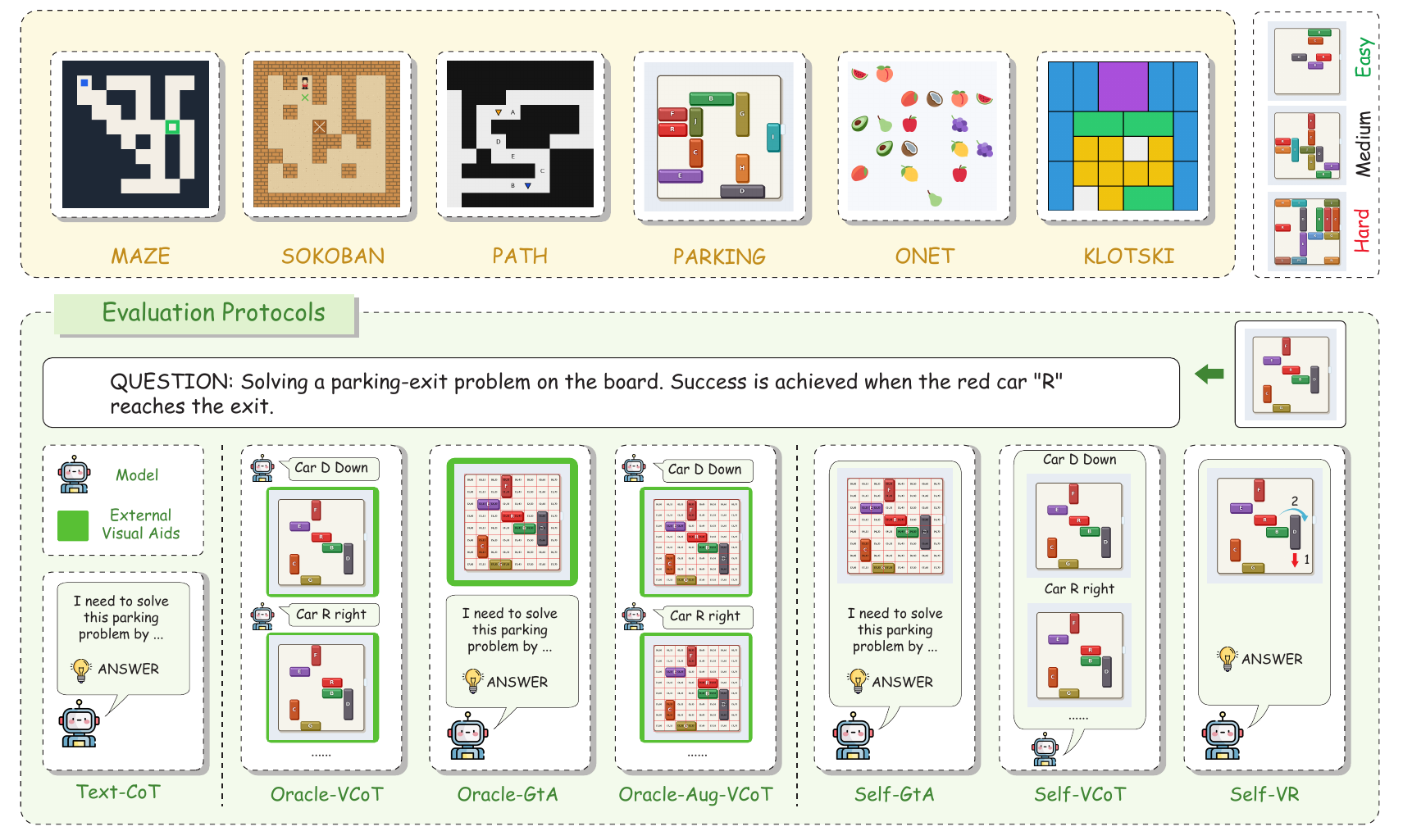}
    \caption{Overview of proposed evaluation framework.}
    \vspace{-5mm}
\end{figure*}

\section{Evaluation Framework}

We evaluate vision generation-assisted understanding in UMMs across six verifiable visual planning task families and three difficulty levels, adopt both closed-loop self-generation and oracle-assisted reference protocols, enabling controlled comparison across visual-aid paradigms and fine-grained diagnosis of generation–understanding bottlenecks.

\subsection{Task Design}

We instantiate the framework with visual planning tasks, whose image-grounded spatial reasoning and executable action outputs provide a verifiable and fine-grained testbed for generation-assisted visual understanding. The suite covers six families: Maze, Sokoban, Path, Onet, Parking, and Klotski, spanning path search, object manipulation, route tracing, pair matching, constrained motion, and sliding-block planning. \textbf{Maze:} Navigate from the start to the goal on a grid while avoiding obstacles. \textbf{Sokoban:} Push all boxes onto their target locations through agent-object interaction. \textbf{Path:} Identify the ordered letters passed along a path from the start to the end. \textbf{Onet:} Connect and eliminate all matching pairs under rule-based path constraints. \textbf{Parking:} Move vehicles in a constrained layout to guide the target vehicle to the exit. \textbf{Klotski:} Slide blocks within a confined board to move the target block to the goal position.

Many existing evaluations of vision generation-assisted understanding rely on multiple-choice questions, where UMMs perform close to random guessing, leading to unstable results. To ensure more reliable evaluation, we require models to generate complete executable solutions instead of selecting from predefined options. For each task family, we construct 100 instances and split them into easy, medium, and hard levels. Detailed settings are provided in Appendix~\ref{sec:task_details}.

\subsection{Evaluation Protocols}

Prior evaluations typically examine only a single visual-aid paradigm, treating one specific implementation as evidence for vision generation-assisted understanding as a whole. Moreover, evaluations based only on UMMs’ self-generated visuals cannot disentangle whether failures arise from poor generation fidelity, weak visual understanding, or the vision generation-assisted understanding paradigm itself. To address these limitations, we use two groups of protocols: closed-loop self-generation protocols, including Self-GtA, Self-VCoT, and Self-VR, and oracle-assisted reference protocols for both VLMs and UMMs, where externally constructed visual aids replace model-generated ones. Together, these protocols enable controlled comparison across vision generation-assisted understanding paradigms and fine-grained diagnosis of generation fidelity, visual understanding, and paradigm-level bottlenecks.

\subsubsection{Self-Generation Protocols for UMMs}

For UMMs, we evaluate three self-generation protocols: Self Generate-then-Answer, Self Vision CoT, and Self Visual Reasoning.

\paragraph{Self Generate-then-Answer (Self-GtA).}
Self-GtA follows a two-stage generation-to-understanding paradigm: the UMM first generates an auxiliary visual output, then performs textual reasoning conditioned on both the auxiliary visual and the original task. For vision planning tasks, we instantiate this paradigm as grid augmentation, where UMMs add grids and coordinates to the original visual input to clarify spatial layouts.

\paragraph{Self Vision CoT (Self-VCoT).}
Self-VCoT requires the model to solve the task through step-by-step visual interaction rather than relying on a single generated aid. Following prior vision-planning evaluations~\cite{xie2026mmeunify, zou-etal-2026-uni, wen2026unig2ubenchunifiedmodelsadvance}, the model predicts one action at each step, after which the next visual state is generated and fed back for the next prediction. This process continues until an optimal-length solution is completed, an invalid action is produced, or the optimal step budget is exceeded.

\paragraph{Self Visual Reasoning (Self-VR).}
Self-VR evaluates whether UMMs can reason directly through visual generation. Instead of using generated visuals as intermediate aids, the model produces a visual output that encodes its predicted solution. For vision planning tasks, such outputs may include a route from the start to the target, a sequence of movement directions, or selected object pairs.

\subsubsection{Oracle-Assisted Reference Protocols}

Self-generation protocols evaluate UMMs in a closed loop, but their failures entangle generation fidelity, visual understanding, and paradigm effectiveness. To disentangle these factors, we introduce oracle-assisted reference protocols, where externally provided visual aids replace model-generated outputs. For VLMs, these aids estimate the attainable performance of each paradigm under strong visual understanding, providing an upper reference under the same visual-aid design and evaluation protocol. For UMMs, oracle conditions control for generation fidelity, so the gap from self-generation reveals whether the bottleneck lies in producing the aid, using it for planning, or both. We consider four reference conditions: \textbf{Text-CoT.} The model solves the task with textual chain-of-thought reasoning only, without auxiliary visual aids. \textbf{Oracle-Vision-CoT (Oracle-VCoT).} Oracle-VCoT follows Self-VCoT but replaces model-generated step-wise visual states with externally constructed ones. \textbf{Oracle-Generate-then-Answer (Oracle-GtA).} Oracle-GtA follows Self-GtA but replaces the model-generated auxiliary visual with an externally constructed aid of the same format. \textbf{Oracle-Augmented-Vision-CoT (Oracle-Aug-VCoT).} Oracle-Aug-VCoT extends Oracle-VCoT by applying the same auxiliary-visual augmentation used in Oracle-GtA to each externally constructed step-wise visual state.

These four conditions enable controlled comparisons across vision generation-assisted paradigms and visual-aid designs. Beyond comparing text-only reasoning with generation-assisted reasoning, the oracle protocols support two targeted analyses: Oracle-VCoT vs. Oracle-Aug-VCoT measures the effect of adding task-relevant structural cues to step-wise visual states, while Oracle-GtA vs. Oracle-Aug-VCoT compares global GtA with step-wise VCoT under the same augmented visual-aid design. Together, these comparisons help disentangle the effect of reasoning paradigm from explicit visual augmentation. The details are provided in ~\ref{appendix:protocol_detail}.

\subsection{Solver-based Verification}
Previous evaluations of generation-assisted understanding in UMMs~\cite{xie2026mmeunify, zou-etal-2026-uni, wen2026unig2ubenchunifiedmodelsadvance} often rely on exact-match or step-wise accuracy metrics. Exact-match evaluation rejects valid alternative solutions, while step-wise accuracy can bias results by rewarding action overlap rather than task success. To address this, we adopt solver-based task-level evaluation and separately report optimal/feasible solution accuracy. The details are provided in ~\ref{sec:task_details}.

\section{Experiment}

\subsection{Experiment Setup}

\noindent\textbf{Models.} 
We categorize the evaluated models  into open-source VLMs, closed-source VLMs and UMMs. The details are provided in Appendix~\ref{appendix:model_detail}.

\noindent\textbf{Evaluation Details.} We set \texttt{max\_new\_tokens} to $8192$. For open-source UMMs, we run each experiment independently three times and report the average performance. 
Additional implementation details are provided in Appendix~\ref{appendix:evaluation_detail}.

\begin{table*}[!t]
\small
\centering
\resizebox{\textwidth}{!}{%
\begin{tabular}{ll*{6}{ccc}}
\toprule
\textbf{Model} & \textbf{Method}
& \multicolumn{3}{c}{\textbf{Maze}}
& \multicolumn{3}{c}{\textbf{Sokoban}}
& \multicolumn{3}{c}{\textbf{Path}}
& \multicolumn{3}{c}{\textbf{Onet}}
& \multicolumn{3}{c}{\textbf{Parking}}
& \multicolumn{3}{c}{\textbf{Klotski}} \\
\cmidrule(lr){3-5}\cmidrule(lr){6-8}\cmidrule(lr){9-11}\cmidrule(lr){12-14}\cmidrule(lr){15-17}\cmidrule(lr){18-20}
&
& \textbf{Easy} & \textbf{Medium} & \textbf{Hard}
& \textbf{Easy} & \textbf{Medium} & \textbf{Hard}
& \textbf{Easy} & \textbf{Medium} & \textbf{Hard}
& \textbf{Easy} & \textbf{Medium} & \textbf{Hard}
& \textbf{Easy} & \textbf{Medium} & \textbf{Hard}
& \textbf{Easy} & \textbf{Medium} & \textbf{Hard} \\
\midrule
\multicolumn{20}{c}{\textbf{Open-source VLMs}} \\
\midrule
\multirow{5}{*}{Qwen2.5-VL-7B-Instruct}
& Direct            & 10.0 & 0.0 & 0.0 & 5.0 & 0.0 & 0.0 & 0.0 & 0.0 & 0.0 & 0.0 & 0.0 & 0.0 & 0.0 & 0.0 & 0.0 & 0.0 & 0.0 & 0.0 \\
& Text-CoT          & 0.0 & 0.0 & 0.0 & 0.0 & 0.0 & 0.0 & 0.0 & 0.0 & 0.0 & 0.0 & 0.0 & 0.0 & \textbf{15.0} & 0.0 & 0.0 & 0.0 & 0.0 & 0.0 \\
& Oracle-VCoT        & 10.0 & 0.0 & 0.0 & 0.0 & 0.0 & 0.0 & 0.0 & 0.0 & 0.0 & 0.0 & 0.0 & 0.0 & 0.0 & 0.0 & 0.0 & 0.0 & 0.0 & 0.0 \\
\addlinespace[0.2em]
\rowcolor{gray!15}
& Oracle-GtA   & 0.0 & 0.0 & 0.0 & 0.0 & 0.0 & 0.0 & 0.0 & 0.0 & 0.0 & 0.0 & 0.0 & 0.0 & 5.0 & 0.0 & 0.0 & 0.0 & 0.0 & 0.0 \\
\rowcolor{gray!15}
& Oracle-Aug-VCoT & \textbf{15.0} & 0.0 & 0.0 & 0.0 & 0.0 & 0.0 & \textbf{15.0} & 0.0 & 0.0 & 0.0 & 0.0 & 0.0 & 0.0 & 0.0 & 0.0 & 0.0 & 0.0 & 0.0 \\
\midrule
\multirow{5}{*}{Nemotron-nano-12b-v2-vl}
& Direct            & 5.0 & 0.0 & 0.0 & 0.0 & 0.0 & 0.0 & \textbf{5.0} & \textbf{6.6} & 0.0 & 0.0 & 0.0 & 0.0 & 0.0 & 0.0 & 0.0 & 0.0 & 0.0 & 0.0 \\
& Text-CoT          & 5.0 & 0.0 & 0.0 & \textbf{10.0} & 0.0 & 0.0 & \textbf{5.0} & \textbf{6.6} & \textbf{2.0} & 0.0 & 0.0 & 0.0 & 15.0 & 0.0 & 0.0 & 0.0 & 0.0 & 0.0 \\
&  Oracle-VCoT         & 15.0 & 0.0 & 0.0 & \textbf{10.0} & 0.0 & 0.0 & \textbf{5.0} & 0.0 & 0.0 & 0.0 & 0.0 & 0.0 & 65.0 & 0.0 & 0.0 & 0.0 & 0.0 & 0.0 \\
\addlinespace[0.2em]
\rowcolor{gray!15}
& Oracle-GtA   & \textbf{35.0} & 0.0 & 0.0 & 0.0 & 0.0 & 0.0 & 0.0 & 3.3 & 0.0 & \textbf{40.0} & 0.0 & 0.0 & 0.0 & 0.0 & 0.0 & 0.0 & 0.0 & 0.0 \\
\rowcolor{gray!15}
& Oracle-Aug-VCoT & 15.0 & 0.0 & 0.0 & \textbf{10.0} & 0.0 & 0.0 & 0.0 & 0.0 & 0.0 & 30.0 & 0.0 & 0.0 & \textbf{75.0} & 0.0 & 0.0 & 0.0 & 0.0 & 0.0 \\
\midrule
\multirow{5}{*}{Qwen3-VL-8B-Instruct}
& Direct            & 0.0 & 0.0 & 0.0 & 0.0 & 0.0 & 0.0 & 5.0 & 0.0 & 0.0 & 0.0 & 0.0 & 0.0 & 15.0 & 0.0 & 0.0 & 0.0 & 0.0 & 0.0 \\
& Text-CoT          & 5.0 & 0.0 & 0.0 & 5.0 & 0.0 & 0.0 & 10.0 & \textbf{3.3} & 0.0 & 0.0 & 0.0 & 0.0 & 10.0 & 0.0 & 0.0 & 0.0 & 0.0 & 0.0 \\
&  Oracle-VCoT         & 20.0 & 0.0 & 0.0 & 5.0 & 0.0 & 0.0 & 0.0 & 0.0 & 0.0 & 0.0 & 0.0 & 0.0 & 65.0 & 0.0 & 0.0 & 0.0 & 0.0 & 0.0 \\
\addlinespace[0.2em]
\rowcolor{gray!15}
& Oracle-GtA   & \textbf{55.0} & 0.0 & 0.0 & 0.0 & 0.0 & 0.0 & \textbf{15.0} & 0.0 & \textbf{2.0} & 55.0 & \textbf{6.6} & 0.0 & 25.0 & 0.0 & 0.0 & 0.0 & 0.0 & 0.0 \\
\rowcolor{gray!15}
& Oracle-Aug-VCoT & 45.0 & 0.0 & 0.0 & \textbf{15.0} & 0.0 & 0.0 & 0.0 & 0.0 & 0.0 & \textbf{75.0} & 3.3 & 0.0 & \textbf{70.0} & 0.0 & 0.0 & 0.0 & 0.0 & 0.0 \\
\midrule
\multirow{5}{*}{Qwen3.5-9B}
& Direct            & 0.0 & 0.0 & 0.0 & 0.0 & 0.0 & 0.0 & 25.0 & \textbf{13.3} & 0.0 & 5.0 & 0.0 & 0.0 & 15.0 & 0.0 & 0.0 & 0.0 & 0.0 & 0.0 \\
& Text-CoT          & 25.0 & 0.0 & 0.0 & 30.0 & 0.0 & 0.0 & 20.0 & 0.0 & 0.0 & 25.0 & 3.3 & 0.0 & 25.0 & 0.0 & 0.0 & 0.0 & 0.0 & 0.0 \\
&  Oracle-VCoT         & 25.0 & 0.0 & 0.0 & 20.0 & 0.0 & 0.0 & 25.0 & 0.0 & 0.0 & 0.0 & 0.0 & 0.0 & 60.0 & 0.0 & 0.0 & 0.0 & 0.0 & 0.0 \\
\addlinespace[0.2em]
\rowcolor{gray!15}
& Oracle-GtA  & \textbf{90.0} & \textbf{33.3} & \textbf{6.0} & \textbf{55.0} & \textbf{10.0} & \textbf{6.6} & \textbf{60.0} & \textbf{13.3} & \textbf{2.0} & \textbf{95.0} & \textbf{43.3} & 0.0 & \textbf{80.0} & \textbf{6.6} & 0.0 & \textbf{50.0} & 0.0 & 0.0 \\
\rowcolor{gray!15}
& Oracle-Aug-VCoT & 50.0 & 0.0 & 0.0 & 20.0 & 0.0 & 0.0 & 15.0 & 0.0 & 0.0 & 85.0 & 16.6 & 0.0 & 70.0 & 0.0 & 0.0 & 0.0 & 0.0 & 0.0 \\
\midrule
\multicolumn{20}{c}{\textbf{Closed-source VLMs}} \\
\midrule
\multirow{4}{*}{\shortstack[l]{Qwen3.6-Plus\\(reasoning mode)}}
& Text-CoT          & 70.0 & 3.3 & 0.0 & 30.0 & 3.3 & 2.0 & 60.0 & 23.3 & 4.0 & 75.0 & 70.0 & 46.0 & 70.0 & 16.6 & 0.0 & 35.0 & 0.0 & 0.0 \\
&  Oracle-VCoT         & 50.0 & 3.3 & 0.0 & 50.0 & 6.6 & 0.0 & 60.0 & 0.0 & 0.0 & 80.0 & 40.0 & 18.0 & 75.0 & 23.3 & 0.0 & 35.0 & 6.6 & 0.0 \\
\addlinespace[0.2em]
\rowcolor{gray!15}
& Oracle-GtA   & \textbf{100.0} & \textbf{100.0} & \textbf{98.0} & \textbf{100.0} & \textbf{86.6} & \textbf{58.0} & \textbf{95.0} & \textbf{63.3} & \textbf{10.0} & \textbf{100.0} & 93.3 & \textbf{58.0} & \textbf{100.0} & \textbf{90.0} & 18.0 & \textbf{70.0} & 3.3 & 0.0 \\
\rowcolor{gray!15}
& Oracle-Aug-VCoT & \textbf{100.0} & \textbf{100.0} & 92.0 & \textbf{100.0} & 26.6 & 12.0 & \textbf{95.0} & 56.6 & 2.0 & \textbf{100.0} & \textbf{96.6} & 26.0 & 90.0 & 16.6 & 0.0 & 55.0 & \textbf{10.0} & 0.0 \\
\midrule
\multirow{4}{*}{\shortstack[l]{Doubao-Seed-2.0-pro\\(reasoning mode)}}
& Text-CoT          & 80.0 & 43.3 & 28.0 & 55.0 & 10.0 & 4.0 & \textbf{100.0} & 66.6 & 30.0 & 20.0 & 30.0 & 16.0 & 35.0 & 10.0 & 0.0 & 45.0 & 10.0 & 0.0 \\
&  Oracle-VCoT         & \textbf{100.0} & 63.3 & 20.0 & 80.0 & 23.3 & 0.0 & 85.0 & 0.0 & 0.0 & 20.0 & 23.3 & 4.0 & 95.0 & 23.3 & 0.0 & 10.0 & 0.0 & 0.0 \\
\addlinespace[0.2em]
\rowcolor{gray!15}
& Oracle-GtA   & \textbf{100.0} & \textbf{93.3} & \textbf{92.0} & \textbf{100.0} & \textbf{60.0} & \textbf{28.0} & \textbf{100.0} & \textbf{86.6} & \textbf{68.0} & \textbf{100.0} & 80.0 & 30.0 & \textbf{100.0} & \textbf{80.0} & \textbf{18.0} & \textbf{90.0} & \textbf{36.6} & 0.0 \\
\rowcolor{gray!15}
& Oracle-Aug-VCoT & \textbf{100.0} & 86.6 & 26.0 & \textbf{100.0} & 33.3 & 8.0 & \textbf{100.0} & 76.6 & 38.0 & \textbf{100.0} & \textbf{100.0} & \textbf{36.0} & 95.0 & 23.3 & 0.0 & 70.0 & 10.0 & \textbf{2.0} \\
\midrule
\multirow{4}{*}{\makecell[l]{Gemini-3.1-Pro-Preview\\(reasoning mode)}}
& Text-CoT          & 75.0 & 20.0 & 24.0 & 60.0 & 13.3 & 0.0 & 60.0 & 36.6 & 8.0 & 90.0 & 0.0 & 0.0 & 30.0 & 6.6 & 0.0 & 55.0 & 3.3 & 0.0 \\
&  Oracle-VCoT         & 90.0 & 23.3 & 0.0 & 90.0 & 13.3 & 0.0 & \textbf{100.0} & 43.3 & 2.0 & 95.0 & 70.0 & 8.0 & 65.0 & 3.3 & 0.0 & 80.0 & \textbf{13.3} & 0.0 \\
\addlinespace[0.2em]
\rowcolor{gray!15}
& Oracle-GtA & \textbf{100.0} & \textbf{93.3} & \textbf{82.0} & 85.0 & \textbf{60.0} & \textbf{32.0} & 85.0 & \textbf{96.6} & \textbf{38.0} & \textbf{100.0} & 10.0 & 0.0 & \textbf{85.0} & \textbf{26.6} & 0.0 & 90.0 & \textbf{13.3} & 0.0 \\
\rowcolor{gray!15}
& Oracle-Aug-VCoT & \textbf{100.0} & 60.0 & 22.0 & \textbf{95.0} & 23.3 & 4.0 & 80.0 & 36.6 & 0.0 & \textbf{100.0} & \textbf{80.0} & 4.0 & 80.0 & 6.6 & 0.0 & \textbf{95.0} & \textbf{13.3} & \textbf{2.0} \\
\midrule
\multirow{4}{*}{\shortstack[l]{GPT-5.4\\(reasoning mode)}}
& Text-CoT          & \textbf{100.0} & 83.3 & \textbf{74.0} & 40.0 & 3.3 & 2.0 & 80.0 & 46.6 & 16.0 & \textbf{100.0} & 83.3 & 32.0 & 85.0 & \textbf{90.0} & 20.0 & \textbf{100.0} & 46.6 & \textbf{2.0} \\
&  Oracle-VCoT     & \textbf{100.0} & 26.6 & 0.0 & 55.0 & 3.3 & 0.0 & 85.0 & 6.6 & 0.0 & 95.0 & 76.6 & 8.0 & \textbf{95.0} & 36.6 & 0.0 & 75.0 & 6.6 & 0.0 \\
\addlinespace[0.2em]
\rowcolor{gray!15}
& Oracle-GtA   & \textbf{100.0} & 76.6 & 72.0 & 90.0 & \textbf{60.0} & \textbf{58.0} & 90.0 & \textbf{70.0} & \textbf{42.0} & \textbf{100.0} & \textbf{90.0} & \textbf{36.0} & \textbf{95.0} & 76.6 & \textbf{26.0} & 95.0 & \textbf{60.0} & \textbf{2.0} \\
\rowcolor{gray!15}
& Oracle-Aug-VCoT & \textbf{100.0} & \textbf{96.6} & 50.0 & \textbf{95.0} & 50.0 & 4.0 & \textbf{100.0} & 66.6 & 14.0 & \textbf{100.0} & 83.3 & 4.0 & \textbf{95.0} & 36.6 & 0.0 & 70.0 & 6.6 & 0.0 \\
\midrule
\multicolumn{20}{c}{\textbf{Unified Models}} \\
\midrule
\multirow{5}{*}{MammothModa2}
& Direct            & 0.0 & 0.0 & 0.0 & 0.0 & 0.0 & 0.0 & 0.0 & 0.0 & 0.0 & 0.0 & 0.0 & 0.0 & 0.0 & 0.0 & 0.0 & 0.0 & 0.0 & 0.0 \\
& Text-CoT          & 0.0 & 0.0 & 0.0 & 0.0 & 0.0 & 0.0 & 0.0 & 0.0 & 0.0 & 0.0 & 0.0 & 0.0 & 0.0 & 0.0 & 0.0 & 0.0 & 0.0 & 0.0 \\
& Oracle-VCoT    & 0.0 & 0.0 & 0.0 & 0.0 & 0.0 & 0.0 & 0.0 & 0.0 & 0.0 & 0.0 & 0.0 & 0.0 & 0.0 & 0.0 & 0.0 & 0.0 & 0.0 & 0.0 \\
\addlinespace[0.2em]
\rowcolor{gray!15}
& Oracle-GtA  & 0.0 & 0.0 & 0.0 & 0.0 & 0.0 & 0.0 & 0.0 & 0.0 & 0.0 & \textbf{15.0} & 0.0 & 0.0 & 0.0 & 0.0 & 0.0 & 0.0 & 0.0 & 0.0 \\
\rowcolor{gray!15}
& Oracle-Aug-VCoT & 0.0 & 0.0 & 0.0 & 0.0 & 0.0 & 0.0 & 0.0 & 0.0 & 0.0 & \textbf{5.0} & 0.0 & 0.0 & 0.0 & 0.0 & 0.0 & 0.0 & 0.0 & 0.0 \\
\midrule
\multirow{5}{*}{STAR-7B}
& Direct            & 0.0 & 0.0 & 0.0 & 0.0 & 0.0 & 0.0 & 0.0 & 0.0 & 0.0 & 0.0 & 0.0 & 0.0 & 0.0 & 0.0 & 0.0 & 0.0 & 0.0 & 0.0 \\
& Text-CoT          & 0.0 & 0.0 & 0.0 & 0.0 & 0.0 & 0.0 & 0.0 & 0.0 & 0.0 & 0.0 & 0.0 & 0.0 & 5.0 & 0.0 & 0.0 & 0.0 & 0.0 & 0.0 \\
&  Oracle-VCoT     & \textbf{10.0} & 0.0 & 0.0 & 10.0 & 0.0 & 0.0 & 0.0 & 0.0 & 0.0 & 0.0 & 0.0 & 0.0 & 5.0 & 0.0 & 0.0 & 0.0 & 0.0 & 0.0 \\
\addlinespace[0.2em]
\rowcolor{gray!15}
& Oracle-GtA& \textbf{10.0} & 0.0 & 0.0 & 0.0 & 0.0 & 0.0 & \textbf{5.0} & 0.0 & 0.0 & 0.0 & 0.0 & 0.0 & \textbf{10.0} & 0.0 & 0.0 & 0.0 & 0.0 & 0.0 \\
\rowcolor{gray!15}
& Oracle-Aug-VCoT & \textbf{10.0} & 0.0 & 0.0 & \textbf{15.0} & 0.0 & 0.0 & 0.0 & 0.0 & 0.0 & \textbf{5.0} & 0.0 & 0.0 & \textbf{10.0} & 0.0 & 0.0 & 0.0 & 0.0 & 0.0 \\
\midrule
\multirow{5}{*}{OmniGen2}
& Direct            & 0.0 & 0.0 & 0.0 & 0.0 & 0.0 & 0.0 & 0.0 & 0.0 & 0.0 & 0.0 & 0.0 & 0.0 & 0.0 & 0.0 & 0.0 & 0.0 & 0.0 & 0.0 \\
& Text-CoT          & 0.0 & 0.0 & 0.0 & 0.0 & 0.0 & 0.0 & 0.0 & 0.0 & 0.0 & 0.0 & 0.0 & 0.0 & 0.0 & 0.0 & 0.0 & 0.0 & 0.0 & 0.0 \\
&  Oracle-VCoT     & 0.0 & 0.0 & 0.0 & 5.0 & 0.0 & 0.0 & \textbf{5.0} & 0.0 & 0.0 & 0.0 & 0.0 & 0.0 & 0.0 & 0.0 & 0.0 & 0.0 & 0.0 & 0.0 \\
\addlinespace[0.2em]
\rowcolor{gray!15}
& Oracle-GtA & 0.0 & 0.0 & 0.0 & 0.0 & 0.0 & 0.0 & 0.0 & 0.0 & 0.0 & 0.0 & 0.0 & 0.0 & 0.0 & 0.0 & 0.0 & 0.0 & 0.0 & 0.0 \\
\rowcolor{gray!15}
& Oracle-Aug-VCoT & 0.0 & 0.0 & 0.0 & \textbf{10.0} & 0.0 & 0.0 & \textbf{5.0} & 0.0 & 0.0 & 0.0 & 0.0 & 0.0 & 0.0 & 0.0 & 0.0 & 0.0 & 0.0 & 0.0 \\
\midrule
\multirow{5}{*}{Janus-Pro}
& Direct            & 0.0 & 0.0 & 0.0 & 0.0 & 0.0 & 0.0 & 0.0 & 0.0 & 0.0 & 0.0 & 0.0 & 0.0 & 5.0 & 0.0 & 0.0 & 0.0 & 0.0 & 0.0 \\
& Text-CoT          & 0.0 & 0.0 & 0.0 & 0.0 & 0.0 & 0.0 & 0.0 & 0.0 & 0.0 & 0.0 & 0.0 & 0.0 & 15.0 & 0.0 & 0.0 & 0.0 & 0.0 & 0.0 \\
& Oracle-VCoT      & 0.0 & 0.0 & 0.0 & \textbf{5.0} & 0.0 & 0.0 & \textbf{5.0} & 0.0 & 0.0 & 0.0 & 0.0 & 0.0 & 0.0 & 0.0 & 0.0 & 0.0 & 0.0 & 0.0 \\
\addlinespace[0.2em]
\rowcolor{gray!15}
& Oracle-GtA  & \textbf{5.0} & 0.0 & 0.0 & \textbf{5.0} & 0.0 & 0.0 & 0.0 & 0.0 & 0.0 & 0.0 & 0.0 & 0.0 & \textbf{25.0} & 0.0 & 0.0 & 0.0 & 0.0 & 0.0 \\
\rowcolor{gray!15}
& Oracle-Aug-VCoT & 0.0 & 0.0 & 0.0 & \textbf{5.0} & 0.0 & 0.0 & \textbf{5.0} & 0.0 & 0.0 & 0.0 & 0.0 & 0.0 & 0.0 & 0.0 & 0.0 & 0.0 & 0.0 & 0.0 \\
\midrule
\multirow{5}{*}{LLaDA2.0-Uni}
& Direct            & 0.0 & 0.0 & 0.0 & 0.0 & 0.0 & 0.0 & 0.0 & 0.0 & 0.0 & 0.0 & 0.0 & 0.0 & 5.0 & 0.0 & 0.0 & 0.0 & 0.0 & 0.0 \\
& Text-CoT          & 5.0 & 0.0 & 0.0 & 0.0 & 0.0 & 0.0 & 0.0 & 0.0 & 0.0 & 0.0 & 0.0 & 0.0 & 5.0 & 0.0 & 0.0 & 0.0 & 0.0 & 0.0 \\
& Oracle-VCoT    & \textbf{10.0} & 0.0 & 0.0 & \textbf{10.0} & 0.0 & 0.0 &\textbf{5.0}  & 0.0 & 0.0 & 0.0 & 0.0 & 0.0 & 35.0 & 0.0 & 0.0 & 0.0 & 0.0 & 0.0 \\
\addlinespace[0.2em]
\rowcolor{gray!15}
& Oracle-GtA & 0.0 & 0.0 & 0.0 & 5.0 & 0.0 & 0.0 & 0.0 & 0.0 & 0.0 & 0.0 & 0.0 & 0.0 & 0.0 & 0.0 & 0.0 & 0.0 & 0.0 & 0.0 \\
\rowcolor{gray!15}
& Oracle-Aug-VCoT & 5.0 & 0.0 & 0.0 & \textbf{10.0} & 0.0 & 0.0 & \textbf{5.0} & 0.0 & 0.0 & 0.0 & 0.0 & 0.0 & \textbf{60.0} & 0.0 & 0.0 & 0.0 & 0.0 & 0.0 \\
\midrule
\multirow{5}{*}{Show-o2}
& Direct            & 0.0 & 0.0 & 0.0 & 0.0 & 0.0 & 0.0 & \textbf{5.0} & 0.0 & 0.0 & 0.0 & 0.0 & 0.0 & 0.0 & 0.0 & 0.0 & 0.0 & 0.0 & 0.0 \\
& Text-CoT          & 5.0 & 0.0 & 0.0 & 0.0 & 0.0 & 0.0 & 0.0 & 0.0 & 0.0 & 0.0 & 0.0 & 0.0 & 0.0 & 0.0 & 0.0 & 0.0 & 0.0 & 0.0 \\
& Oracle-VCoT    & \textbf{15.0} & 0.0 & 0.0 & \textbf{10.0} & 0.0 & 0.0 & \textbf{5.0} & 0.0 & 0.0 & 0.0 & 0.0 & 0.0 & \textbf{70.0} & 0.0 & 0.0 & 0.0 & 0.0 & 0.0 \\
\addlinespace[0.2em]
\rowcolor{gray!15}
& Oracle-GtA   & 11.6 & 0.0 & 0.0 & 0.0 & 0.0 & 0.0 & \textbf{5.0} & 0.0 & 0.0 & 0.0 & 0.0 & 0.0 & 10.0 & 0.0 & 0.0 & 0.0 & 0.0 & 0.0 \\
\rowcolor{gray!15}
& Oracle-Aug-VCoT & 10.0 & 0.0 & 0.0 & \textbf{10.0} & 0.0 & 0.0 & 1.6 & 0.0 & 0.0 & 0.0 & 0.0 & 0.0 & 0.0 & 0.0 & 0.0 & 0.0 & 0.0 & 0.0 \\
\midrule
\multirow{5}{*}{BAGEL}
& Direct            & 0.0 & 0.0 & 0.0 & 0.0 & 0.0 & 0.0 & 15.0 & 0.0 & 0.0 & 0.0 & 0.0 & 0.0 & 0.0 & 0.0 & 0.0 & 0.0 & 0.0 & 0.0 \\
& Text-CoT          & 5.0 & 0.0 & 0.0 & 0.0 & 0.0 & 0.0 & 5.0 & 0.0 & 0.0 & 0.0 & 0.0 & 0.0 & 5.0 & 0.0 & 0.0 & 0.0 & 0.0 & 0.0 \\
&  Oracle-VCoT       & 10.0 & 0.0 & 0.0 & 20.0 & 0.0 & 0.0 & 5.0 & 0.0 & 0.0 & 0.0 & 0.0 & 0.0 & 0.0 & 0.0 & 0.0 & 0.0 & 0.0 & 0.0 \\
\addlinespace[0.2em]
\rowcolor{gray!15}
& Oracle-GtA  & \textbf{15.0} & 0.0 & 0.0 & 0.0 & 0.0 & 0.0 & \textbf{20.0} & 0.0 & 0.0 & 40.0 & 0.0 & 0.0 & 0.0 & 0.0 & 0.0 & 0.0 & 0.0 & 0.0 \\
\rowcolor{gray!15}
& Oracle-Aug-VCoT & 0.0 & 0.0 & 0.0 & \textbf{40.0} & 0.0 & 0.0 & 5.0 & 0.0 & 0.0 & \textbf{56.6} & 0.0 & 0.0 & \textbf{30.0} & 0.0 & 0.0 & 0.0 & 0.0 & 0.0 \\
\midrule
\multirow{5}{*}{LongCat-Next}
& Direct            & 0.0 & 0.0 & 0.0 & 0.0 & 0.0 & 0.0 & 5.0 & 0.0 & 0.0 & 0.0 & 0.0 & 0.0 & 0.0 & 0.0 & 0.0 & 0.0 & 0.0 & 0.0 \\
& Text-CoT          & 10.0 & 0.0 & 0.0 & 0.0 & 0.0 & 0.0 & 10.0 & 0.0 & 0.0 & 0.0 & 0.0 & 0.0 & 15.0 & 0.0 & 0.0 & 0.0 & 0.0 & 0.0 \\
&  Oracle-VCoT    & 13.3 & 0.0 & 0.0 & 6.6 & 0.0 & 0.0 & 5.0 & 0.0 & 0.0 & 0.0 & 0.0 & 0.0 & 10.0 & 0.0 & 0.0 & 0.0 & 0.0 & 0.0 \\
\addlinespace[0.2em]
\rowcolor{gray!15}
& Oracle-GtA  & \textbf{60.0} & \textbf{13.3} & 0.0 & \textbf{15.0} & 0.0 & 0.0 & \textbf{30.0} & 0.0 & 0.0 & \textbf{60.0} & \textbf{26.6} & 0.0 & 10.0 & 0.0 & 0.0 & 0.0 & 0.0 & 0.0 \\
\rowcolor{gray!15}
& Oracle-Aug-VCoT & 5.0 & 0.0 & 0.0 & 8.3 & 0.0 & 0.0 & 15.0 & 0.0 & 0.0 & \textbf{6.6} & 0.0 & 0.0 & \textbf{61.6} & 0.0 & 0.0 & 0.0 & 0.0 & 0.0 \\
\midrule
\multirow{5}{*}{Nano-Banana}
& Direct            & 5.0 & 0.0 & 0.0 & 0.0 & 0.0 & 0.0 & 15.0 & 0.0 & 0.0 & 0.0 & 0.0 & 0.0 & 0.0 & 0.0 & 0.0 & 0.0 & 0.0 & 0.0 \\
& Text-CoT          & 5.0 & 0.0 & 0.0 & 15.0 & 0.0 & 0.0 & 20.0 & 0.0 & 0.0 & 0.0 & 0.0 & 0.0 & 5.0 & 0.0 & 0.0 & 5.0 & 0.0 & 0.0 \\
&  Oracle-VCoT    & 15.0 & 0.0 & 0.0 & \textbf{25.0} & 0.0 & 0.0 & 15.0 & 0.0 & 0.0 & 0.0 & 0.0 & 0.0 & 5.0 & 0.0 & 0.0 & 0.0 & 0.0 & 0.0 \\
\addlinespace[0.2em]
\rowcolor{gray!15}
& Oracle-GtA & 35.0 & 0.0 & 0.0 & \textbf{25.0} & 0.0 & 0.0 & \textbf{55.0} & 0.0 & 0.0 & \textbf{85.0} & \textbf{33.3} & 0.0 & \textbf{15.0} & 0.0 & 0.0 & \textbf{15.0} & 0.0 & 0.0 \\
\rowcolor{gray!15}
& Oracle-Aug-VCoT & \textbf{55.0} & 0.0 & 0.0 & \textbf{25.0} & 0.0 & 0.0 & 40.0 & 0.0 & 0.0 & 55.0 & 0.0 & 0.0 & 5.0 & 0.0 & 0.0 & 5.0 & 0.0 & 0.0 \\
\midrule
\multirow{5}{*}{Nano Banana 2}
& Direct            & 20.0 & 0.0 & 0.0 & 10.0 & 0.0 & 0.0 & 35.0 & 23.3 & 8.0 & 10.0 & 0.0 & 0.0 & 0.0 & 0.0 & 0.0 & 0.0 & 0.0 & 0.0 \\
& Text-CoT          & 35.0 & 0.0 & 0.0 & 20.0 & 0.0 & 0.0 & 25.0 & 10.0 & 2.0 & 35.0 & 10.0 & 0.0 & 10.0 & 3.3 & 0.0 & 10.0 & 0.0 & 0.0 \\
&  Oracle-VCoT    & 20.0 & 0.0 & 0.0 & 30.0 & 0.0 & 0.0 & 30.0 & 0.0 & 0.0 & 25.0 & 0.0 & 0.0 & 0.0 & 0.0 & 0.0 & 25.0 & 0.0 & 0.0 \\
\addlinespace[0.2em]
\rowcolor{gray!15}
&Oracle-GtA & \textbf{85.0} & \textbf{40.0} & \textbf{38.0} & \textbf{30.0} & 0.0 & 0.0 & \textbf{90.0} & \textbf{66.6} & \textbf{32.0} & \textbf{100.0} & \textbf{40.0} & \textbf{6.0} & \textbf{70.0} & \textbf{10.0} & 0.0 & \textbf{60.0} & \textbf{13.3} & 0.0 \\
\rowcolor{gray!15}
& Oracle-Aug-VCoT & 45.0 & 13.3 & 0.0 & 25.0 & 0.0 & 0.0 & 40.0 & 13.3 & 0.0 & 70.0 & 20.0 & 0.0 & 35.0 & 0.0 & 0.0 & 30.0 & 0.0 & 0.0 \\
\bottomrule
\end{tabular}%
}
\caption{Accuracy (\%) on Maze, Sokoban, Path, Onet, Parking, and Klotski under distinct evaluation paradigms.}
\label{tab:task_method_acc_updated}
\vspace{-5mm}
\end{table*}

\subsection{Main Results and Analysis}

In this section, we analyze the main results across oracle-assisted and self-generation settings. Table~\ref{tab:task_method_acc_updated} reports the performance of VLMs and UMMs under Text-CoT, Oracle-VCoT, Oracle-GtA and Oracle-Aug-VCoT across six tasks. Table~\ref{tab:unified_models_vision} presents the results of UMMs under three self-generation protocols, including Self-GtA, Self-VCoT, and Self-VR. Given the generally low performance on harder cases under oracle-assisted settings, we evaluate UMM self-generation protocols only on easy cases. Feasible-solution accuracy under the corresponding oracle-assisted and self-generation settings is reported in Table~\ref{tab:additional-reach-any} and Table~\ref{tab:unified_models_reach_any} in Appendix.

\noindent\textbf{Insight 1}:  \textit{Difficulty-aware evaluation is essential for revealing when visual generation improves visual understanding.} 

Previous evaluations of vision generation-assisted visual understanding usually mix samples of different difficulty levels into a single evaluation set, which makes the effect of visual generation difficult to observe: gains on easy cases can be masked by failures on harder ones, leading to an overly coarse conclusion about whether visual generation helps visual understanding. In contrast, as illustrated in Table~\ref{tab:task_method_acc_updated}, after dividing samples into easy, medium, and hard levels according to the optimal solution length, we can conduct a more fine-grained analysis of generation-assisted reasoning across different planning difficulties. Under this difficulty-aware setting, we find that for UMMs, visual generation can already bring observable improvements on easy instances, indicating that current unified models are able to use generated visual aids when the visual state is relatively simple and the planning horizon is short. However, as task difficulty increases, these gains become much less stable and often disappear, suggesting that current UMMs still struggle to convert visual generation into reliable understanding gains for more complex long-horizon tasks.

% \begin{table*}[!t]
% \centering
% \caption{Results across generated models, understanding model is Qwen3.5-9B.}
% \label{tab:qwen_models}
% \resizebox{\textwidth}{!}{
% \begin{tabular}{lcccccc}
% \toprule
% Method 
% & qwen-image
% & Qwen-Image-Edit
% & Qwen-Image-Edit-plus 
% & Qwen-Image-Edit-max
% & Nano Banana 2
% & Nano-Banana-pro \\
% \midrule
% Text-CoT & \multicolumn{6}{c}{25.0} \\
% \midrule
% Self-Grid-GtA & 0.0 \textcolor{blue}{(-25.0)} & 20.0 \textcolor{blue}{(-5.0)} & 30.0 \textcolor{red}{(+5.0)} & 30.0 \textcolor{red}{(+5.0)} & 40.0 \textcolor{red}{(+15.0)} & 60.0 \textcolor{red}{(+30.0)} \\
% Self-Vision-CoT & 10.0 \textcolor{blue}{(-15.0)} & 10.0 \textcolor{blue}{(-15.0)} & 10.0 \textcolor{blue}{(-15.0)} & 10.0 \textcolor{blue}{(-15.0)} & 15.0 \textcolor{blue}{(-10.0)} & 5.0 \textcolor{blue}{(-20.0)} \\
% Self-Reasoning-GtA & 0.0 \textcolor{blue}{(-25.0)} & 0.0 \textcolor{blue}{(-25.0)} & 0.0 \textcolor{blue}{(-25.0)} & 5.0 \textcolor{blue}{(-20.0)} & 0.0 \textcolor{blue}{(-25.0)} & 10.0 \textcolor{blue}{(-15.0)} \\
% \midrule
% Oracle-Grid-GtA & \multicolumn{6}{c}{95.0 \textcolor{red}{(+70.0)}} \\
% \bottomrule
% \end{tabular}
% }
% \end{table*}

\noindent\textbf{Insight 2:} \textit{Visual understanding side is the primary bottleneck for generation-assisted understanding.}

Under the oracle setting, proprietary VLMs obtain substantial gains from the same type of visual aids, indicating that these visual aids are effective when the model has sufficient visual understanding capability. In contrast, most open-source UMMs still achieve near-zero performance under Oracle-GtA, showing that their failures cannot be mainly attributed to poor visual generation. Instead, the oracle results suggest that current UMMs often lack the visual understanding ability needed to effectively leverage the provided visual aids. This conclusion is further supported by the comparison between Self-GtA and Oracle-GtA: among $48$ task-level comparisons from $8$ open-source UMMs across $6$ tasks, only $12$ cases fall below their Oracle-GtA upper reference. In other words, for most task-level comparisons, replacing self-generated visual aids with oracle visual aids brings little improvement. Overall, the primary bottleneck in current generation-assisted understanding lies in visual understanding side rather than visual generation side.

% The paradigm determines gain stability. Self-VCoT shows unstable performance: it can bring large gains in some cases, but also often degrades performance in long-horizon planning due to step-wise error accumulation and protocol-level effects. In contrast, Self-GtA exposes task-relevant spatial structure in a single global visual aid, making its gains cleaner and more stable. Self-VR remains the hardest setting because current UMMs still lack the ability to accurately generate visual reasoning that encode complete planning solutions.

\begin{table*}[!t]
\small
\centering
\resizebox{\textwidth}{!}{%
\begin{tabular}{ll*{6}{c}}
\toprule
\textbf{Model} & \textbf{Setting} & \textbf{Maze} & \textbf{Sokoban} & \textbf{Path} & \textbf{Onet} & \textbf{Parking} & \textbf{Klotski} \\
\midrule
\multirow{3}{*}{MammothModa2}
& \cellcolor{gray!15}Self-GtA     &  \cellcolor{gray!15} 0.0 \textcolor{gray}{(+0.0)} & \cellcolor{gray!15} 0.0 \textcolor{gray}{(+0.0)} &  0.0 \cellcolor{gray!15}\textcolor{gray}{(+0.0)} & 0.0 \cellcolor{gray!15}\textcolor{gray}{(+0.0)} & 0.0 \cellcolor{gray!15}\textcolor{gray}{(+0.0)} &  \cellcolor{gray!15}0.0 \textcolor{gray}{(+0.0)} \\
& Self-VCoT    & 0.0 \textcolor{gray}{(+0.0)} & 0.0 \textcolor{gray}{(+0.0)} & 0.0 \textcolor{gray}{(+0.0)} & 0.0 \textcolor{gray}{(+0.0)} & 0.0 \textcolor{gray}{(+0.0)} & 0.0 \textcolor{gray}{(+0.0)} \\
&Self-VR  & 0.0 \textcolor{gray}{(+0.0)} & 0.0 \textcolor{gray}{(+0.0)} & 0.0 \textcolor{gray}{(+0.0)} & 0.0 \textcolor{gray}{(+0.0)} & 0.0 \textcolor{gray}{(+0.0)} & 0.0 \textcolor{gray}{(+0.0)} \\
\midrule
\multirow{3}{*}{STAR-7B}
& \cellcolor{gray!15}Self-GtA     & \cellcolor{gray!15}0.0 \textcolor{gray}{(+0.0)} & \cellcolor{gray!15}0.0 \textcolor{gray}{(+0.0)} & \cellcolor{gray!15}0.0 \textcolor{gray}{(+0.0)} & \cellcolor{gray!15}0.0 \textcolor{gray}{(+0.0)} & \cellcolor{gray!15}0.0 \textcolor{blue}{(-5.0)} & \cellcolor{gray!15}0.0 \textcolor{gray}{(+0.0)} \\
& Self-VCoT     & \textbf{1.6} \textcolor{red}{(+1.6)} & 0.0 \textcolor{gray}{(+0.0)} & 0.0 \textcolor{gray}{(+0.0)} & 0.0 \textcolor{gray}{(+0.0)} & 0.0 \textcolor{blue}{(-5.0)} & 0.0 \textcolor{gray}{(+0.0)} \\
& Self-VR & 0.0 \textcolor{gray}{(+0.0)} & 0.0 \textcolor{gray}{(+0.0)} & 0.0 \textcolor{gray}{(+0.0)} &0.0 \textcolor{gray}{(+0.0)} & 0.0 \textcolor{blue}{(-5.0)} & 0.0 \textcolor{gray}{(+0.0)} \\
\midrule
\multirow{3}{*}{OmniGen2}
& \cellcolor{gray!15}Self-GtA     & \cellcolor{gray!15}0.0 \textcolor{gray}{(+0.0)} & \cellcolor{gray!15}0.0 \textcolor{gray}{(+0.0)} & \cellcolor{gray!15}0.0 \textcolor{gray}{(+0.0)} & \cellcolor{gray!15}0.0 \textcolor{gray}{(+0.0)} & \cellcolor{gray!15}0.0 \textcolor{gray}{(+0.0)} & \cellcolor{gray!15} 0.0 \textcolor{gray}{(+0.0)} \\
& Self-VCoT      & 0.0 \textcolor{gray}{(+0.0)} & \textbf{5.0} \textcolor{red}{(+5.0)} & \textbf{5.0} \textcolor{red}{(+5.0)} & 0.0 \textcolor{gray}{(+0.0)} & 0.0 \textcolor{gray}{(+0.0)} & 0.0 \textcolor{gray}{(+0.0)} \\
& Self-VR & 0.0 \textcolor{gray}{(+0.0)} & 0.0 \textcolor{gray}{(+0.0)} & 0.0 \textcolor{gray}{(+0.0)} & 0.0 \textcolor{gray}{(+0.0)} & \textbf{1.6} \textcolor{red}{(+1.6)}  & 0.0 \textcolor{gray}{(+0.0)} \\
\midrule
\multirow{3}{*}{Janus-Pro}
& \cellcolor{gray!15}Self-GtA      & \cellcolor{gray!15}\textbf{5.0} \textcolor{red}{(+5.0)} & \cellcolor{gray!15}\textbf{5.0} \textcolor{red}{(+5.0)}  & \cellcolor{gray!15}0.0 \textcolor{gray}{(+0.0)} & \cellcolor{gray!15}0.0 \textcolor{gray}{(+0.0)} & \cellcolor{gray!15}\textbf{25.0} \textcolor{red}{(+10.0)} & \cellcolor{gray!15}0.0 \textcolor{gray}{(+0.0)} \\
& Self-VCoT     & 0.0 \textcolor{gray}{(+0.0)} & \textbf{5.0} \textcolor{red}{(+5.0)}  & \textbf{5.0} \textcolor{red}{(+5.0)}  & 0.0 \textcolor{gray}{(+0.0)} & 0.0 \textcolor{blue}{(-15.0)}  & 0.0 \textcolor{gray}{(+0.0)} \\
& Self-VR & 0.0 \textcolor{gray}{(+0.0)} & 0.0 \textcolor{gray}{(+0.0)} & 0.0 \textcolor{gray}{(+0.0)} & 0.0 \textcolor{gray}{(+0.0)} & 20.0 \textcolor{red}{(+5.0)} & 0.0 \textcolor{gray}{(+0.0)} \\
\midrule
\multirow{3}{*}{LLaDA2.0-Uni}
& \cellcolor{gray!15}Self-GtA  & \cellcolor{gray!15}0.0 \textcolor{blue}{(-5.0)} & \cellcolor{gray!15}0.0 \textcolor{gray}{(+0.0)} & \cellcolor{gray!15}0.0 \textcolor{gray}{(+0.0)} & \cellcolor{gray!15}0.0 \textcolor{gray}{(+0.0)} & \cellcolor{gray!15}0.0 \textcolor{blue}{(-5.0)} & \cellcolor{gray!15}0.0 \textcolor{gray}{(+0.0)} \\
& Self-VCoT     & \textbf{10.0} \textcolor{red}{(+5.0)} & \textbf{10.0} \textcolor{red}{(+10.0)} & \textbf{5.0} \textcolor{red}{(+5.0)} & 0.0 \textcolor{gray}{(+0.0)} & \textbf{40.0} \textcolor{red}{(+35.0)} & 0.0 \textcolor{gray}{(+0.0)} \\
& Self-VR  & 0.0 \textcolor{blue}{(-5.0)} & 5.0 \textcolor{red}{(+5.0)} & 0.0 \textcolor{gray}{(+0.0)} & 0.0 \textcolor{gray}{(+0.0)} & 0.0 \textcolor{blue}{(-5.0)} & 0.0 \textcolor{gray}{(+0.0)} \\
\midrule
\multirow{3}{*}{Show-o2}
& \cellcolor{gray!15}Self-GtA & \cellcolor{gray!15} 10.0 \textcolor{red}{(+5.0)} & \cellcolor{gray!15} 0.0 \textcolor{gray}{(+0.0)} & \cellcolor{gray!15}5.0 \textcolor{red}{(+5.0)}& \cellcolor{gray!15}0.0 \textcolor{gray}{(+0.0)} & \cellcolor{gray!15}10.0 \textcolor{red}{(+10.0)} & \cellcolor{gray!15}0.0 \textcolor{gray}{(+0.0)} \\
& Self-VCoT     &  \textbf{11.6} \textcolor{red}{(+6.6)} & \textbf{10.0} \textcolor{red}{(+10.0)} &  \textbf{6.6} \textcolor{red}{(+6.6)} &  0.0 \textcolor{gray}{(+0.0)} &  \textbf{56.6} \textcolor{red}{(+56.6)} &  0.0 \textcolor{gray}{(+0.0)} \\
& Self-VR & 0.0 \textcolor{blue}{(-5.0)}& 0.0 \textcolor{gray}{(+0.0)} & 0.0 \textcolor{gray}{(+0.0)} & 0.0 \textcolor{gray}{(+0.0)} & 10.0 \textcolor{red}{(+10.0)} & 0.0 \textcolor{gray}{(+0.0)} \\
\midrule
\multirow{3}{*}{BAGEL}
& \cellcolor{gray!15}Self-GtA     & \cellcolor{gray!15}0.0 \textcolor{blue}{(-5.0)} & \cellcolor{gray!15}0.0 \textcolor{gray}{(+0.0)} & \cellcolor{gray!15}\textbf{20.0} \textcolor{red}{(+15.0)} & \cellcolor{gray!15}0.0 \textcolor{gray}{(+0.0)} & \cellcolor{gray!15}\textbf{5.0} \textcolor{gray}{(+0.0)}  & \cellcolor{gray!15}0.0 \textcolor{gray}{(+0.0)} \\
& Self-VCoT    & \textbf{6.6} \textcolor{red}{(+1.6)} & \textbf{13.3} \textcolor{red}{(+13.3)} & 5.0 \textcolor{gray}{(+0.0)} & 0.0 \textcolor{gray}{(+0.0)} & 0.0 \textcolor{blue}{(-5.0)}  & 0.0 \textcolor{gray}{(+0.0)} \\
& Self-VR & 0.0 \textcolor{blue}{(-5.0)} & 0.0 \textcolor{gray}{(+0.0)} & 5.0 \textcolor{gray}{(+0.0)} & 0.0 \textcolor{gray}{(+0.0)} & 0.0 \textcolor{blue}{(-5.0)}  & 0.0 \textcolor{gray}{(+0.0)} \\
\midrule
\multirow{3}{*}{LongCat-Next}
& \cellcolor{gray!15}Self-GtA      & \cellcolor{gray!15}\textbf{20.0} \textcolor{red}{(+10.0)} & \cellcolor{gray!15}\textbf{5.0} \textcolor{red}{(+5.0)} & \cellcolor{gray!15}\textbf{15.0} \textcolor{red}{(+5.0)} & \cellcolor{gray!15}\textbf{5.0} \textcolor{red}{(+5.0)} & \cellcolor{gray!15}\textbf{10.0} \textcolor{blue}{(-5.0)} & \cellcolor{gray!15} 0.0 \textcolor{gray}{(+0.0)} \\
& Self-VCoT   & 10.0 \textcolor{gray}{(+0.0)} & \textbf{5.0} \textcolor{red}{(+5.0)} & 1.6 \textcolor{blue}{(-8.4)} & 0.0 \textcolor{gray}{(+0.0)} & 5.0 \textcolor{blue}{(-10.0)} & 0.0 \textcolor{gray}{(+0.0)} \\
& Self-VR  & 5.0 \textcolor{blue}{(-5.0)} & 0.0 \textcolor{gray}{(+0.0)} & 0.0 \textcolor{blue}{(-10.0)} & 0.0 \textcolor{gray}{(+0.0)} & 5.0 \textcolor{blue}{(-10.0)} & 0.0 \textcolor{gray}{(+0.0)} \\
\midrule
\multirow{3}{*}{Nano-Banana}
& \cellcolor{gray!15}Self-GtA      & \cellcolor{gray!15} \textbf{15.0} \textcolor{red}{(+10.0)} & \cellcolor{gray!15} 20.0 \textcolor{red}{(+5.0)} &  \textbf{25.0} \cellcolor{gray!15}\textcolor{red}{(+5.0)} & 0.0 \cellcolor{gray!15}\textcolor{gray}{(+0.0)} & \textbf{15.0} \cellcolor{gray!15}\textcolor{red}{(+10.0)} & \textbf{10.0} \cellcolor{gray!15}\textcolor{red}{(+5.0)} \\
& Self-VCoT   & 0.0 \textcolor{blue}{(-5.0)} & \textbf{25.0} \textcolor{red}{(+10.0)} & 20.0 \textcolor{gray}{(+0.0)} & 0.0 \textcolor{gray}{(+0.0)} & 10.0 \textcolor{red}{(+5.0)} & 0.0 \textcolor{blue}{(-5.0)} \\
& Self-VR  & 5.0 \textcolor{gray}{(+0.0)} & 10.0 \textcolor{blue}{(-5.0)} & 20.0 \textcolor{gray}{(+0.0)} & 0.0 \textcolor{gray}{(+0.0)} & 10.0 \textcolor{red}{(+5.0)} & 0.0 \textcolor{blue}{(-5.0)} \\
\midrule
\multirow{3}{*}{Nano Banana 2}
& \cellcolor{gray!15}Self-GtA      & \cellcolor{gray!15}\textbf{55.0} \textcolor{red}{(+20.0)} & \cellcolor{gray!15}\textbf{25.0} \textcolor{red}{(+5.0)} & \cellcolor{gray!15}\textbf{50.0} \textcolor{red}{(+25.0)} & \cellcolor{gray!15}\textbf{70.0} \textcolor{red}{(+35.0)} & \cellcolor{gray!15}\textbf{40.0} \textcolor{red}{(+30.0)} & \cellcolor{gray!15}\textbf{30.0} \textcolor{red}{(+20.0)} \\
&Self-VCoT     & 10.0 \textcolor{blue}{(-25.0)} & 20.0 \textcolor{gray}{(+0.0)} & 30.0 \textcolor{red}{(+5.0)} & 15.0 \textcolor{blue}{(-20.0)} & 10.0 \textcolor{gray}{(+0.0)} & 5.0 \textcolor{blue}{(-5.0)} \\
& Self-VR & 15.0 \textcolor{blue}{(-20.0)} & 0.0 \textcolor{blue}{(-20.0)} & 45.0 \textcolor{red}{(+20.0)} & 25.0 \textcolor{blue}{(-10.0)} & 10.0 \textcolor{gray}{(+0.0)} & 10.0 \textcolor{gray}{(+0.0)} \\
\bottomrule
\end{tabular}%
}
\caption{Accuracy (\%) of unified models on Maze, Sokoban, Path, Onet, Parking, and Klotski on easy cases.}
\vspace{-5mm}
\label{tab:unified_models_vision}
\end{table*}

\subsection{Further Analysis of Self-Generation Paradigms for UMMs}

A remaining question is \textbf{why UMMs do not exhibit the same paradigm trend as specialist understanding models, either under oracle-assisted evaluation or under self-generation}. This discrepancy is especially pronounced in the comparison between GtA and VCoT on Maze, Sokoban, Path, and Parking, where the paradigm-dependent instability is most evident in our results. While GtA shows relatively stable benefits in stronger understanding models, VCoT exhibits much larger fluctuations in UMMs: it can bring substantial gains on some tasks, but often fails to improve over the Text-CoT or even degrades performance on others.

\textbf{To investigate the source of this instability, we construct a controlled diagnostic setup that separates visual understanding capability from the fidelity of generated visual aids.} For visual understanding, we use the Qwen model series with different parameter scales as visual understanding models, providing a proxy for progressively stronger understanding capability. For visual generation, we vary the fidelity of auxiliary visual aids along three levels. \textbf{The first level} uses task-irrelevant visuals, where intermediate images are replaced with Gaussian noise to simulate complete generation failure. \textbf{The second level} uses real but incorrect visual aids, including mismatched images from other instances of the same task and outputs from a base generation model, to simulate plausible but inaccurate generation. \textbf{The third level} uses high-fidelity task-aligned visuals produced by stronger editing and generation models, such as Nano Banana 2, to approximate reliable visual generation.

\begin{table*}[!t]
\centering 
\resizebox{\textwidth}{!}{
\begin{tabular}{lcccccc}
\toprule
Method 
& Qwen3.5-0.8B 
& Qwen3.5-2B  
& Qwen3.5-4B 
& Qwen3.5-9B 
& Qwen3.5-27B 
& Qwen3.6-plus \\
\midrule
Text-CoT & 5.0 & 5.0 & 20.0 & 25.0 & 60.0 & 70.0 \\
\midrule
\multicolumn{7}{c}{\textit{Vision: Gaussian noise}} \\
\midrule
Self-GtA & 0.0 \textcolor{blue}{(-5.0)} & 1.6 \textcolor{blue}{(-3.4)}  & 8.3 \textcolor{blue}{(-11.7)} & 8.3 \textcolor{blue}{(-16.7)}  & 20.0 \textcolor{blue}{(-40.0)}  & 60.0 \textcolor{blue}{(-10.0)} \\
Self-VCoT & \cellcolor{gray!15}\textbf{6.6} \textcolor{red}{(+1.6)} & \cellcolor{gray!15}\textbf{6.6} \textcolor{red}{(+1.6)} & \cellcolor{gray!15}\textbf{11.6} \textcolor{blue}{(-8.4)} & \cellcolor{gray!15}\textbf{13.3} \textcolor{blue}{(-11.7)} & \cellcolor{gray!15}\textbf{35.0} \textcolor{blue}{(-25.0)} & \textbf{61.6} \textcolor{blue}{(-8.4)} \\
Self-VR & 0.0 \textcolor{blue}{(-5.0)} & 0.0 \textcolor{blue}{(-5.0)} & 6.6 \textcolor{blue}{(-13.4)} & 6.6 \textcolor{blue}{(-18.4)} & 18.3 \textcolor{blue}{(-41.7)} & 53.3 \textcolor{blue}{(-16.7)} \\
\midrule
\multicolumn{7}{c}{\textit{Vision: Mismatch Image}} \\
\midrule
Self-GtA & 3.3 \textcolor{blue}{(-1.7)} & 1.6 \textcolor{blue}{(-3.4)}  & 6.6 \textcolor{blue}{(-13.4)} & 5.0 \textcolor{blue}{(-20.0)}  & 10.0 \textcolor{blue}{(-50.0)}  & 15.0 \textcolor{blue}{(-55.0)} \\
Self-VCoT & \cellcolor{gray!15}\textbf{6.6} \textcolor{red}{(+1.6)} & \cellcolor{gray!15}\textbf{6.6} \textcolor{red}{(+1.6)} & \cellcolor{gray!15}\textbf{8.3} \textcolor{blue}{(-11.7)} & \cellcolor{gray!15}\textbf{15.0} \textcolor{blue}{(-10.0)} & \cellcolor{gray!15}\textbf{30.0} \textcolor{blue}{(-30.0)} & \textbf{35.0} \cellcolor{gray!15}\textcolor{blue}{(-35.0)} \\
Self-VR & 0.0 \textcolor{blue}{(-5.0)} & 0.0 \textcolor{blue}{(-5.0)} & 1.6 \textcolor{blue}{(-18.4)} & 3.3 \textcolor{blue}{(-21.7)} & 1.6 \textcolor{blue}{(-58.4)} & 21.6 \textcolor{blue}{(-48.4)} \\
\midrule
\multicolumn{7}{c}{\textit{Vision generator:  Qwen-Image~\cite{wu2025qwenimagetechnicalreport}}} \\
\midrule
Self-GtA & 0.0 \textcolor{blue}{(-5.0)} & 0.0 \textcolor{blue}{(-5.0)} & 0.0 \textcolor{blue}{(-20.0)} & 0.0 \textcolor{blue}{(-25.0)} & 1.6 \textcolor{blue}{(-58.4)} & 38.3 \textcolor{blue}{(-31.7)} \\
Self-VCoT & \cellcolor{gray!15}\textbf{5.0 \textcolor{gray}{(+0.0)}} & \cellcolor{gray!15}\textbf{5.0 \textcolor{gray}{(+0.0)}} & \cellcolor{gray!15}\textbf{11.6 \textcolor{blue}{(-8.4)}} & \cellcolor{gray!15}\textbf{13.3 \textcolor{blue}{(-11.7)}} & \cellcolor{gray!15}\textbf{35.0 \textcolor{blue}{(-25.0)}} & \cellcolor{gray!15}\textbf{73.3 \textcolor{red}{(+3.3)}} \\
Self-VR & 0.0 \textcolor{blue}{(-5.0)} & 0.0 \textcolor{blue}{(-5.0)} & 0.0 \textcolor{blue}{(-20.0)} & 0.0 \textcolor{blue}{(-25.0)} & 1.6 \textcolor{blue}{(-58.4)} & 23.3 \textcolor{blue}{(-46.7)} \\
\midrule
\multicolumn{7}{c}{\textit{Vision generator: Qwen-Image-Edit~\cite{wu2025qwenimagetechnicalreport}}} \\
\midrule
Self-GtA & 0.0  \textcolor{blue}{(-5.0)} & 3.3 \textcolor{blue}{(-1.7)} & \textbf{18.3} \textcolor{blue}{(-1.7)} & \textbf{26.6} \textcolor{red}{(+1.6)} & \textbf{31.6} \textcolor{blue}{(-28.4)} & \textbf{58.3} \textcolor{blue}{(-11.7)} \\
Self-VCoT & \cellcolor{gray!15} \textbf{3.3} \textcolor{blue}{(-1.7)} & \cellcolor{gray!15}\textbf{5.0} \textcolor{gray}{(+0.0)} & 10.0 \textcolor{blue}{(-10.0)} & 13.3 \textcolor{blue}{(-11.7)} & 33.3 \textcolor{blue}{(-26.7)} & 33.3 \textcolor{blue}{(-36.7)} \\
Self-VR & 0.0 \textcolor{blue}{(-5.0)} & 0.0 \textcolor{blue}{(-5.0)} & 5.0 \textcolor{blue}{(-15.0)} & 5.0 \textcolor{blue}{(-20.0)} & 11.6 \textcolor{blue}{(-48.4)} & 25.0 \textcolor{blue}{(-45.0)} \\
\midrule
\multicolumn{7}{c}{\textit{Vision generator: Nano Banana 2~\cite{raisinghani2026nanobanana2}}} \\
\midrule
Self-GtA & 5.0 \textcolor{gray}{(+0.0)} & \textbf{8.3} \textcolor{red}{(+3.3)} & \textbf{35.0} \textcolor{red}{(+15.0)} & \textbf{45.0} \textcolor{red}{(+20.0)} & \textbf{58.3} \textcolor{blue}{(-1.7)} & \textbf{83.3} \textcolor{red}{(+13.3)} \\
Self-VCoT & \cellcolor{gray!15}\textbf{6.6} \textcolor{red}{(+1.6)} & 3.3 \textcolor{blue}{(-1.7)} & 11.6 \textcolor{blue}{(-8.4)} & 25.0 \textcolor{gray}{(+0.0)} & 33.3 \textcolor{blue}{(-26.7)} & \textcolor{red}{45.0} \textcolor{blue}{(-25.0)} \\
Self-VR & 1.6 \textcolor{blue}{(-3.4)} & 0.0 \textcolor{blue}{(-5.0)} & 8.3 \textcolor{blue}{(-11.7)} & 8.3 \textcolor{blue}{(-16.7)} & 13.3 \textcolor{blue}{(-46.7)} & 25.0 \textcolor{blue}{(-45.0)} \\
\midrule
\multicolumn{7}{c}{\textit{Oracle generation}} \\
\midrule
Oracle-GtA & 0.0 \textcolor{blue}{(-5.0)} & \textbf{40.0} \textcolor{red}{(+35.0)} & \textbf{85.0} \textcolor{red}{(+65.0)} & \textbf{90.0} \textcolor{red}{(+65.0)} & \textbf{100.0} \textcolor{red}{(+40.0)} & \textbf{100.0} \textcolor{red}{(+30.0)} \\
Oracle-VCoT & \cellcolor{gray!15}\textbf{5.0} \textcolor{gray}{(+0.0)}& 10.0 \textcolor{red}{(+5.0)}& 20.0 \textcolor{gray}{(+0.0)} & 25.0 \textcolor{gray}{(+0.0)} & 45.0 \textcolor{blue}{(-15.0)} & 50.0 \textcolor{blue}{(-20.0)} \\
\bottomrule
\end{tabular}
}
\vspace{-2mm}
\caption{Diagnosis of visual understanding scale and generation fidelity across Qwen models on easy cases.}
\vspace{-3mm}
\label{tab:qwen_reasoners_maze}
\end{table*}

Based on Table~\ref{tab:qwen_reasoners_maze}, we observe two key patterns. First, when generation quality is fixed, performance generally improves as visual understanding capability becomes stronger. Second, when visual understanding capability is fixed, improving generation fidelity does not lead to monotonic gains. Instead, performance often drops when moving from task-irrelevant noise to real but incorrect visual aids, and improves substantially only when high-fidelity task-aligned visual aids are provided.

\textbf{Insight 3:} \textit{Generation-assisted understanding follows a three-stage transition from interference, to misleading guidance, and finally to assistance.}

In the \textbf{interference stage}, task-irrelevant visuals such as noise contain no task-useful information and act as distractions. 
In the \textbf{misleading stage}, incorrect visual aids become more harmful: although they appear plausible, they encode an incorrect task state, causing the model to rely on false visual evidence. 
In the \textbf{assistance stage}, high-fidelity task-aligned visual aids provide reliable task-relevant information, reduce the burden of visual understanding, and support stable performance gains.

\begin{table*}[!t]
\small
\centering
\resizebox{\textwidth}{!}{%
\begin{tabular}{ll*{6}{c}}
\toprule
\textbf{Model} & \textbf{Setting} & \textbf{Maze} & \textbf{Sokoban} & \textbf{Path} & \textbf{Onet} & \textbf{Parking} & \textbf{Klotski} \\
\midrule
\multirow{2}{*}{MammothModa2}
& Self-VCoT & 0.0 \textcolor{gray}{(+0.0)} & 0.0 \textcolor{gray}{(+0.0)} & 0.0 \textcolor{gray}{(+0.0)} & 0.0 \textcolor{gray}{(+0.0)} & 0.0 \textcolor{gray}{(+0.0)} & 0.0 \textcolor{gray}{(+0.0)} \\
& Self-VCoT-fixed &0.0 \textcolor{gray}{(+0.0)} & 0.0 \textcolor{gray}{(+0.0)}& 0.0 \textcolor{gray}{(+0.0)}& 0.0 \textcolor{gray}{(+0.0)}& 0.0 \textcolor{gray}{(+0.0)}& 0.0 \textcolor{gray}{(+0.0)}\\
\midrule

\multirow{2}{*}{STAR-7B}
& Self-VCoT & 1.6 \textcolor{red}{(+1.6)} & 0.0 \textcolor{gray}{(+0.0)} & 0.0 \textcolor{gray}{(+0.0)} & 0.0 \textcolor{gray}{(+0.0)} & 0.0 \textcolor{blue}{(-5.0)} & 0.0 \textcolor{gray}{(+0.0)} \\
& Self-VCoT-fixed & 0.0 \textcolor{gray}{(+0.0)}& 0.0 \textcolor{gray}{(+0.0)}&0.0 \textcolor{gray}{(+0.0)} & 0.0 \textcolor{gray}{(+0.0)}&0.0 \textcolor{blue}{(-5.0)} & 0.0 \textcolor{gray}{(+0.0)}\\
\midrule

\multirow{2}{*}{OmniGen2}
& Self-VCoT & 0.0 \textcolor{gray}{(+0.0)} & 5.0 \textcolor{red}{(+5.0)} & 5.0 \textcolor{red}{(+5.0)} & 0.0 \textcolor{gray}{(+0.0)} & 0.0 \textcolor{gray}{(+0.0)} & 0.0 \textcolor{gray}{(+0.0)} \\
& Self-VCoT-fixed & 0.0 \textcolor{gray}{(+0.0)}&0.0 \textcolor{gray}{(+0.0)} & 0.0 \textcolor{gray}{(+0.0)}& 0.0 \textcolor{gray}{(+0.0)}&0.0 \textcolor{gray}{(+0.0)} &0.0 \textcolor{gray}{(+0.0)} \\
\midrule

\multirow{2}{*}{Janus-Pro}
& Self-VCoT & 0.0 \textcolor{gray}{(+0.0)} & 5.0 \textcolor{red}{(+5.0)} & 5.0 \textcolor{red}{(+5.0)} & 0.0 \textcolor{gray}{(+0.0)} & 0.0 \textcolor{blue}{(-15.0)} & 0.0 \textcolor{gray}{(+0.0)} \\
& Self-VCoT-fixed & 0.0 \textcolor{gray}{(+0.0)} & 0.0 \textcolor{gray}{(+0.0)}& 0.0 \textcolor{gray}{(+0.0)}& 0.0 \textcolor{gray}{(+0.0)}& 0.0 \textcolor{blue}{(-15.0)}& 0.0 \textcolor{gray}{(+0.0)}\\
\midrule

\multirow{2}{*}{LLaDA2.0-Uni}
& Self-VCoT & 10.0 \textcolor{red}{(+5.0)} & 10.0 \textcolor{red}{(+10.0)} & 5.0 \textcolor{red}{(+5.0)} & 0.0 \textcolor{gray}{(+0.0)} & 40.0 \textcolor{red}{(+35.0)} & 0.0 \textcolor{gray}{(+0.0)} \\
& Self-VCoT-fixed & 0.0 \textcolor{blue}{(-5.0)}& 0.0 \textcolor{gray}{(+0.0)}& 0.0 \textcolor{gray}{(+0.0)}& 0.0 \textcolor{gray}{(+0.0)}& 0.0 \textcolor{blue}{(-5.0)}& 0.0 \textcolor{gray}{(+0.0)}\\
\midrule

\multirow{2}{*}{Show-o2}
& Self-VCoT & 11.6 \textcolor{red}{(+6.6)} & 10.0 \textcolor{red}{(+10.0)} & 6.6 \textcolor{red}{(+6.6)} & 0.0 \textcolor{gray}{(+0.0)} & 56.6 \textcolor{red}{(+56.6)} & 0.0 \textcolor{gray}{(+0.0)} \\
& Self-VCoT-fixed &0.0 \textcolor{blue}{(-5.0)} & 0.0 \textcolor{gray}{(+0.0)}& 0.0 \textcolor{gray}{(+0.0)}& 0.0 \textcolor{gray}{(+0.0)}& 0.0 \textcolor{gray}{(+0.0)}& 0.0 \textcolor{gray}{(+0.0)}\\
\midrule

\multirow{2}{*}{BAGEL}
& Self-VCoT & 6.6 \textcolor{red}{(+1.6)} & 13.3 \textcolor{red}{(+13.3)} & 5.0 \textcolor{gray}{(+0.0)} & 0.0 \textcolor{gray}{(+0.0)} & 0.0 \textcolor{blue}{(-5.0)} & 0.0 \textcolor{gray}{(+0.0)} \\
& Self-VCoT-fixed & 5.0 \textcolor{gray}{(+0.0)} &0.0 \textcolor{gray}{(+0.0)} & 0.0 \textcolor{blue}{(-5.0)}& 0.0 \textcolor{gray}{(+0.0)}& 0.0 \textcolor{blue}{(-5.0)}& 0.0 \textcolor{gray}{(+0.0)}\\
\midrule

\multirow{2}{*}{LongCat-Next}
& Self-VCoT & 10.0 \textcolor{gray}{(+0.0)} & 5.0 \textcolor{red}{(+5.0)} & 1.6 \textcolor{blue}{(-8.4)} & 0.0 \textcolor{gray}{(+0.0)} & 5.0 \textcolor{blue}{(-10.0)} & 0.0 \textcolor{gray}{(+0.0)} \\
& Self-VCoT-fixed & 0.0 \textcolor{blue}{(-10.0)} & 5.0 \textcolor{red}{(+5.0)} & 0.0 \textcolor{blue}{(-10.0)} & 0.0 \textcolor{gray}{(+0.0)}& 0.0 \textcolor{blue}{(-15.0)}& 0.0 \textcolor{gray}{(+0.0)} \\
\bottomrule
\end{tabular}%
}
\caption{Accuracy (\%) of unified models on Maze, Sokoban, Path, Onet, Parking, and Klotski under Self-VCoT and fixed Self-VCoT evaluation settings on easy cases.}
\label{tab:corrected_self_vcot}
\end{table*}

\begin{figure*}[htbp]
    \small
    \centering
    \includegraphics[width=\textwidth]{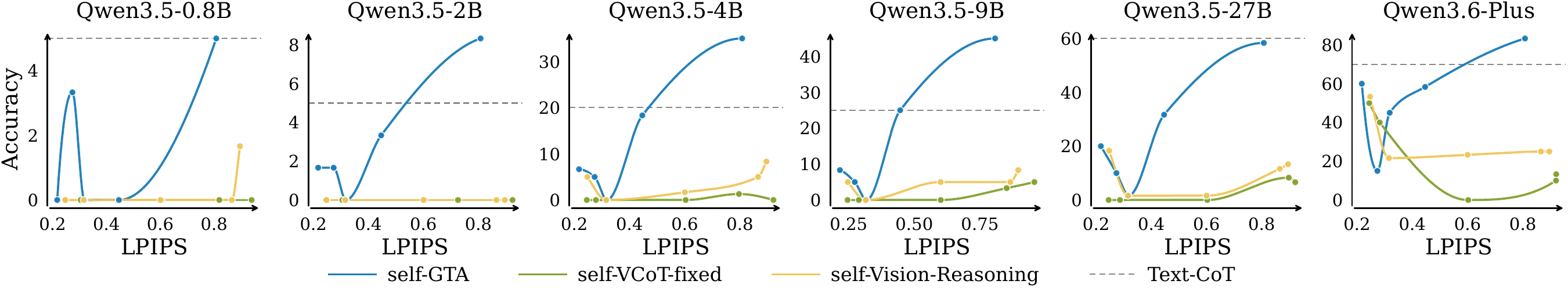}
    \caption{Effect of visual generation fidelity under fixed self-generation protocols across Qwen models. Accuracy is plotted against LPIPS for Self-GtA, Self-VCoT-fixed, Self-Vision-Reasoning, and Text-CoT.}
    \label{fig:result}
    \vspace{-3mm}
\end{figure*}

\vspace{-2mm}
\subsection{Fixing Self-VCoT Prior Leakage and Comparing Self-Generation Paradigms}
\vspace{-1mm}

As shown in Tables~\ref{tab:qwen_reasoners_maze}, \ref{tab:qwen_reasoners_sokoban}, and \ref{tab:qwen_reasoners_parking}, Self-VCoT exhibits relatively unstable behavior across different visual conditions. In particular, Self-VCoT can still \textbf{obtain unexpectedly high accuracy even when the intermediate visual states are replaced with task-irrelevant Gaussian noise}.

\textbf{This behavior suggests that the measured gain of Self-VCoT does not necessarily come from generated visual states.} In the implementation of Self-VCoT, the optimal number of steps is first obtained from the answer, and the model is then asked to perform interleaved generation for exactly this number of rounds, producing one action and one corresponding image at each round. This prior access to the optimal step count forces the generated answer to have the same length as the reference solution, thereby making the task easier to solve.

To fix this protocol-induced bias, we remove the prior access to the optimal step count from the Self-VCoT protocol. 
The model still performs interleaved generation one step at a round. 
After each round is completed, we ask the model to estimate the number of remaining steps. 
The interaction stops when the model predicts zero remaining steps. 
The results under this fixed protocol are reported in Table~\ref{tab:corrected_self_vcot}. We further apply the same fix to the disentangled diagnostic setting used in Table~\ref{tab:qwen_reasoners_maze}, and report the corrected results in Figure~\ref{fig:result}.

\noindent \textbf{Insight 4}: \textit{Effective visual generation should target the bottleneck of visual understanding rather than simply provide more visual reasoning steps.}

As shown in Table~\ref{tab:corrected_self_vcot}, performance under Self-VCoT-fixed declines substantially after removing the optimal-step prior. For example, Show-o2's Self-VCoT accuracy on Parking drops from $56.6\%$ to $0.0\%$ under the fixed protocol. Similarly, LLaDA2.0-Uni's Self-VCoT accuracy on Parking drops from $40.0\%$ to $0.0\%$ after the optimal-step prior is removed. Without this optimal-step prior, each intermediate visual state must be sufficiently accurate and useful for the next decision, making the iterative process more susceptible to accumulated errors. In contrast, Self-GtA remains more reliable because it directly targets the visual understanding bottleneck with a single auxiliary visual aid. These results suggest that current UMMs benefit more from targeted auxiliary visual generation than from repeatedly generating intermediate visual states. Overall, effective visual generation should address the visual understanding bottleneck rather than simply add more visual reasoning steps.

\vspace{-3mm}
\section{Conclusion}
\vspace{-3mm}
In this work, we studied whether visual generation can improve visual understanding in UMMs. To this end, we introduce \textit{VGAU-Diag}, a fine-grained evaluation framework that stratifies samples by difficulty, unifies the evaluation of multiple vision generation-assisted reasoning paradigms, and uses oracle-assisted protocols. Our results show that generated visual aids help more reliably on easier instances but become unstable as reasoning complexity increases. Oracle-assisted diagnosis further reveals that the main bottleneck often lies in visual understanding rather than generation, as current UMMs struggle to use even faithful visual aids. We also show that effective visual generation should target visual-understanding bottlenecks rather than add more reasoning steps and identify a three-stage transition from task-irrelevant interference, to misleading plausible guidance, and finally to effective assistance. Together, these findings would be useful to guide the development of better UMMs.

\section{Limitations}

Although our framework provides a controlled setting for evaluating generation-assisted visual understanding, it is limited to static two-dimensional planning environments. These tasks are well suited for analyzing spatial state parsing, planning horizons, and intermediate visual reasoning, but do not fully capture temporal dynamics, embodied interaction, or continuous three-dimensional scenes. In addition, the evaluation relies on procedurally generated tasks and structured visual states to ensure controllability, unique solutions, and reliable automatic evaluation. This design may reduce visual diversity and underrepresent the noise, ambiguity, and appearance variation commonly observed in natural images. Future work can extend to video-based planning, interactive environments, embodied decision-making, and more realistic visual inputs while preserving verifiable solutions.

\section{Acknowledgements}

We sincerely thank all the anonymous reviewers
for their constructive feedback. This work was supported in part by the National Natural Science Foundation of China (NSFC) under Grants No. 62372226 and 62272215.

\bibliography{custom}

@misc{nvidia2025nvidianemotronnanov2,
      title={NVIDIA Nemotron Nano V2 VL}, 
      author={NVIDIA and : and Amala Sanjay Deshmukh and Kateryna Chumachenko and Tuomas Rintamaki and Matthieu Le and Tyler Poon and Danial Mohseni Taheri and Ilia Karmanov and Guilin Liu and Jarno Seppanen and Guo Chen and Karan Sapra and Zhiding Yu and Adi Renduchintala and Charles Wang and Peter Jin and Arushi Goel and Mike Ranzinger and Lukas Voegtle and Philipp Fischer and Timo Roman and Wei Ping and Boxin Wang and Zhuolin Yang and Nayeon Lee and Shaokun Zhang and Fuxiao Liu and Zhiqi Li and Di Zhang and Greg Heinrich and Hongxu Yin and Song Han and Pavlo Molchanov and Parth Mannan and Yao Xu and Jane Polak Scowcroft and Tom Balough and Subhashree Radhakrishnan and Paris Zhang and Sean Cha and Ratnesh Kumar and Zaid Pervaiz Bhat and Jian Zhang and Darragh Hanley and Pritam Biswas and Jesse Oliver and Kevin Vasques and Roger Waleffe and Duncan Riach and Oluwatobi Olabiyi and Ameya Sunil Mahabaleshwarkar and Bilal Kartal and Pritam Gundecha and Khanh Nguyen and Alexandre Milesi and Eugene Khvedchenia and Ran Zilberstein and Ofri Masad and Natan Bagrov and Nave Assaf and Tomer Asida and Daniel Afrimi and Amit Zuker and Netanel Haber and Zhiyu Cheng and Jingyu Xin and Di Wu and Nik Spirin and Maryam Moosaei and Roman Ageev and Vanshil Atul Shah and Yuting Wu and Daniel Korzekwa and Unnikrishnan Kizhakkemadam Sreekumar and Wanli Jiang and Padmavathy Subramanian and Alejandra Rico and Sandip Bhaskar and Saeid Motiian and Kedi Wu and Annie Surla and Chia-Chih Chen and Hayden Wolff and Matthew Feinberg and Melissa Corpuz and Marek Wawrzos and Eileen Long and Aastha Jhunjhunwala and Paul Hendricks and Farzan Memarian and Benika Hall and Xin-Yu Wang and David Mosallanezhad and Soumye Singhal and Luis Vega and Katherine Cheung and Krzysztof Pawelec and Michael Evans and Katherine Luna and Jie Lou and Erick Galinkin and Akshay Hazare and Kaustubh Purandare and Ann Guan and Anna Warno and Chen Cui and Yoshi Suhara and Shibani Likhite and Seph Mard and Meredith Price and Laya Sleiman and Saori Kaji and Udi Karpas and Kari Briski and Joey Conway and Michael Lightstone and Jan Kautz and Mohammad Shoeybi and Mostofa Patwary and Jonathen Cohen and Oleksii Kuchaiev and Andrew Tao and Bryan Catanzaro},
      year={2025},
      eprint={2511.03929},
      archivePrefix={arXiv},
      primaryClass={cs.LG},
      url={https://arxiv.org/abs/2511.03929}, 
}

@inproceedings{dai2025see,
  title={See different, think better: Visual variations mitigating hallucinations in lvlms},
  author={Dai, Ziyun and Li, Xiaoqiang and Zhang, Shaohua and Wu, Yuanchen and Li, Jide},
  booktitle={Proceedings of the 33rd ACM International Conference on Multimedia},
  pages={3310--3319},
  year={2025}
}

@misc{wei2026youtuvlunleashingvisualpotential,
      title={Youtu-VL: Unleashing Visual Potential via Unified Vision-Language Supervision}, 
      author={Zhixiang Wei and Yi Li and Zhehan Kan and Xinghua Jiang and Zuwei Long and Shifeng Liu and Hongze Shen and Wei Liu and Xiaoyu Tan and Haojia Lin and Yubo Zhu and Qianyu Li and Di Yin and Haoyu Cao and Weibo Gu and Xin Li and Yinsong Liu and Deqiang Jiang and Xing Sun and Yunsheng Wu and Mingkong Tang and Shuangyin Liu and Lexiang Tang and Haodong Lin and Junru Lu and Jiarui Qin and Lingfeng Qiao and Ruizhi Qiao and Bo Ke and Jianfeng He and Ke Li and Yangning Li and Yunhang Shen and Mengdan Zhang and Peixian Chen and Kun Yin and Bing Liu and Yunfei Wu and Huang Chen and Zhongpeng Cai and Xiaotian Li},
      year={2026},
      eprint={2601.19798},
      archivePrefix={arXiv},
      primaryClass={cs.CV},
      url={https://arxiv.org/abs/2601.19798}, 
}

@misc{wu2025qwenimagetechnicalreport,
      title={Qwen-Image Technical Report}, 
      author={Chenfei Wu and Jiahao Li and Jingren Zhou and Junyang Lin and Kaiyuan Gao and Kun Yan and Sheng-ming Yin and Shuai Bai and Xiao Xu and Yilei Chen and Yuxiang Chen and Zecheng Tang and Zekai Zhang and Zhengyi Wang and An Yang and Bowen Yu and Chen Cheng and Dayiheng Liu and Deqing Li and Hang Zhang and Hao Meng and Hu Wei and Jingyuan Ni and Kai Chen and Kuan Cao and Liang Peng and Lin Qu and Minggang Wu and Peng Wang and Shuting Yu and Tingkun Wen and Wensen Feng and Xiaoxiao Xu and Yi Wang and Yichang Zhang and Yongqiang Zhu and Yujia Wu and Yuxuan Cai and Zenan Liu},
      year={2025},
      eprint={2508.02324},
      archivePrefix={arXiv},
      primaryClass={cs.CV},
      url={https://arxiv.org/abs/2508.02324}, 
}

@inproceedings{zhang-etal-2025-lmms,
    title = "{LMM}s-Eval: Reality Check on the Evaluation of Large Multimodal Models",
    author = "Zhang, Kaichen  and
      Li, Bo  and
      Zhang, Peiyuan  and
      Pu, Fanyi  and
      Cahyono, Joshua Adrian  and
      Hu, Kairui  and
      Liu, Shuai  and
      Zhang, Yuanhan  and
      Yang, Jingkang  and
      Li, Chunyuan  and
      Liu, Ziwei",
    editor = "Chiruzzo, Luis  and
      Ritter, Alan  and
      Wang, Lu",
    booktitle = "Findings of the Association for Computational Linguistics: NAACL 2025",
    month = apr,
    year = "2025",
    address = "Albuquerque, New Mexico",
    publisher = "Association for Computational Linguistics",
    url = "https://aclanthology.org/2025.findings-naacl.51/",
    doi = "10.18653/v1/2025.findings-naacl.51",
    pages = "881--916",
    ISBN = "979-8-89176-195-7"
}

@article{wang2025autoregressive,
  title={Autoregressive semantic visual reconstruction helps vlms understand better},
  author={Wang, Dianyi and Song, Wei and Wang, Yikun and Wang, Siyuan and Yu, Kaicheng and Wei, Zhongyu and Wang, Jiaqi},
  journal={arXiv preprint arXiv:2506.09040},
  year={2025}
}

@inproceedings{
wang2025reconstructive,
title={Reconstructive Visual Instruction Tuning},
author={Haochen Wang and Anlin Zheng and Yucheng Zhao and Tiancai Wang and Zheng Ge and Xiangyu Zhang and Zhaoxiang Zhang},
booktitle={The Thirteenth International Conference on Learning Representations},
year={2025},
url={https://openreview.net/forum?id=8q9NOMzRDg}
}

@article{shao2025anchoring,
  title={Anchoring Values in Temporal and Group Dimensions for Flow Matching Model Alignment},
  author={Shao, Yawen and Xiao, Jie and Zhu, Kai and Liu, Yu and Zhai, Wei and Cao, Yang and Zha, Zheng-Jun},
  journal={arXiv preprint arXiv:2512.12387},
  year={2025}
}

@misc{shao2026trackaligningrewardsstates,
      title={Back on Track: Aligning Rewards and States for Reasoning in Diffusion Large Language Models}, 
      author={Yawen Shao and Jie Xiao and Kai Zhu and Yu Liu and Hongchen Luo and Xueyang Fu and Yang Cao and Wei Zhai and Zheng-Jun Zha},
      year={2026},
      eprint={2606.08501},
      archivePrefix={arXiv},
      primaryClass={cs.CL},
      url={https://arxiv.org/abs/2606.08501}, 
}

@inproceedings{
li2025imagine,
title={Imagine While Reasoning in Space: Multimodal Visualization-of-Thought},
author={Chengzu Li and Wenshan Wu and Huanyu Zhang and Yan Xia and Shaoguang Mao and Li Dong and Ivan Vuli{\'c} and Furu Wei},
booktitle={Forty-second International Conference on Machine Learning},
year={2025},
url={https://openreview.net/forum?id=6vk6Xg24ZC}
}

@inproceedings{
fu2025refocus,
title={ReFocus: Visual Editing as a Chain of Thought for Structured Image Understanding},
author={Xingyu Fu and Minqian Liu and Zhengyuan Yang and John Richard Corring and Yijuan Lu and Jianwei Yang and Dan Roth and Dinei Florencio and Cha Zhang},
booktitle={Forty-second International Conference on Machine Learning},
year={2025},
url={https://openreview.net/forum?id=a7qFlPOTix}
}

@misc{wu2025vicbenchbenchmarkingvisualinterleavedchainofthought,
      title={ViC-Bench: Benchmarking Visual-Interleaved Chain-of-Thought Capability in MLLMs with Free-Style Intermediate State Representations}, 
      author={Xuecheng Wu and Jiaxing Liu and Danlei Huang and Yifan Wang and Yunyun Shi and Kedi Chen and Junxiao Xue and Yang Liu and Chunlin Chen and Hairong Dong and Dingkang Yang},
      year={2025},
      eprint={2505.14404},
      archivePrefix={arXiv},
      primaryClass={cs.CV},
      url={https://arxiv.org/abs/2505.14404}, 
}

@inproceedings{
leng2025crosswordbench,
title={CrossWordBench: Evaluating the Reasoning Capabilities of {LLM}s and {LVLM}s with Controllable Puzzle Generation},
author={Jixuan Leng and Chengsong Huang and Langlin Huang and Bill Yuchen Lin and William W. Cohen and Haohan Wang and Jiaxin Huang},
booktitle={Second Conference on Language Modeling},
year={2025},
url={https://openreview.net/forum?id=bJCQMKwPVq}
}

@misc{cheng2025comtnovelbenchmarkchain,
      title={CoMT: A Novel Benchmark for Chain of Multi-modal Thought on Large Vision-Language Models}, 
      author={Zihui Cheng and Qiguang Chen and Jin Zhang and Hao Fei and Xiaocheng Feng and Wanxiang Che and Min Li and Libo Qin},
      year={2025},
      eprint={2412.12932},
      archivePrefix={arXiv},
      primaryClass={cs.CV},
      url={https://arxiv.org/abs/2412.12932}, 
}

@misc{bai2025qwen3vltechnicalreport,
      title={Qwen3-VL Technical Report}, 
      author={Shuai Bai and Yuxuan Cai and Ruizhe Chen and Keqin Chen and Xionghui Chen and Zesen Cheng and Lianghao Deng and Wei Ding and Chang Gao and Chunjiang Ge and Wenbin Ge and Zhifang Guo and Qidong Huang and Jie Huang and Fei Huang and Binyuan Hui and Shutong Jiang and Zhaohai Li and Mingsheng Li and Mei Li and Kaixin Li and Zicheng Lin and Junyang Lin and Xuejing Liu and Jiawei Liu and Chenglong Liu and Yang Liu and Dayiheng Liu and Shixuan Liu and Dunjie Lu and Ruilin Luo and Chenxu Lv and Rui Men and Lingchen Meng and Xuancheng Ren and Xingzhang Ren and Sibo Song and Yuchong Sun and Jun Tang and Jianhong Tu and Jianqiang Wan and Peng Wang and Pengfei Wang and Qiuyue Wang and Yuxuan Wang and Tianbao Xie and Yiheng Xu and Haiyang Xu and Jin Xu and Zhibo Yang and Mingkun Yang and Jianxin Yang and An Yang and Bowen Yu and Fei Zhang and Hang Zhang and Xi Zhang and Bo Zheng and Humen Zhong and Jingren Zhou and Fan Zhou and Jing Zhou and Yuanzhi Zhu and Ke Zhu},
      year={2025},
      eprint={2511.21631},
      archivePrefix={arXiv},
      primaryClass={cs.CV},
      url={https://arxiv.org/abs/2511.21631}, 
}

@misc{qin2025starstackedautoregressivescheme,
      title={STAR: STacked AutoRegressive Scheme for Unified Multimodal Learning}, 
      author={Jie Qin and Jiancheng Huang and Limeng Qiao and Lin Ma},
      year={2025},
      eprint={2512.13752},
      archivePrefix={arXiv},
      primaryClass={cs.CV},
      url={https://arxiv.org/abs/2512.13752}, 
}

@InProceedings{Wu_2026_CVPR,
    author    = {Wu, Chenyuan and Wang, Jiahao and Zheng, Pengfei and Yan, Ruiran and Xiao, Shitao and Luo, Xin and Wang, Yueze and Li, Wanli and Jiang, Xiyan and Liu, Yexin and Zhou, Junjie and Xia, Ziyi and Liu, Ze and Li, Chaofan and Deng, Haoge and Luo, Kun and Zhang, Bo and Zhang, Jiajun and Liu, Dong and Lian, Defu and Wang, Xinlong and Wang, Zhongyuan and Huang, Tiejun and Liu, Zheng},
    title     = {OmniGen2: Towards Instruction-Aligned Multimodal Generation},
    booktitle = {Proceedings of the IEEE/CVF Conference on Computer Vision and Pattern Recognition (CVPR)},
    month     = {June},
    year      = {2026},
    pages     = {21964-21975}
}

@misc{qwen2026qwen35,
    author = {{QwenTeam}},
    title = {Qwen3.5: Towards Native Multimodal Agents},
    year = {2026},
    howpublished = {\url{https://qwen.ai/blog?id=qwen3.5}},
}

@misc{geminiteam2026gemini31,
  author       = {{The Gemini Team}},
  title        = {Gemini 3.1 Pro: A smarter model for your most complex tasks},
  howpublished = {\url{https://blog.google/innovation-and-ai/models-and-research/gemini-models/gemini-3-1-pro/}},
  year         = {2026},
  month        = {February},
}

@misc{qwen2026qwen36,
    author = {{QwenTeam}},
    title = {Qwen3.6-Plus: Towards Real World Agents},
    year = {2026},
    howpublished = {\url{https://qwen.ai/blog?id=qwen3.6}},
}

@misc{openaiteam2026gpt54,
  author       = {{OpenAI}},
  title        = {Introducing GPT-5.4},
  howpublished = {\url{https://openai.com/index/introducing-gpt-5-4/}},
  year         = {2026},
  month        = {March},
}

@misc{seedteam2026seed20,
  author       = {{The Seed Team}},
  title        = {Seed 2.0 Official Launch},
  howpublished = {\url{https://seed.bytedance.com/en/blog/seed-2-0-official-launch}},
  year         = {2026},
  month        = {February},
  note         = {ByteDance Seed Blog}
}

@misc{bai2025qwen25vltechnicalreport,
      title={Qwen2.5-VL Technical Report}, 
      author={Shuai Bai and Keqin Chen and Xuejing Liu and Jialin Wang and Wenbin Ge and Sibo Song and Kai Dang and Peng Wang and Shijie Wang and Jun Tang and Humen Zhong and Yuanzhi Zhu and Mingkun Yang and Zhaohai Li and Jianqiang Wan and Pengfei Wang and Wei Ding and Zheren Fu and Yiheng Xu and Jiabo Ye and Xi Zhang and Tianbao Xie and Zesen Cheng and Hang Zhang and Zhibo Yang and Haiyang Xu and Junyang Lin},
      year={2025},
      eprint={2502.13923},
      archivePrefix={arXiv},
      primaryClass={cs.CV},
      url={https://arxiv.org/abs/2502.13923}, 
}

@misc{ai2026llada20uniunifyingmultimodalunderstanding,
      title={LLaDA2.0-Uni: Unifying Multimodal Understanding and Generation with Diffusion Large Language Model}, 
      author={{Inclusion AI} and Tiwei Bie and Haoxing Chen and Tieyuan Chen and Zhenglin Cheng and Long Cui and Kai Gan and Zhicheng Huang and Zhenzhong Lan and Haoquan Li and Jianguo Li and Tao Lin and Qi Qin and Hongjun Wang and Xiaomei Wang and Haoyuan Wu and Yi Xin and Junbo Zhao},
      year={2026},
      eprint={2604.20796},
      archivePrefix={arXiv},
      primaryClass={cs.CV},
      url={https://arxiv.org/abs/2604.20796}, 
}

@misc{fortin2025nanobanana,
  author       = {Alisa Fortin and Guillaume Vernade and Kat Kampf and Ammaar Reshi},
  title        = {Introducing Gemini 2.5 Flash Image, our state-of-the-art image model},
  year         = {2025},
  month        = {8},
  howpublished = {\url{https://developers.googleblog.com/en/introducing-gemini-2-5-flash-image/}},
}

@misc{raisinghani2026nanobanana2,
  author = {Naina Raisinghani},
  title = {Nano {B}anana 2: {C}ombining {P}ro capabilities with lightning-fast speed},
  howpublished = {\url{https://blog.google/innovation-and-ai/technology/ai/nano-banana-2/}},
  year = {2026},
  month = {February},
  day = {26},
  note = {Google Blog. Accessed: 2026-05-09}
}

@misc{meituanlongcatteam2026longcatnextlexicalizingmodalitiesdiscrete,
      title={LongCat-Next: Lexicalizing Modalities as Discrete Tokens}, 
      author={{Meituan LongCat Team} and Bin Xiao and Chao Wang and Chengjiang Li and Chi Zhang and Chong Peng and Hang Yu and Hao Yang and Haonan Yan and Haoze Sun and Haozhe Zhao and Hong Liu and Hui Su and Jiaqi Zhang and Jiawei Wang and Jing Li and Kefeng Zhang and Manyuan Zhang and Minhao Jing and Peng Pei and Quan Chen and Taofeng Xue and Tongxin Pan and Xiaotong Li and Xiaoyang Li and Xiaoyu Zhao and Xing Hu and Xinyang Lin and Xunliang Cai and Yan Bai and Yan Feng and Yanjie Li and Yao Qiu and Yerui Sun and Yifan Lu and Ying Luo and Yipeng Mei and Yitian Chen and Yuchen Xie and Yufang Liu and Yufei Chen and Yulei Qian and Yuqi Peng and Zhihang Yu and Zhixiong Han and Changran Wang and Chen Chen and Dian Zheng and Fengjiao Chen and Ge Yang and Haowei Guo and Haozhe Wang and Hongyu Li and Huicheng Jiang and Jiale Hong and Jialv Zou and Jiamu Li and Jianping Lin and Jiaxing Liu and Jie Yang and Jing Jin and Jun Kuang and Juncheng She and Kunming Luo and Kuofeng Gao and Lin Qiu and Linsen Guo and Mianqiu Huang and Qi Li and Qian Wang and Rumei Li and Siyu Ren and Wei Wang and Wenlong He and Xi Chen and Xiao Liu and Xiaoyu Li and Xu Huang and Xuanyu Zhu and Xuezhi Cao and Yaoming Zhu and Yifei Cao and Yimeng Jia and Yizhen Jiang and Yufei Gao and Zeyang Hu and Zhenlong Yuan and Zijian Zhang and Ziwen Wang},
      year={2026},
      eprint={2603.27538},
      archivePrefix={arXiv},
      primaryClass={cs.CV},
      url={https://arxiv.org/abs/2603.27538}, 
}

@inproceedings{
        xie2025showo,
        title={Show-o2: Improved Native Unified Multimodal Models},
        author={Jinheng Xie and Zhenheng Yang and Mike Zheng Shou},
        booktitle={The Thirty-ninth Annual Conference on Neural Information Processing Systems},
        year={2025},
        url={https://openreview.net/forum?id=7VMg7Jb7AL}
}

@misc{zhu2025llmknowsestimatingllmperceived,
      title={The LLM Already Knows: Estimating LLM-Perceived Question Difficulty via Hidden Representations}, 
      author={Yubo Zhu and Dongrui Liu and Zecheng Lin and Wei Tong and Sheng Zhong and Jing Shao},
      year={2025},
      eprint={2509.12886},
      archivePrefix={arXiv},
      primaryClass={cs.CL},
      url={https://arxiv.org/abs/2509.12886}, 
}

@inproceedings{
xie2026mmeunify,
title={{MME}-Unify: A Comprehensive Benchmark for Unified Multimodal Understanding and Generation Models},
author={Wulin Xie and YiFan Zhang and Chaoyou Fu and Yang Shi and Jianshu Zeng and Bingyan Nie and Hongkai Chen and Zhang Zhang and Liang Wang},
booktitle={The Fourteenth International Conference on Learning Representations},
year={2026},
url={https://openreview.net/forum?id=7x6TxVIarj}
}

@misc{wen2026unig2ubenchunifiedmodelsadvance,
      title={UniG2U-Bench: Do Unified Models Advance Multimodal Understanding?}, 
      author={Zimo Wen and Boxiu Li and Wanbo Zhang and Junxiang Lei and Xiaoyu Chen and Yijia Fan and Qi Zhang and Yujiang Wang and Lili Qiu and Bo Li and Ziwei Liu and Caihua Shan and Yifan Yang and Yifei Shen},
      year={2026},
      eprint={2603.03241},
      archivePrefix={arXiv},
      primaryClass={cs.CV},
      url={https://arxiv.org/abs/2603.03241}, 
}

@misc{deng2025emergingpropertiesunifiedmultimodal,
      title={Emerging Properties in Unified Multimodal Pretraining}, 
      author={Chaorui Deng and Deyao Zhu and Kunchang Li and Chenhui Gou and Feng Li and Zeyu Wang and Shu Zhong and Weihao Yu and Xiaonan Nie and Ziang Song and Guang Shi and Haoqi Fan},
      year={2025},
      eprint={2505.14683},
      archivePrefix={arXiv},
      primaryClass={cs.CV},
      url={https://arxiv.org/abs/2505.14683}, 
}

@inproceedings{
zeller2026mentisoculi,
title={MentisOculi: Revealing the Limits of Reasoning with Mental Imagery},
author={Jana Ricarda Zeller and Thadd{\"a}us Wiedemer and Fanfei Li and Thomas Klein and Prasanna Mayilvahanan and Matthias Bethge and Felix A. Wichmann and Ryan Cotterell and Wieland Brendel},
booktitle={Forty-third International Conference on Machine Learning},
year={2026},
url={https://openreview.net/forum?id=sxvuK2x3eA}
}

@inproceedings{zou-etal-2026-uni,
    title = "Uni-{MMMU}: A Massive Multi-discipline Multimodal Unified Benchmark",
    author = "Zou, Kai  and
      Huang, Ziqi  and
      Dong, Yuhao  and
      Tian, Shulin  and
      Zheng, Dian  and
      Liu, Hongbo  and
      He, Jingwen  and
      Liu, Bin  and
      Qiao, Yu  and
      Liu, Ziwei",
    editor = "Liakata, Maria  and
      Moreira, Viviane P.  and
      Zhang, Jiajun  and
      Jurgens, David",
    booktitle = "Proceedings of the 64th Annual Meeting of the {A}ssociation for {C}omputational {L}inguistics (Volume 1: Long Papers)",
    month = jul,
    year = "2026",
    address = "San Diego, California, United States",
    publisher = "Association for Computational Linguistics",
    url = "https://aclanthology.org/2026.acl-long.40/",
    doi = "10.18653/v1/2026.acl-long.40",
    pages = "908--924",
    ISBN = "979-8-89176-390-6"
}

@misc{chen2025janusprounifiedmultimodalunderstanding,
      title={Janus-Pro: Unified Multimodal Understanding and Generation with Data and Model Scaling}, 
      author={Xiaokang Chen and Zhiyu Wu and Xingchao Liu and Zizheng Pan and Wen Liu and Zhenda Xie and Xingkai Yu and Chong Ruan},
      year={2025},
      eprint={2501.17811},
      archivePrefix={arXiv},
      primaryClass={cs.AI},
      url={https://arxiv.org/abs/2501.17811}, 
}

@misc{shen2025mammothmoda2unifiedardiffusionframework,
      title={MammothModa2: A Unified AR-Diffusion Framework for Multimodal Understanding and Generation}, 
      author={Tao Shen and Xin Wan and Taicai Chen and Rui Zhang and Junwen Pan and Dawei Lu and Fanding Lei and Zhilin Lu and Yunfei Yang and Chen Cheng and Qi She and Chang Liu and Zhenbang Sun},
      year={2025},
      eprint={2511.18262},
      archivePrefix={arXiv},
      primaryClass={cs.CV},
      url={https://arxiv.org/abs/2511.18262}, 
}

@inproceedings{
liang2026rover,
title={{ROVER}: Benchmarking Reciprocal Cross-Modal Reasoning for Omnimodal Generation},
author={Yongyuan Liang and Wei Chow and Feng Li and Ziqiao Ma and Xiyao Wang and Jiageng Mao and Jiuhai Chen and Jiatao Gu and Yue Wang and Furong Huang},
booktitle={The Fourteenth International Conference on Learning Representations},
year={2026},
url={https://openreview.net/forum?id=gu3DRaDWiI}
}

@InProceedings{Shi_2026_CVPR,
    author    = {Shi, Yang and Dong, Yuhao and Ding, Yue and Wang, Yuran and Zhu, Xuanyu and Zhou, Sheng and Liu, Wenting and Tian, Haochen and Wang, Rundong and Wang, Huanqian and Liu, Zuyan and Zeng, Bohan and Chen, Ruizhe and Wang, Qixun and Zhang, Zhuoran and Chen, Xinlong and Tong, Chengzhuo and Li, Bozhou and Liu, Qiang and Wang, Haotian and Yang, Wenjing and Zhang, Yuanxing and Wan, Pengfei and Zhang, Yi-Fan and Liu, Ziwei},
    title     = {RealUnify: Do Unified Models Truly Benefit from Unification? A Comprehensive Benchmark},
    booktitle = {Proceedings of the IEEE/CVF Conference on Computer Vision and Pattern Recognition (CVPR)},
    month     = {June},
    year      = {2026},
    pages     = {22488-22497}
}

\clearpage

\appendix
\section{Appendix}
\label{sec:appendix}

\subsection{Tasks Details}
\label{sec:task_details}

\subsubsection{Maze}

\textbf{Task Definition.} Given an image of a $10 \times 10$ maze, the model must move the agent from the blue start marker to the green goal marker. Black cells are walls, off-white cells are walkable, and the green goal cell is also passable. The only legal actions are \texttt{up}, \texttt{down}, \texttt{left}, and \texttt{right}, each moving the agent by exactly one grid cell. The model outputs a final action sequence as a JSON array wrapped in \texttt{<ANSWER\_JSON>...</ANSWER\_JSON>}. In the step-by-step setting, the model additionally generates one state image after each move, while preserving the original maze layout and moving only the blue agent marker.

\textbf{Construction.} We procedurally generate 100 mazes with controlled difficulty. Each maze is rendered on a $10 \times 10$ map with a fixed color palette: dark cells for walls, off-white cells for paths, a blue square for the agent, and a green hollow square for the goal. To create an instance, we first sample a simple path of a target length using randomized depth-first search. We then grow the set of walkable cells by repeatedly adding leaf cells, i.e., cells adjacent to exactly one existing walkable cell. This keeps the walkable-cell graph acyclic, guaranteeing a unique path between any two connected cells. We verify each sample by BFS and by explicitly counting simple paths from start to goal, keeping only mazes whose unique shortest path length falls into the desired bucket. The dataset contains easy cases with path lengths 1--6, medium cases with path lengths 7--17, and hard cases with path lengths 18--30.

\textbf{Evaluation.} We parse the model output by extracting the final JSON action list from \texttt{<ANSWER\_JSON>...<ANSWER\_JSON>} and normalizing each action token. Starting from the annotated start cell, we replay the predicted moves on the ground-truth walkable-cell set. A prediction is invalid if it contains an unknown action, moves outside the grid, or enters a wall. We report \texttt{reach\_any} when the replayed trajectory is legal and ends at the goal, regardless of length. We report \texttt{reach\_optimal} when the trajectory reaches the goal and its length equals the BFS shortest-path length.

\subsubsection{Sokoban}

\textbf{Task Definition.} Given an image of a $10 \times 10$ single-box Sokoban puzzle, the model must control the player to push the wooden box onto the green target cell. Brick cells are walls, sand cells are walkable, the small person marks the player start, and the wooden crate marks the box. The only legal actions are \texttt{up}, \texttt{down}, \texttt{left}, and \texttt{right}, each moving the player by exactly one grid cell. If the player moves into the box, the box is pushed one cell in the same direction, and the push is legal only when the destination cell is not a wall. The model outputs a final action sequence as a JSON array wrapped in \texttt{<ANSWER\_JSON>...</ANSWER\_JSON>}. In the step-by-step setting, the model additionally generates one state image after each move, while preserving the original layout and updating only the player and box positions.

\textbf{Construction.} We procedurally generate 100 Sokoban puzzles with controlled difficulty. Each puzzle is rendered on a $10 \times 10$ map with a fixed visual style for walls, floor, player, box, and target. Each instance contains exactly one player, one movable box, and one target. We sample candidate wall layouts and object placements, then solve each candidate with a programmatic Sokoban solver to obtain the shortest valid action sequence. Unsolvable puzzles and puzzles whose optimal solution length falls outside the desired range are discarded. The dataset contains easy cases with optimal lengths 1--8, medium cases with optimal lengths 10--16, and hard cases with optimal lengths 20--39.

\textbf{Evaluation.} We parse the model output by extracting the final JSON action list from \texttt{<ANSWER\_JSON>...</ANSWER\_JSON>} and normalizing each action token. Starting from the annotated player and box positions, we replay the predicted moves on the ground-truth grid. A prediction is invalid if it contains an unknown action, moves the player into a wall, or attempts to push the box into a wall. We report \texttt{reach\_any} when the replayed trajectory is legal and the box ends on the target cell, regardless of length. We report \texttt{reach\_optimal} when the puzzle is solved and the predicted action sequence length equals the annotated optimal solution length.

\subsubsection{Path}

\textbf{Task Definition.} Given an image of a $10 \times 10$ map, the model must identify which labeled points are passed along the valid simple path from the orange triangle to the blue triangle. Black cells are walls, light-gray cells are walkable, the orange triangle marks the start, and the blue triangle marks the destination. Five walkable cells are labeled with letters A, B, C, D, and E. Legal movement is restricted to \texttt{up}, \texttt{down}, \texttt{left}, and \texttt{right}, one cell per step, without entering black cells. The path is required to be simple, with no revisits or backtracking. The model outputs only the passed labels in path order, wrapped in \texttt{<answer>...</answer>}. In the step-by-step setting, the model additionally generates one state image after each move, while preserving the original layout and moving only the current-position marker.

\textbf{Construction.} We procedurally generate 100 path-following instances with controlled difficulty. Each instance is rendered on a $10 \times 10$ map with a fixed visual style for walls, walkable cells, start, destination, and letter labels. We sample a valid simple path between the start and destination, place five labeled points on walkable cells, and record the subset of labels that lie on the path in traversal order as the ground-truth answer. Candidate instances are filtered to ensure that the path does not cross walls, uses only orthogonal moves, and satisfies the target path-length range. The dataset contains easy cases with path lengths 2--5, medium cases with path lengths 8--14, and hard cases with path lengths 18--25.

\textbf{Evaluation.} We parse the model output by extracting the label sequence from \texttt{<answer>...</answer>} and normalizing all labels to uppercase letters among A--E. A prediction is considered correct only if the extracted label sequence exactly matches the ground-truth labels in both content and order. We report accuracy separately over easy, medium, and hard subsets, as well as over the full dataset.

\subsubsection{Parking}

\textbf{Task Definition.} Given an image of an $8 \times 8$ parking-exit puzzle, the model must slide vehicles to move the red car \texttt{R} to the exit on the right boundary. Each vehicle has a fixed orientation, either horizontal or vertical, and can only move along that orientation. In one move, exactly one vehicle slides by one grid cell. Vehicles cannot overlap or leave the board. Success is achieved when the rightmost cell of the red car reaches the right boundary on the exit row. The model outputs a final move sequence as a JSON array of objects wrapped in \texttt{<ANSWER\_JSON>...</ANSWER\_JSON>}, where each object specifies the vehicle id and direction, e.g., \texttt{\{"car":"A","direction":"up"\}}. In the step-by-step setting, the model additionally generates one state image after each move, while preserving the original layout and updating only the moved vehicle.

\textbf{Construction.} We procedurally generate 100 parking-exit puzzles with controlled difficulty. Each puzzle is rendered on an $8 \times 8$ board with a fixed visual style for the board, exit, and labeled vehicles. The red car \texttt{R} is always horizontal and aligned with the exit row. We sample vehicle sets with different numbers of cars, orientations, and lengths, initialize solvable states, and then scramble them through legal vehicle moves. For each candidate puzzle, we run BFS to compute the shortest solution and discard instances that are unsolvable or outside the target step range. Additional filtering enforces increasing structural complexity, such as more blocking vehicles in the red car's exit corridor and more non-red vehicles that must be moved. The dataset contains easy cases with optimal lengths 1--5, medium cases with optimal lengths 8--16, and hard cases with optimal lengths 18--34.

\textbf{Evaluation.} We parse the model output by extracting the JSON move list from \texttt{<ANSWER\_JSON>...</ANSWER\_JSON>} and normalizing vehicle ids and directions. Starting from the annotated vehicle positions, we replay the predicted moves on the ground-truth board. A prediction is invalid if it names an unknown vehicle, moves a vehicle against its orientation, causes overlap, or moves any vehicle outside the board. We report \texttt{reach\_solved} when the replayed trajectory is legal and the red car reaches the exit. We report \texttt{reach\_optimal} when the puzzle is solved and the predicted move sequence length equals the annotated shortest solution length.

\subsubsection{Onet}

\textbf{Task Definition.} Given an image of a $6 \times 6$ Onet board, the model must decide a sequence of pair removals that clears the entire board. Each occupied cell contains a fruit icon, and empty cells can be used as connection space. In each step, the model may remove two identical fruit icons if they can be connected by a path composed of horizontal and vertical segments with at most two turns. The connection path must not pass through any other fruit. The model outputs the full removal sequence as a JSON array of coordinate pairs wrapped in \texttt{<answer\_json>...</answer\_json>}, where each step has the form \texttt{[[r1,c1],[r2,c2]]}. In the step-by-step setting, the model additionally generates one board image after each pair removal, while preserving the original layout and clearing only the removed fruit cells.

\textbf{Construction.} We construct 100 solvable Onet puzzles with controlled difficulty. Each puzzle is rendered on a $6 \times 6$ board with a fixed image-only style and fruit icons as tile types. Each fruit type appears exactly twice, forming removable pairs, while remaining cells may be empty. For each instance, we store the initial board, the fruit legend, and one valid clearing sequence. Candidate boards are filtered so that all pairs can be removed under the Onet rule: each removal must connect identical fruits through empty cells using at most two turns. The dataset contains easy cases with 4--6 removal steps, medium cases with 7--10 removal steps, and hard cases with 12--16 removal steps.

\textbf{Evaluation.} We parse the model output by extracting the JSON removal sequence from \texttt{<answer\_json>...</answer\_json>}. Starting from the annotated board, we replay each predicted pair removal. A prediction is invalid if either coordinate is out of bounds, the two cells are empty, the two fruits are not identical, or the pair cannot be connected by a valid path with at most two turns through empty cells. After each valid step, the two fruit cells are cleared. We report accuracy when the entire predicted sequence is legal and the board is fully cleared.

\subsubsection{Klotski}

\textbf{Task Definition.} Given an image of a $6 \times 6$ Klotski board, the model must move the $2 \times 2$ target block \texttt{T} to the goal position. The board contains one target block, multiple $1 \times 2$ vertical blocks, multiple $2 \times 1$ horizontal blocks, several $1 \times 1$ single blocks, and exactly two empty cells. In each step, the model slides exactly one block by one cell in one of \texttt{UP}, \texttt{DOWN}, \texttt{LEFT}, or \texttt{RIGHT}. Blocks cannot rotate, jump, move diagonally, leave the board, or overlap other blocks. A block is identified in the output by its current top-left coordinate at the time of the move. The goal is achieved when the top-left coordinate of the target block reaches $(4,2)$. The model outputs the full move sequence as a JSON object with a \texttt{solution} field, where each step has the form \texttt{\{"from":[r,c],"direction":"UP"\}}. In the step-by-step setting, the model additionally generates one board image after each move, while preserving the original layout and updating only the moved block.

\textbf{Construction.} We construct 100 Klotski puzzles with controlled difficulty. Each puzzle is rendered on a $6 \times 6$ board with a fixed color palette for empty cells, the target block, vertical blocks, horizontal blocks, and single blocks. All instances use the same block inventory: one $2 \times 2$ target block, eight vertical blocks, three horizontal blocks, and eight single blocks, leaving exactly two empty cells. We sample solvable states by starting from goal configurations and walking backward through legal one-cell block slides, then reverse the sampled trajectory to obtain a valid solution from the initial state. Duplicate states and already-solved states are discarded. The dataset contains easy cases with optimal lengths 1--4, medium cases with optimal lengths 6--10, and hard cases with optimal lengths 15--20.

\textbf{Evaluation.} We parse the model output by extracting the JSON object containing the \texttt{solution} list. Starting from the annotated initial state, we replay each predicted move. A prediction is invalid if the referenced top-left coordinate does not correspond to a movable block in the current state, the direction is invalid, the move leaves the board, or any destination cell is occupied by another block. We report \texttt{solved} when the replayed trajectory is legal and the target block reaches $(4,2)$. We report \texttt{optimal} when the puzzle is solved and the predicted move sequence length equals the annotated optimal solution length.

\subsubsection{Difficulty validation.}
Optimal solution length serves as a controllable proxy for difficulty stratification rather than the sole definition of difficulty. To illustrate that this split captures broader structural complexity, we take Maze as an example. We compute instance-level and ground-truth path-level statistics across difficulty levels. Walkable cells measure the size of the reachable state space, branch nodes are walkable cells with at least three walkable neighbors, and corridor nodes are walkable cells with exactly two walkable neighbors. As shown in Tables~\ref{tab:maze-structure} and~\ref{tab:maze-path-structure}, all metrics increase consistently from easy to hard cases, showing that the length-based split reflects structural complexity rather than merely longer output sequences.

\begin{table}[t]
\centering
\small
\begin{tabular}{lccc}
\hline
Average metric & Easy & Medium & Hard \\
\hline
Walkable cells & 20.30 & 30.10 & 39.12 \\
Branch nodes & 4.25 & 5.83 & 6.98 \\
Corridor nodes & 9.35 & 16.07 & 22.76 \\
\hline
\end{tabular}
\caption{Structural statistics of Maze instances.}
\label{tab:maze-structure}
\end{table}

\begin{table}[t]
\centering
\small
\begin{tabular}{lccc}
\hline
Average path metric & Easy & Medium & Hard \\
\hline
Path nodes & 4.10 & 13.27 & 24.64 \\
Branch nodes on path & 1.55 & 3.70 & 6.00 \\
Corridor nodes on path & 2.00 & 8.97 & 18.00 \\
\hline
\end{tabular}
\caption{Ground-truth solution-path statistics.}
\label{tab:maze-path-structure}
\end{table}

\subsection{Protocol Details}
\label{appendix:protocol_detail}

\subsubsection{Text-CoT}

\paragraph{Maze. } 
In this setting, the prompt asks the model to first reason in text about the maze  and planned move sequence. The prompt is detailed in Table~\ref{tab:maze-Text-CoT}.

\paragraph{Sokoban. }
In this setting, the prompt asks the model to first reason in text about the player position, box position, target cell, and planned move sequence. The prompt is detailed in Table~\ref{tab:sokoban-Text-CoT}.

\paragraph{Path. }
In this setting, the prompt asks the model to first reason in text about the start point, destination point, valid simple path, and labeled points passed along the route. The prompt is detailed in Table~\ref{tab:path-Text-CoT}.

\paragraph{Parking. }
In this setting, the prompt asks the model to first reason in text about the vehicle layout, blocking vehicles, exit row, and planned slide sequence for moving the red car to the exit. The prompt is detailed in Table~\ref{tab:parking-Text-CoT}.

\paragraph{Onet. }
In this setting, the prompt asks the model to first reason in text about matching fruit pairs, valid connection paths, and the planned pair-removal sequence for clearing the board. The prompt is detailed in Table~\ref{tab:onet-Text-CoT}.

\paragraph{Klotski. }
In this setting, the prompt asks the model to first reason in text about the block layout, empty cells, target block position, and planned sequence of legal block slides. The prompt is detailed in Table~\ref{tab:klotski-Text-CoT}.

\subsubsection{Self-GtA}

\paragraph{Maze. }
In this setting, the model is first prompted to generate an auxiliary annotated maze image by adding red grid-boundary lines and coordinate labels while preserving the original maze layout, walls, start marker, and goal marker. The model then uses the generated auxiliary image to reason about the maze and output the final move sequence. The prompt is detailed in Table~\ref{tab:maze-Self-GtA}.

\paragraph{Sokoban. }
In this setting, the model is first prompted to generate an auxiliary annotated Sokoban image by adding red grid-boundary lines and coordinate labels while preserving the original walls, floor, player, box, and target. The model then uses the generated auxiliary image to reason about box-pushing dynamics and output the final move sequence. The prompt is detailed in Table~\ref{tab:sokoban-Self-GtA}.

\paragraph{Path. }
In this setting, the model is first prompted to generate an auxiliary annotated path image by adding red grid-boundary lines and coordinate labels while preserving the original walls, start marker, destination marker, and labeled points. The model then uses the generated auxiliary image to reason about the valid simple path and output the labels passed along the route. The prompt is detailed in Table~\ref{tab:path-Self-GtA}.

\paragraph{Parking. }
In this setting, the model is first prompted to generate an auxiliary annotated parking image by adding red grid-boundary lines and coordinate labels while preserving the original board, exit, vehicle positions, labels, and colors. The model then uses the generated auxiliary image to reason about the blocking vehicles and output the final vehicle-slide sequence. The prompt is detailed in Table~\ref{tab:parking-Self-GtA}.

\paragraph{Onet. }
In this setting, the model is first prompted to generate an auxiliary annotated Onet image by adding red grid-boundary lines and coordinate labels while preserving the original board geometry and fruit icons. The model then uses the generated auxiliary image to reason about valid matching pairs and output the final pair-removal sequence. The prompt is detailed in Table~\ref{tab:onet-Self-GtA}.

\paragraph{Klotski. }
In this setting, the model is first prompted to generate an auxiliary annotated Klotski image by adding red grid-boundary lines and coordinate labels while preserving the original board geometry, block positions, labels, and colors. The model then uses the generated auxiliary image to reason about legal block slides and output the final solution sequence. The prompt is detailed in Table~\ref{tab:klotski-Self-GtA}.

\subsubsection{self-VCoT}

\paragraph{Maze. }
In this setting, the model solves the maze iteratively by predicting one move at each step. After each predicted move, the model generates the corresponding next-state image, where only the agent marker should move while the maze layout remains unchanged. The process continues until the model reaches the goal and outputs the final move sequence. The prompt is detailed in Table~\ref{tab:maze-self-VCoT}.

\paragraph{Sokoban. }
In this setting, the model solves the Sokoban puzzle iteratively by predicting one player move at each step. After each predicted move, the model generates the corresponding next-state image, updating the player and box positions while preserving the walls, floor, and target. The process continues until the box reaches the target and the model outputs the final move sequence. The prompt is detailed in Table~\ref{tab:sokoban-self-VCoT}.

\paragraph{Path. }
In this setting, the model traces the path iteratively by predicting one movement step at a time. After each predicted move, the model generates the corresponding next-state image, where only the current-position marker should move while the map and labels remain unchanged. The process continues until the destination is reached, and the model outputs the labels passed along the path. The prompt is detailed in Table~\ref{tab:path-self-VCoT}.

\paragraph{Parking. }
In this setting, the model solves the parking-exit puzzle iteratively by predicting one vehicle slide at each step. After each predicted move, the model generates the corresponding next-state image, updating only the moved vehicle while preserving the board, exit, and other vehicles. The process continues until the red car reaches the exit and the model outputs the final vehicle-slide sequence. The prompt is detailed in Table~\ref{tab:parking-self-VCoT}.

\paragraph{Onet. }
In this setting, the model solves the Onet puzzle iteratively by predicting one removable fruit pair at each step. After each predicted removal, the model generates the corresponding next-state image, where only the selected pair is cleared and the remaining fruit icons stay unchanged. The process continues until the board is cleared and the model outputs the final pair-removal sequence. The prompt is detailed in Table~\ref{tab:onet-self-VCoT}.

\paragraph{Klotski. }
In this setting, the model solves the Klotski puzzle iteratively by predicting one block slide at each step. After each predicted move, the model generates the corresponding next-state image, updating only the moved block while preserving the board geometry and all other blocks. The process continues until the target block reaches the goal position and the model outputs the final solution sequence. The prompt is detailed in Table~\ref{tab:klotski-self-VCoT}.

\subsubsection{self-VR}

\paragraph{Maze. }
In this setting, the model is prompted to generate a visual reasoning image that overlays a continuous red path from the start marker to the goal while preserving the original maze. The prompt is detailed in Table~\ref{tab:maze-self-VR}.

\paragraph{Sokoban. }
In this setting, the model is prompted to generate a visual reasoning image that highlights a feasible player trajectory for pushing the box onto the target while preserving the original puzzle state. The prompt is detailed in Table~\ref{tab:sokoban-self-VR}.

\paragraph{Path. }
In this setting, the model is prompted to generate a visual reasoning image that overlays a continuous red path from the orange start triangle to the blue destination triangle while preserving the walls and labeled points. The prompt is detailed in Table~\ref{tab:path-self-VR}.

\paragraph{Parking. }
In this setting, the model is prompted to generate a visual reasoning image that adds concise visual hints for a feasible vehicle-moving strategy while preserving the original parking board and vehicle positions. The prompt is detailed in Table~\ref{tab:parking-self-VR}.

\paragraph{Onet. }
In this setting, the model is prompted to generate a visual reasoning image that highlights valid connection paths between removable matching fruit pairs while preserving the original Onet board. The prompt is detailed in Table~\ref{tab:onet-self-VR}.

\paragraph{Klotski. }
In this setting, the model is prompted to generate a visual reasoning image that adds visual hints for a feasible sequence of legal block slides while preserving the original Klotski board state. The prompt is detailed in Table~\ref{tab:klotski-self-VR}.

\subsubsection{oracle-GtA}

\paragraph{Maze. }
Compared with \textit{Self-GtA}, this setting does not ask the model to generate the auxiliary annotated image. Instead, we directly provide an auxiliary maze image by overlaying red grid-boundary lines on the $10 \times 10$ maze and placing coordinate labels on each cell.

\paragraph{Sokoban. }
Compared with \textit{Self-GtA}, this setting does not ask the model to generate the auxiliary annotated image. Instead, we directly provide an auxiliary Sokoban image by overlaying red grid-boundary lines on the board and placing coordinate labels on each cell.

\paragraph{Path. }
Compared with \textit{Self-GtA}, this setting does not ask the model to generate the auxiliary annotated image. Instead, we directly provide an auxiliary path image by overlaying red grid-boundary lines on the $10 \times 10$ map and placing coordinate labels on each cell.

\paragraph{Parking. }
Compared with \textit{Self-GtA}, this setting does not ask the model to generate the auxiliary annotated image. Instead, we directly provide an auxiliary parking image by overlaying red grid-boundary lines on the board and placing coordinate labels on each cell.

\paragraph{Onet. }
Compared with \textit{Self-GtA}, this setting does not ask the model to generate the auxiliary annotated image. Instead, we directly provide an auxiliary Onet image by overlaying red grid-boundary lines on the $6 \times 6$ board and placing coordinate labels on each cell.

\paragraph{Klotski. }
Compared with \textit{Self-GtA}, this setting does not ask the model to generate the auxiliary annotated image. Instead, we directly provide an auxiliary Klotski image by overlaying red dashed grid-boundary lines on the $6 \times 6$ board and placing coordinate labels on each cell.

\subsubsection{oracle-VCoT}

\paragraph{Maze. }
Compared with \textit{self-VCoT}, this setting does not ask the model to generate the next-state image after each predicted move. Instead, after the model predicts one move, we directly provide the updated maze state image by moving the agent marker according to that move while keeping the maze layout unchanged.

\paragraph{Sokoban. }
Compared with \textit{self-VCoT}, this setting does not ask the model to generate the next-state image after each predicted move. Instead, after the model predicts one player move, we directly provide the updated Sokoban state image by applying the move to the player and box positions while keeping the walls, floor, and target unchanged.

\paragraph{Path. }
Compared with \textit{self-VCoT}, this setting does not ask the model to generate the next-state image after each predicted move. Instead, after the model predicts one movement step, we directly provide the updated path state image by moving the current-position marker while keeping the map and labels unchanged.

\paragraph{Parking. }
Compared with \textit{self-VCoT}, this setting does not ask the model to generate the next-state image after each predicted move. Instead, after the model predicts one vehicle slide, we directly provide the updated parking state image by moving the selected vehicle while keeping the board, exit, and other vehicles unchanged.

\paragraph{Onet. }
Compared with \textit{self-VCoT}, this setting does not ask the model to generate the next-state image after each predicted pair removal. Instead, after the model predicts one removable pair, we directly provide the updated Onet state image by clearing the selected cells while keeping the remaining board unchanged.

\paragraph{Klotski. }
Compared with \textit{self-VCoT}, this setting does not ask the model to generate the next-state image after each predicted move. Instead, after the model predicts one block slide, we directly provide the updated Klotski state image by moving the selected block while keeping the board and other blocks unchanged.

\subsubsection{oracle-Aug-VCoT}

\paragraph{Maze. }
Compared with \textit{oracle-VCoT}, this setting additionally provides an auxiliary maze image before the iterative solving process. The auxiliary image is created by overlaying red grid-boundary lines on the $10 \times 10$ maze and placing coordinate labels on each cell. After each predicted move, the updated state image is also provided with the same grid and coordinate annotations.

\paragraph{Sokoban. }
Compared with \textit{oracle-VCoT}, this setting additionally provides an auxiliary Sokoban image before the iterative solving process. The auxiliary image is created by overlaying red grid-boundary lines on the board and placing coordinate labels on each cell. After each predicted move, the updated state image is also provided with the same grid and coordinate annotations.

\paragraph{Path. }
Compared with \textit{oracle-VCoT}, this setting additionally provides an auxiliary path image before the iterative solving process. The auxiliary image is created by overlaying red grid-boundary lines on the $10 \times 10$ map and placing coordinate labels on each cell. After each predicted move, the updated state image is also provided with the same grid and coordinate annotations.

\paragraph{Parking. }
Compared with \textit{oracle-VCoT}, this setting additionally provides an auxiliary parking image before the iterative solving process. The auxiliary image is created by overlaying red grid-boundary lines on the board and placing coordinate labels on each cell. After each predicted move, the updated state image is also provided with the same grid and coordinate annotations.

\paragraph{Onet. }
Compared with \textit{oracle-VCoT}, this setting additionally provides an auxiliary Onet image before the iterative solving process. The auxiliary image is created by overlaying red grid-boundary lines on the $6 \times 6$ board and placing coordinate labels on each cell. After each predicted pair removal, the updated state image is also provided with the same grid and coordinate annotations.

\paragraph{Klotski. }
Compared with \textit{oracle-VCoT}, this setting additionally provides an auxiliary Klotski image before the iterative solving process. The auxiliary image is created by overlaying red dashed grid-boundary lines on the $6 \times 6$ board and placing coordinate labels on each cell. After each predicted move, the updated state image is also provided with the same grid and coordinate annotations.

\subsection{Model Details}
\label{appendix:model_detail}

The evaluated models are categorized into three groups: 

1) \textbf{Open-source VLMs:} We include widely adopted open-weights models such as Nemotron-nano-12b-v2-vl~\cite{nvidia2025nvidianemotronnanov2}, Qwen2.5-VL-7B-Instruct~\cite{bai2025qwen25vltechnicalreport}, Qwen3-VL-8B-Instruct~\cite{bai2025qwen3vltechnicalreport}, and Qwen3.5-9B~\cite{qwen2026qwen35}. These models are specifically selected because their parameter scales are highly comparable to those of the majority of the evaluated open-source UMMs, serving as a scale-matched reference to investigate the behavioral differences between standard visual-language models and unified architectures.

2) \textbf{Closed-source VLMs:} We evaluate state-of-the-art (SOTA) proprietary models, including Qwen3.6-Plus~\cite{qwen2026qwen36}, Doubao-Seed-2.0-pro~\cite{seedteam2026seed20}, Gemini-3.1-Pro-Preview~\cite{geminiteam2026gemini31}, and GPT-5.4~\cite{openaiteam2026gpt54}. We include these models for two primary reasons: (1) to investigate whether the SOTA models can still achieve additional performance gains through generative visual augmentations; and (2) to establish an empirical upper bound for complex visual reasoning, serving as a critical reference point to guide the future evolution of UMMs. 

3) \textbf{Unified Multimodal Models (UMMs):} We evaluate recent UMMs, including MammothModa2~\cite{shen2025MammothModa2unifiedardiffusionframework}, STAR-7B~\cite{qin2025starstackedautoregressivescheme}, OmniGen2~\cite{Wu_2026_CVPR}, Janus-Pro~\cite{chen2025janusprounifiedmultimodalunderstanding}, BAGEL~\cite{deng2025emergingpropertiesunifiedmultimodal}, Show-o2~\cite{xie2025showo}, LongCat-Next~\cite{meituanlongcatteam2026longcatnextlexicalizingmodalitiesdiscrete}, LLaDA2.0-Uni~\cite{ai2026llada20uniunifyingmultimodalunderstanding}. Consistent with prior work, we treat Nano-Banana~\cite{fortin2025nanobanana} and Nano Banana 2~\cite{raisinghani2026nanobanana2} as a unified model for comparison.

\subsection{Evaluation Details}
\label{appendix:evaluation_detail}

We set \texttt{max\_new\_tokens} to $8192$ and use greedy decoding with temperature set to $0$ for all evaluated models.  For UMMs, we run each experiment independently three times and report the average performance. Our evaluation implementation is based on lmms-eval~\cite{zhang-etal-2025-lmms}.

Additionally, due to the current capability limitations of UMMs, they require manual iterative calls to achieve interleaved text-image generation. For models that support only single-image input, such as Show-o2 and LLaDA2.0-Uni, we use the output image from the previous step as the visual input for the next step. This evaluation setting is consistent with previous works~\cite{zou-etal-2026-uni, wen2026unig2ubenchunifiedmodelsadvance}. To assess the impact of the single-image-input, we evaluate Qwen3.5-9B under the same setting, where only the most recently generated image is provided at each answering step. As shown in Table~\ref{tab:last-image-input}, using only the last image does not substantially change Oracle-VCoT performance on these tasks.

\begin{table}[t]
\centering
\small
\begin{tabular}{lcc}
\hline
Task & Oracle-VCoT & Oracle-VCoT-last-image \\
\hline
Maze    & 25.0 & 20.0 \\
Sokoban & 20.0 & 20.0 \\
Path    & 25.0 & 20.0 \\
\hline
\end{tabular}
\caption{Oracle-VCoT accuracy with all images or only the last image on Qwen3.5-9B.}
\label{tab:last-image-input}
\end{table}

\begin{table*}[!t]
\small
\centering
\resizebox{0.9\textwidth}{!}{%
\begin{tabular}{ll*{4}{ccc}}
\toprule
\textbf{Model} & \textbf{Method}
& \multicolumn{3}{c}{\textbf{Maze}}
& \multicolumn{3}{c}{\textbf{Sokoban}}
& \multicolumn{3}{c}{\textbf{Parking}}
& \multicolumn{3}{c}{\textbf{Klotski}} \\
\cmidrule(lr){3-5}\cmidrule(lr){6-8}\cmidrule(lr){9-11}\cmidrule(lr){12-14}
&
& \textbf{Easy} & \textbf{Medium} & \textbf{Hard}
& \textbf{Easy} & \textbf{Medium} & \textbf{Hard}
& \textbf{Easy} & \textbf{Medium} & \textbf{Hard}
& \textbf{Easy} & \textbf{Medium} & \textbf{Hard} \\
\midrule
\multicolumn{14}{c}{\textbf{Open-source VLMs}} \\
\midrule
\multirow{3}{*}{Qwen2.5-VL-7B-Instruct}
& Direct & 10.0 & 0.0 & 0.0 & 5.0 & 0.0 & 0.0 & 0.0 & 0.0 & 0.0 & 0.0 & 0.0 & 0.0 \\
& Text-CoT & 0.0 & 0.0 & 0.0 & 0.0 & 0.0 & 0.0 & 15.0 & 0.0 & 0.0 & 0.0 & 0.0 & 0.0 \\
& Oracle-GtA & 0.0 & 0.0 & 0.0 & 0.0 & 0.0 & 0.0 & 5.0 & 0.0 & 0.0 & 0.0 & 0.0 & 0.0 \\
\midrule
\multirow{3}{*}{Nemotron-nano-12b-v2-vl}
& Direct & 5.0 & 0.0 & 0.0 & 0.0 & 0.0 & 0.0 & 0.0 & 0.0 & 0.0 & 0.0 & 0.0 & 0.0 \\
& Text-CoT & 5.0 & 0.0 & 0.0 & 10.0 & 0.0 & 0.0 & 15.0 & 0.0 & 0.0 & 0.0 & 0.0 & 0.0 \\
& Oracle-GtA & 35.0 & 0.0 & 0.0 & 0.0 & 0.0 & 0.0 & 0.0 & 0.0 & 0.0 & 0.0 & 0.0 & 0.0 \\
\midrule
\multirow{3}{*}{Qwen3-VL-8B-Instruct}
& Direct & 0.0 & 0.0 & 0.0 & 0.0 & 0.0 & 0.0 & 15.0 & 0.0 & 0.0 & 0.0 & 0.0 & 0.0 \\
& Text-CoT & 5.0 & 0.0 & 0.0 & 5.0 & 0.0 & 0.0 & 10.0 & 0.0 & 0.0 & 0.0 & 0.0 & 0.0 \\
& Oracle-GtA & 55.0 & 0.0 & 0.0 & 0.0 & 0.0 & 0.0 & 25.0 & 0.0 & 0.0 & 0.0 & 0.0 & 0.0 \\
\midrule
\multirow{3}{*}{Qwen3.5-9B}
& Direct & 0.0 & 0.0 & 0.0 & 0.0 & 0.0 & 0.0 & 15.0 & 0.0 & 0.0 & 0.0 & 0.0 & 0.0 \\
& Text-CoT & 30.0 & 0.0 & 0.0 & 30.0 & 0.0 & 0.0 & 25.0 & 0.0 & 0.0 & 0.0 & 0.0 & 0.0 \\
& Oracle-GtA & 95.0 & 33.3 & 6.0 & 55.0 & 10.0 & 6.6 & 80.0 & 6.6 & 0.0 & 65.0 & 0.0 & 0.0 \\
\midrule
\multicolumn{14}{c}{\textbf{Closed-source VLMs}} \\
\midrule
\multirow{2}{*}{\shortstack[l]{Qwen3.6-Plus}}
& Text-CoT & 70.0 & 3.3 & 0.0 & 30.0 & 3.3 & 2.0 & 95.0 & 73.3 & 14.0 & 60.0 & 0.0 & 0.0 \\
& Oracle-GtA & 100.0 & 100.0 & 98.0 & 100.0 & 86.6 & 58.0 & 100.0 & 90.0 & 18.0 & 100.0 & 3.3 & 0.0 \\
\midrule
\multirow{2}{*}{\shortstack[l]{Doubao-Seed-2.0-pro}}
& Text-CoT & 80.0 & 43.3 & 28.0 & 55.0 & 20.0 & 6.0 & 35.0 & 10.0 & 0.0 & 45.0 & 10.0 & 0.0 \\
& Oracle-GtA & 100.0 & 93.3 & 92.0 & 100.0 & 100.0 & 52.0 & 100.0 & 80.0 & 18.0 & 90.0 & 36.6 & 0.0 \\
\midrule
\multirow{2}{*}{\makecell[l]{Gemini-3.1-Pro-Preview}}
& Text-CoT & 75.0 & 20.0 & 24.0 & 60.0 & 23.3 & 2.0 & 30.0 & 6.6 & 0.0 & 55.0 & 3.3 & 0.0 \\
& Oracle-GtA & 100.0 & 93.3 & 82.0 & 85.0 & 70.0 & 42.0 & 85.0 & 26.6 & 0.0 & 90.0 & 13.3 & 0.0 \\
\midrule
\multirow{2}{*}{\shortstack[l]{GPT-5.4}}
& Text-CoT & 100.0 & 83.3 & 74.0 & 40.0 & 10.0 & 4.0 & 85.0 & 90.0 & 20.0 & 100.0 & 46.6 & 2.0 \\
& Oracle-GtA & 100.0 & 76.6 & 72.0 & 90.0 & 70.0 & 60.0 & 95.0 & 76.6 & 26.0 & 95.0 & 60.0 & 2.0 \\
\midrule
\multicolumn{14}{c}{\textbf{Unified Models}} \\
\midrule
\multirow{3}{*}{MammothModa2}
& Direct & 0.0 & 0.0 & 0.0 & 0.0 & 0.0 & 0.0 & 0.0 & 0.0 & 0.0 & 0.0 & 0.0 & 0.0 \\
& Text-CoT & 0.0 & 0.0 & 0.0 & 0.0 & 0.0 & 0.0 & 0.0 & 0.0 & 0.0 & 0.0 & 0.0 & 0.0 \\ 
& Oracle-GtA & 0.0 & 0.0 & 0.0 & 0.0 & 0.0 & 0.0 & 0.0 & 0.0 & 0.0 & 0.0 & 0.0 & 0.0 \\
\midrule
\multirow{3}{*}{STAR-7B}
& Direct & 0.0 & 0.0 & 0.0 & 0.0 & 0.0 & 0.0 & 10.0 & 0.0 & 0.0 & 0.0 & 0.0 & 0.0 \\
& Text-CoT & 0.0 & 0.0 & 0.0 & 0.0 & 0.0 & 0.0 & 10.0 & 0.0 & 0.0 & 0.0 & 0.0 & 0.0 \\
& Oracle-GtA & 10.0 & 0.0 & 0.0 & 0.0 & 0.0 & 0.0 & 10.0 & 0.0 & 0.0 & 0.0 & 0.0 & 0.0 \\
\midrule
\multirow{3}{*}{OmniGen2}
& Direct & 0.0 & 0.0 & 0.0 & 0.0 & 0.0 & 0.0 & 0.0 & 0.0 & 0.0 & 0.0 & 0.0 & 0.0 \\
& Text-CoT & 0.0 & 0.0 & 0.0 & 0.0 & 0.0 & 0.0 & 0.0 & 0.0 & 0.0 & 0.0 & 0.0 & 0.0 \\
& Oracle-GtA & 0.0 & 0.0 & 0.0 & 0.0 & 0.0 & 0.0 & 0.0 & 0.0 & 0.0 & 0.0 & 0.0 & 0.0 \\
\midrule
\multirow{3}{*}{Janus-Pro}
& Direct & 0.0 & 0.0 & 0.0 & 0.0 & 0.0 & 0.0 & 5.0 & 0.0 & 0.0 & 0.0 & 0.0 & 0.0 \\
& Text-CoT & 0.0 & 0.0 & 0.0 & 0.0 & 0.0 & 0.0 & 15.0 & 0.0 & 0.0 & 0.0 & 0.0 & 0.0 \\
& Oracle-GtA & 5.0 & 0.0 & 0.0 & 5.0 & 0.0 & 0.0 & 25.0 & 0.0 & 0.0 & 0.0 & 0.0 & 0.0 \\
\midrule
\multirow{3}{*}{LLaDA2.0-Uni}
& Direct & 0.0 & 0.0 & 0.0 & 0.0 & 0.0 & 0.0 & 5.0 & 0.0 & 0.0 & 0.0 & 0.0 & 0.0 \\
& Text-CoT & 5.0 & 0.0 & 0.0 & 0.0 & 0.0 & 0.0 & 5.0 & 0.0 & 0.0 & 0.0 & 0.0 & 0.0 \\
& Oracle-GtA & 0.0 & 0.0 & 0.0 & 5.0 & 0.0 & 0.0 & 10.0 & 0.0 & 0.0 & 0.0 & 0.0 & 0.0 \\
\midrule
\multirow{3}{*}{Show-o2}
& Direct & 0.0 & 0.0 & 0.0 & 0.0 & 0.0 & 0.0 & 0.0 & 0.0 & 0.0 & 0.0 & 0.0 & 0.0 \\
& Text-CoT & 5.0 & 0.0 & 0.0 & 0.0 & 0.0 & 0.0 & 0.0 & 0.0 & 0.0 & 0.0 & 0.0 & 0.0 \\
& Oracle-GtA & 15.0 & 0.0 & 0.0 & 0.0 & 0.0 & 0.0 & 10.0 & 0.0 & 0.0 & 0.0 & 0.0 & 0.0 \\
\midrule
\multirow{3}{*}{BAGEL}
& Direct & 0.0 & 0.0 & 0.0 & 0.0 & 0.0 & 0.0 & 5.0 & 0.0 & 0.0 & 0.0 & 0.0 & 0.0 \\
& Text-CoT & 5.0 & 0.0 & 0.0 & 0.0 & 0.0 & 0.0 & 10.0 & 0.0 & 0.0 & 0.0 & 0.0 & 0.0 \\
& Oracle-GtA & 15.0 & 0.0 & 0.0 & 0.0 & 0.0 & 0.0 & 15.0 & 0.0 & 0.0 & 0.0 & 0.0 & 0.0 \\
\midrule
\multirow{3}{*}{LongCat-Next}
& Direct & 0.0 & 0.0 & 0.0 & 0.0 & 0.0 & 0.0 & 0.0 & 0.0 & 0.0 & 0.0 & 0.0 & 0.0 \\
& Text-CoT & 10.0 & 0.0 & 0.0 & 0.0 & 0.0 & 0.0 & 15.0 & 0.0 & 0.0 & 0.0 & 0.0 & 0.0 \\
& Oracle-GtA & 60.0 & 13.3 & 0.0 & 15.0 & 0.0 & 0.0 & 10.0 & 0.0 & 0.0 & 30.0 & 0.0 & 0.0 \\
\midrule
\multirow{3}{*}{Nano-Banana}
& Direct & 20.0 & 0.0 & 0.0 & 15.0 & 0.0 & 0.0 & 0.0 & 0.0 & 0.0 & 0.0 & 0.0 & 0.0 \\
& Text-CoT & 40.0 & 3.3 & 0.0 & 20.0 & 0.0 & 0.0 & 15.0 & 3.3 & 0.0 & 10.0 & 0.0 & 0.0 \\
& Oracle-GtA & 85.0 & 40.0 & 38.0 & 30.0 & 0.0 & 0.0 & 90.0 & 43.3 & 4.0 & 60.0 & 16.6 & 0.0 \\
\midrule
\multirow{3}{*}{Nano Banana 2}
& Direct & 20.0 & 0.0 & 0.0 & 10.0 & 0.0 & 0.0 & 5.0 & 0.0 & 0.0 & 0.0 & 0.0 & 0.0 \\
& Text-CoT & 35.0 & 0.0 & 0.0 & 20.0 & 0.0 & 0.0 & 15.0 & 3.3 & 0.0 & 10.0 & 0.0 & 0.0 \\
& Oracle-GtA & 85.0 & 40.0 & 38.0 & 30.0 & 0.0 & 0.0 & 70.0 & 10.0 & 0.0 & 60.0 & 13.3 & 0.0 \\
\bottomrule
\end{tabular}%
}
\caption{Feasible-solution accuracy (\%) on Maze, Sokoban, Parking, and Klotski under oracle settings.}
\label{tab:additional-reach-any}
\end{table*}

\begin{table*}[!t]
\small
\centering
\resizebox{0.72\textwidth}{!}{%
\begin{tabular}{ll*{4}{c}}
\toprule
\textbf{Model} & \textbf{Setting} & \textbf{Maze} & \textbf{Sokoban} & \textbf{Parking} & \textbf{Klotski} \\
\midrule
\multirow{2}{*}{MammothModa2}
& Self-GtA   & 0.0 & 0.0 & 0.0 & 0.0 \\
& Self-VR    & 0.0 & 0.0 & 0.0 & 0.0 \\
\midrule
\multirow{2}{*}{STAR-7B}
& Self-GtA   & 0.0 & 0.0 & 0.0 & 0.0 \\
& Self-VR    & 0.0 & 0.0 & 0.0 & 0.0 \\
\midrule
\multirow{2}{*}{OmniGen2}
& Self-GtA   & 0.0 & 0.0 & 0.0 & 0.0 \\
& Self-VR    & 0.0 & 0.0 & 10.0 & 0.0 \\
\midrule
\multirow{2}{*}{Janus-Pro}
& Self-GtA   & 5.0 & 5.0 & 25.0 & 0.0 \\
& Self-VR    & 0.0 & 0.0 & 20.0 & 0.0 \\
\midrule
\multirow{2}{*}{LLaDA2.0-Uni}
& Self-GtA   & 0.0 & 0.0 & 0.0 & 0.0 \\
& Self-VR    & 0.0 & 5.0 & 0.0 & 0.0 \\
\midrule
\multirow{2}{*}{Show-o2}
& Self-GtA   & 10.0 & 0.0 & 10.0 & 0.0 \\
& Self-VR    & 0.0 & 0.0 & 10.0 & 0.0 \\
\midrule
\multirow{2}{*}{BAGEL}
& Self-GtA   & 0.0 & 0.0 & 5.0 & 0.0 \\
& Self-VR    & 0.0 & 0.0 & 0.0 & 0.0 \\
\midrule
\multirow{2}{*}{LongCat-Next}
& Self-GtA   & 20.0 & 5.0 & 10.0 & 0.0 \\
& Self-VR    & 5.0 & 0.0 & 5.0 & 0.0 \\
\midrule
\multirow{2}{*}{Nano-Banana}
& Self-GtA   & 15.0 & 20.0 & 15.0 & 10.0 \\
& Self-VR    & 5.0 & 10.0 & 10.0 & 0.0 \\
\midrule
\multirow{2}{*}{Nano Banana 2}
& Self-GtA   & 55.0 & 25.0 & 40.0 & 30.0 \\
& Self-VR    & 15.0 & 0.0 & 10.0 & 10.0 \\
\bottomrule
\end{tabular}%
}
\caption{Feasible-solution accuracy (\%) of unified models on Maze, Sokoban, Parking, and Klotski under two self-generation evaluation settings.}
\label{tab:unified_models_reach_any}
\end{table*}

\begin{table*}[!t]
\centering
\caption{Results on Sokoban.}
\label{tab:qwen_reasoners_sokoban}
\resizebox{\textwidth}{!}{
\begin{tabular}{lcccccc}
\toprule
Method 
& Qwen3.5-0.8B 
& Qwen3.5-2B  
& Qwen3.5-4B 
& Qwen3.5-9B 
& Qwen3.5-27B 
& Qwen3.6-plus \\
\midrule
Text-CoT & 0.0 & 10.0 & 25.0 & 30.0 & 15.0 & 30.0 \\
\midrule
\multicolumn{7}{c}{\textit{Vision: Gaussian noise}} \\
\midrule
Self-GtA & 0.0 \textcolor{gray}{(+0.0)} & \textbf{8.3} \textcolor{blue}{(-1.7)}  & 11.6 \textcolor{blue}{(-13.4)} & \textbf{21.6} \textcolor{blue}{(-8.4)}  & 20.0 \textcolor{red}{(+5.0)}  & 23.3 \textcolor{blue}{(-6.7)} \\
Self-VCoT & \cellcolor{gray!15}\textbf{6.6} \textcolor{red}{(+6.6)} & \cellcolor{gray!15}\textbf{8.3} \textcolor{blue}{(-1.7)} & \cellcolor{gray!15}\textbf{15.0} \textcolor{blue}{(-10.0)} & 15.0 \textcolor{blue}{(-15.0)} & \textbf{21.6} \textcolor{red}{(+6.6)} & \textbf{41.6} \textcolor{red}{(+11.6)} \\
Self-VR & 0.0 \textcolor{gray}{(+0.0)} & 3.3 \textcolor{blue}{(-6.7)} & 11.6 \textcolor{blue}{(-13.4)} & 16.6 \textcolor{blue}{(-13.4)} & 18.3 \textcolor{red}{(+3.3)} & 21.6 \textcolor{blue}{(-8.4)} \\
\midrule
\multicolumn{7}{c}{\textit{Vision: Mismatch Image}} \\
\midrule
Self-GtA & 0.0 \textcolor{gray}{(+0.0)} & 6.6 \textcolor{blue}{(-3.4)}  & 10.0 \textcolor{blue}{(-15.0)} & 20.0 \textcolor{blue}{(-10.0)}  & 31.6 \textcolor{red}{(+16.6)}  & 11.6 \textcolor{blue}{(-18.4)} \\
Self-VCoT & \cellcolor{gray!15}\textbf{11.6} \textcolor{red}{(+11.6)} & \cellcolor{gray!15}\textbf{6.6} \textcolor{blue}{(-3.4)} & \cellcolor{gray!15}\textbf{10.0} \textcolor{blue}{(-15.0)} & \cellcolor{gray!15}\textbf{13.3} \textcolor{blue}{(-16.7)} & \cellcolor{gray!15}\textbf{33.3} \textcolor{red}{(+18.3)} & \textbf{26.6} \cellcolor{gray!15}\textcolor{blue}{(-3.4)} \\
Self-VR & 0.0 \textcolor{gray}{(+0.0)} & 5.0 \textcolor{blue}{(-5.0)} & 10.0 \textcolor{blue}{(-15.0)} & 10.0 \textcolor{blue}{(-20.0)} & 15.0 \textcolor{gray}{(+0.0)} & 18.3 \textcolor{blue}{(-11.7)} \\
\midrule
\multicolumn{7}{c}{\textit{Vision generator:  Qwen-Image}} \\
\midrule
Self-GtA & 0.0 \textcolor{gray}{(+0.0)} & 0.0 \textcolor{blue}{(-10.0)} & 8.3 \textcolor{blue}{(-16.7)} & 20.0 \textcolor{blue}{(-10.0)} & 23.3 \textcolor{red}{(+8.3)} & 20.0 \textcolor{blue}{(-10.0)} \\
Self-VCoT & \cellcolor{gray!15}\textbf{6.6 \textcolor{red}{(+6.6)}} & \cellcolor{gray!15}\textbf{6.6 \textcolor{blue}{(-3.4)}} & \cellcolor{gray!15}\textbf{10.0 \textcolor{blue}{(-15.0)}} & \cellcolor{gray!15}\textbf{11.6 \textcolor{blue}{(-18.4)}} & \cellcolor{gray!15}\textbf{23.3 \textcolor{red}{(+8.3)}} & \cellcolor{gray!15}\textbf{45.0 \textcolor{red}{(+15.0)}} \\
Self-VR & 0.0 \textcolor{gray}{(+0.0)} & 0.0 \textcolor{blue}{(-10.0)} & 6.6 \textcolor{blue}{(-18.4)} & 8.3 \textcolor{blue}{(-21.7)} & 15.0 \textcolor{gray}{(+0.0)} & 16.6 \textcolor{blue}{(-13.4)} \\
\midrule
\multicolumn{7}{c}{\textit{Vision generator: Qwen-Image-Edit}} \\
\midrule
Self-GtA & 0.0  \textcolor{gray}{(+0.0)} & 1.6 \textcolor{blue}{(-8.4)} & \textbf{20.0} \textcolor{blue}{(-5.0)} & \textbf{38.3} \textcolor{red}{(+8.3)} & \textbf{40.0} \textcolor{red}{(+25.0)} & \textbf{78.3} \textcolor{red}{(+48.3)} \\
Self-VCoT & \cellcolor{gray!15} \textbf{3.3} \textcolor{red}{(+3.3)} & \cellcolor{gray!15}\textbf{8.3} \textcolor{blue}{(-1.7)} & 8.3 \textcolor{blue}{(-16.7)} & 8.3 \textcolor{blue}{(-21.7)} & 25.0 \textcolor{red}{(+10.0)} & 31.6 \textcolor{red}{(+1.6)} \\
Self-VR & 0.0 \textcolor{gray}{(+0.0)} & 1.6 \textcolor{blue}{(-8.4)} & 6.6 \textcolor{blue}{(-18.4)} & 6.6 \textcolor{blue}{(-23.4)} & 10.0 \textcolor{blue}{(-5.0)} & 23.3 \textcolor{blue}{(-6.7)} \\
\midrule
\multicolumn{7}{c}{\textit{Vision generator: Nano Banana 2}} \\
\midrule
Self-GtA & 0.0 \textcolor{gray}{(+0.0)} & \textbf{1.6} \textcolor{blue}{(-8.4)} & \textbf{28.3} \textcolor{red}{(+3.3)} & \textbf{50.0} \textcolor{red}{(+20.0)} & \textbf{48.3} \textcolor{red}{(+33.3)} & \textbf{83.3} \textcolor{red}{(+53.3)} \\
Self-VCoT & \cellcolor{gray!15}\textbf{8.3} \textcolor{red}{(+8.3)} & 10.0 \textcolor{gray}{(+0.0)} & 15.0 \textcolor{blue}{(-10.0)} & 18.3 \textcolor{blue}{(-11.7)} & 33.3 \textcolor{red}{(+18.3)} & \textcolor{red}{36.6} \textcolor{red}{(+6.6)} \\
Self-VR & 0.0 \textcolor{gray}{(+0.0)} & 6.6 \textcolor{blue}{(-3.4)} & 6.6 \textcolor{blue}{(-18.4)} & 16.6 \textcolor{blue}{(-13.4)} & 15.0 \textcolor{gray}{(+0.0)} & 21.6 \textcolor{blue}{(-8.4)} \\
\midrule
\multicolumn{7}{c}{\textit{Oracle generation}} \\
\midrule
Oracle-GtA & 0.0 \textcolor{gray}{(+0.0)} & \textbf{10.0} \textcolor{gray}{(+0.0)} & \textbf{55.0} \textcolor{red}{(+30.0)} & \textbf{55.0} \textcolor{red}{(+25.0)} & \textbf{75.0} \textcolor{red}{(+60.0)} & \textbf{100.0} \textcolor{red}{(+70.0)} \\
Oracle-VCoT & \cellcolor{gray!15}\textbf{10.0} \textcolor{red}{(+10.0)}&  5.0 \textcolor{blue}{(-5.0)}& 20.0 \textcolor{blue}{(-5.0)} & 20.0 \textcolor{blue}{(-10.0)} & 15.0 \textcolor{gray}{(+0.0)} & 50.0 \textcolor{red}{(+20.0)} \\
\bottomrule
\end{tabular}
}
\end{table*}

\begin{table*}[!t]
\centering
\caption{Results on Parking.}
\label{tab:qwen_reasoners_parking}
\resizebox{\textwidth}{!}{
\begin{tabular}{lcccccc}
\toprule
Method 
& Qwen3.5-0.8B 
& Qwen3.5-2B  
& Qwen3.5-4B 
& Qwen3.5-9B 
& Qwen3.5-27B 
& Qwen3.6-plus \\
\midrule
Text-CoT & 5.0 & 5.0 & 25.0 & 25.0 & 40.0 & 70.0 \\
\midrule
\multicolumn{7}{c}{\textit{Vision: Gaussian noise}} \\
\midrule
Self-GtA & \textbf{3.3} \textcolor{blue}{(-1.7)} & 3.3 \textcolor{blue}{(-1.7)} & \textbf{20.0} \textcolor{blue}{(-5.0)} & \textbf{33.3} \textcolor{red}{(+8.3)} & 6.6 \textcolor{blue}{(-33.4)} & \textbf{88.3} \textcolor{red}{(+18.3)} \\
Self-VCoT & \cellcolor{gray!15}\textbf{3.3} \textcolor{blue}{(-1.7)} & \cellcolor{gray!15}\textbf{26.6} \textcolor{red}{(+21.6)} & 8.3 \textcolor{blue}{(-16.7)} & 23.3 \textcolor{blue}{(-1.7)} & 10.0 \textcolor{blue}{(-30.0)} & 75.0 \textcolor{red}{(+5.0)} \\
Self-VR & 0.0 \textcolor{blue}{(-5.0)} & 3.3 \textcolor{blue}{(-1.7)} & 10.0 \textcolor{blue}{(-15.0)} & 16.6 \textcolor{blue}{(-8.4)} & \textbf{30.0} \textcolor{blue}{(-10.0)} & 56.6 \textcolor{blue}{(-13.4)} \\
\midrule
\multicolumn{7}{c}{\textit{Vision: Mismatch Image}} \\
\midrule
Self-GtA & 0.0 \textcolor{blue}{(-5.0)} & 5.0 \textcolor{gray}{(+0.0)} & \textbf{20.0} \textcolor{blue}{(-5.0)} & \textbf{41.6} \textcolor{red}{(+16.6)} & 0.0 \textcolor{blue}{(-40.0)} & 30.0 \textcolor{blue}{(-40.0)} \\
Self-VCoT & \cellcolor{gray!15}\textbf{3.3} \textcolor{blue}{(-1.7)} & \cellcolor{gray!15}\textbf{28.3} \textcolor{red}{(+23.3)} & \cellcolor{gray!15}8.3 \textcolor{blue}{(-16.7)} & \cellcolor{gray!15}6.6 \textcolor{blue}{(-18.4)} & \cellcolor{gray!15}\textbf{10.0} \textcolor{blue}{(-30.0)} & 23.3 \textcolor{blue}{(-46.7)} \\
Self-VR & \textbf{3.3} \textcolor{blue}{(-1.7)} & 1.6 \textcolor{blue}{(-3.4)} & 5.0 \textcolor{blue}{(-20.0)} & 8.3 \textcolor{blue}{(-16.7)} & 3.3 \textcolor{blue}{(-36.7)} & \textbf{51.6} \textcolor{blue}{(-18.4)} \\
\midrule
\multicolumn{7}{c}{\textit{Vision generator:  Qwen-Image}} \\
\midrule
Self-GtA & \textbf{1.6} \textcolor{blue}{(-3.4)} & 10.0 \textcolor{red}{(+5.0)} & \textbf{15.0} \textcolor{blue}{(-10.0)} & 21.6 \textcolor{blue}{(-3.4)} & \textbf{16.6} \textcolor{blue}{(-23.4)} & 63.3 \textcolor{blue}{(-6.7)} \\
Self-VCoT & \cellcolor{gray!15}\textbf{1.6} \textcolor{blue}{(-3.4)} & \cellcolor{gray!15}\textbf{25.0} \textcolor{red}{(+20.0)} & \cellcolor{gray!15}10.0 \textcolor{blue}{(-15.0)} & \cellcolor{gray!15}\textbf{23.3} \textcolor{blue}{(-1.7)} & \cellcolor{gray!15}10.0 \textcolor{blue}{(-30.0)} & \cellcolor{gray!15}\textbf{86.6} \textcolor{red}{(+16.6)} \\
Self-VR & 0.0 \textcolor{blue}{(-5.0)} & 1.6 \textcolor{blue}{(-3.4)} & 3.3 \textcolor{blue}{(-21.7)} & 5.0 \textcolor{blue}{(-20.0)} & 3.3 \textcolor{blue}{(-36.7)} & 43.3 \textcolor{blue}{(-26.7)} \\
\midrule
\multicolumn{7}{c}{\textit{Vision generator: Qwen-Image-Edit}} \\
\midrule
Self-GtA & 1.6 \textcolor{blue}{(-3.4)} & 8.3 \textcolor{red}{(+3.3)} & \textbf{51.6} \textcolor{red}{(+26.6)} & \textbf{51.6} \textcolor{red}{(+26.6)} & \textbf{71.6} \textcolor{red}{(+31.6)} & \textbf{90.0} \textcolor{red}{(+20.0)} \\
Self-VCoT & \cellcolor{gray!15}0.0 \textcolor{blue}{(-5.0)} & \cellcolor{gray!15}\textbf{35.0} \textcolor{red}{(+30.0)} & 5.0 \textcolor{blue}{(-20.0)} & 23.3 \textcolor{blue}{(-1.7)} & 10.0 \textcolor{blue}{(-30.0)} & 73.3 \textcolor{red}{(+3.3)} \\
Self-VR & \textbf{5.0} \textcolor{gray}{(+0.0)} & 3.3 \textcolor{blue}{(-1.7)} & 5.0 \textcolor{blue}{(-20.0)} & 3.3 \textcolor{blue}{(-21.7)} & 13.3 \textcolor{blue}{(-26.7)} & 38.3 \textcolor{blue}{(-31.7)} \\
\midrule
\multicolumn{7}{c}{\textit{Vision generator: Nano Banana 2}} \\
\midrule
Self-GtA & \textbf{8.3} \textcolor{red}{(+3.3)} & 16.6 \textcolor{red}{(+11.6)} & \textbf{50.0} \textcolor{red}{(+25.0)} & \textbf{80.0} \textcolor{red}{(+55.0)} & \textbf{93.3} \textcolor{red}{(+53.3)} & \textbf{100.0} \textcolor{red}{(+30.0)} \\
Self-VCoT & \cellcolor{gray!15}0.0 \textcolor{blue}{(-5.0)} &  \cellcolor{gray!15}\textbf{31.6} \textcolor{red}{(+26.6)} & 10.0 \textcolor{blue}{(-15.0)} & 15.0 \textcolor{blue}{(-10.0)} & 10.0 \textcolor{blue}{(-30.0)} & 81.6 \textcolor{red}{(+11.6)} \\
Self-VR & 1.6 \textcolor{blue}{(-3.4)} & 3.3 \textcolor{blue}{(-1.7)} & 5.0 \textcolor{blue}{(-20.0)} & 16.6 \textcolor{blue}{(-8.4)} & 20.0 \textcolor{blue}{(-20.0)} & 53.3 \textcolor{blue}{(-16.7)} \\
\midrule
\multicolumn{7}{c}{\textit{Oracle generation}} \\
\midrule
Oracle-GtA & \textbf{20.0} \textcolor{red}{(+15.0)} & 25.0 \textcolor{red}{(+20.0)} & \textbf{90.0} \textcolor{red}{(+65.0)} & \textbf{80.0} \textcolor{red}{(+55.0)} & \textbf{100.0} \textcolor{red}{(+60.0)} & \textbf{100.0} \textcolor{red}{(+30.0)} \\
Oracle-VCoT & 0.0 \textcolor{blue}{(-5.0)} & \cellcolor{gray!15}\textbf{55.0} \textcolor{red}{(+50.0)} & 5.0 \textcolor{blue}{(-20.0)} & 60.0 \textcolor{red}{(+35.0)} & 5.0 \textcolor{blue}{(-35.0)} & 75.0 \textcolor{red}{(+5.0)} \\
\bottomrule
\end{tabular}
}
\end{table*}

\clearpage

\begin{table*}[h!]
\centering
\begin{tcolorbox}[colback=white, colframe=black!50, sharp corners=south, boxrule=0.4pt, width=0.95\linewidth]
You are a precise maze solver.\\
\\
SEMANTICS

- Black squares: walls (impassable)

- White squares: path (walkable)

- Blue dot: start (the agent)

- The green square marks the goal cell. It is passable, not a wall, and success is achieved only when the agent moves onto the cell occupied by the green square.

- Legal moves: up, down, left, right only. One cell per step; no diagonals, no jumps; never cross walls.\\

OUTPUT FORMAT

1) Describe your reasoning briefly: identify the start (agent) and goal, explain the planned move sequence, and confirm never enters black-wall cells.

2) Output exactly one final move list as a JSON array of lowercase strings, wrapped as:
\texttt{\detokenize{<ANSWER_JSON>["move1","move2",...]</ANSWER_JSON>}}

\end{tcolorbox}
\caption{prompt for maze under Text-CoT setting.}
\label{tab:maze-Text-CoT}
\end{table*}

\begin{table*}[h!]
\centering
\begin{tcolorbox}[colback=white, colframe=black!50, sharp corners=south, boxrule=0.4pt, width=0.95\linewidth]
You are solving a single-box Sokoban puzzle.\\

Legend:

- Brick wall: blocked

- Sand floor: walkable

- Small person: player start

- Wooden crate with X: box

- Green X: target\\

Rules:

1) Allowed moves: up, down, left, right (one cell per step).

2) If the player moves into the box, the box is pushed by one cell in the same direction.

3) A push is valid only if the box destination is not a wall.

4) No diagonal movement.\\

Goal:

Push the box onto the target cell.\\

OUTPUT FORMAT

1) Think briefly about legality of moves and box position updates.

2) Then output strictly:

\texttt{\detokenize{<ANSWER_JSON>["move1","move2",...]</ANSWER_JSON>}}

\end{tcolorbox}
\caption{prompt for sokoban under Text-CoT setting.}
\label{tab:sokoban-Text-CoT}
\end{table*}

\begin{table*}[h!]
\centering
\begin{tcolorbox}[colback=white, colframe=black!50, sharp corners=south, boxrule=0.4pt, width=0.95\linewidth]
In the image there are five labeled points: A, B, C, D, E.

From the orange triangle (start) to the blue triangle (destination), determine which labeled points lie on valid simple path without entering black grids.\\

SEMANTICS

- Black squares: walls (impassable)

- Light-gray cells are walkable.

- Orange triangle: path start.

- Blue triangle: destination.

- A/B/C/D/E: a label point, which is walkable.

- Legal moves: up, down, left, right only. One cell per step; no diagonals, no jumps; never cross walls.

- Follow the valid simple path from start to destination: no backtracking and no revisits.\\

Task:

From the orange triangle (start) to the blue triangle (destination), determine which labeled points lie on the valid simple path\\

OUTPUT FORMAT

1) Describe your reasoning briefly: identify the start and destination, explain the planned move sequence, and confirm never enters black-wall cells. 

2) Answer with only the passed letters in path order, using this format:

\texttt{\detokenize{<answer>point1,point2,...</answer>}}
\end{tcolorbox}
\caption{prompt for path under Text-CoT setting.}
\label{tab:path-Text-CoT}
\end{table*}

\begin{table*}[h!]
\centering
\begin{tcolorbox}[colback=white, colframe=black!50, sharp corners=south, boxrule=0.4pt, width=0.95\linewidth]
You are solving a parking-exit puzzle on a \{grid\_size\}$\times$ \{grid\_size\} board.\\

Rules:

1) Vehicle orientations are fixed (horizontal or vertical).

2) In one move, exactly one vehicle slides by one cell along its orientation.

3) No overlap and no leaving the board.

4) Success is achieved when the rightmost cell of the red car "R" reaches the right boundary on row index \{exit\_row - 1\} (indices are 0-based).\\

OUTPUT FORMAT

1) Describe your reasoning briefly.

2) Answer using this format:

\texttt{\detokenize{<ANSWER_JSON>[{"car":"A","direction":"up"}, ...]</ANSWER_JSON>}}

\end{tcolorbox}
\caption{prompt for parking under Text-CoT setting.}
\label{tab:parking-Text-CoT}
\end{table*}

\begin{table*}[h!]
\centering
\begin{tcolorbox}[colback=white, colframe=black!50, sharp corners=south, boxrule=0.4pt, width=0.95\linewidth]
You are solving an Onet puzzle on a 6x6 board.\\

Coordinate system

- Use 0-based coordinates.

- (r, c) means row r, column c.

- (0,0) is the top-left cell.

- r increases downward, c increases to the right.\\

Goal:

- Decide whether the whole board can be fully cleared.

- In each step, you may remove a pair of identical fruits if they can be linked.

- A valid link uses only horizontal/vertical segments and can have at most 2 turns.

- The path must not pass through any other fruit.

OUTPUT FORMAT:\\

1) Brief reasoning.

2) Final answer on a separate line:

\texttt{\detokenize{<answer_json>[[[r1,c1],[r2,c2]],[[r3,c3],[r4,c4]],...]</answer_json>}}

\end{tcolorbox}
\caption{prompt for onet under Text-CoT setting.}
\label{tab:onet-Text-CoT}
\end{table*}

\begin{table*}[h!]
\centering
\begin{tcolorbox}[colback=white, colframe=black!50, sharp corners=south, boxrule=0.4pt, width=0.95\linewidth]
You are an expert Klotski puzzle solver.\\

Coordinate system

- Use 0-based coordinates.

- (r, c) means row r, column c.

- (0,0) is the top-left cell.

- r increases downward, c increases to the right.\\

Board

- Grid size: width=6, height=6.

- There are exactly two empty cells at any time.

- Every non-empty cell belongs to exactly one block from the listed block shapes.\\

Block shapes

\quad - One 2x2 target block.

\quad - Multiple 1x2 vertical blocks.

\quad - Multiple 2x1 horizontal blocks.

\quad - Multiple 1x1 single blocks.\\

Block reference in actions

- In the action output, a block is identified by its CURRENT TOP-LEFT coordinate at the time of that move.\\

Legal move definition

1. Each step slides ONE block by exactly ONE cell in {UP, DOWN, LEFT, RIGHT}.

2. No rotation, no diagonal move, no jump.

3. A move is legal only if EVERY destination cell of that block is currently empty.

  \quad - Let current\_cells be the cells currently occupied by the moving block.
   
\quad   - Let destination\_cells be the cells occupied after translating the block by one cell in the chosen direction.
   
 \quad  - A move is legal iff every cell in (destination\_cells - current\_cells) is empty before the move.
   
 \quad  - For 1x2 / 2x1 blocks, required empty destination cells can be 1 or 2 depending on move direction.
   
 \quad  - For 2x2 blocks, required empty destination cells are 2.
   
4. Cells occupied by OTHER blocks are never passable.\\

Goal

- Make the 2x2 target block's TOP-LEFT coordinate reach (4,2).\\

OUTPUT FORMAT

Before outputting the final sequence, briefly describe your reasoning for the solution.

Finally, output the solution in the following format:

\{

 \quad "solution": [
  
 \quad\quad   {"from": [<r1>, <c1>], "direction": "<DIRECTION>"},
    
 \quad\quad   {"from": [<r2>, <c2>], "direction": "<DIRECTION>"}
    
 \quad ]
  
\}
\end{tcolorbox}
\caption{prompt for klotski under Text-CoT setting.}
\label{tab:klotski-Text-CoT}
\end{table*}

\begin{table*}[h!]
\centering
\begin{tcolorbox}[colback=white, colframe=black!50, sharp corners=south, boxrule=0.4pt, width=0.95\linewidth]

\textit{[stage1: generate]}\\

Edit the input maze image to create exactly one auxiliary annotated image.\\

Maze semantics:

\quad - Black squares are walls and must remain unchanged.

\quad - White squares are walkable cells.

\quad - The blue dot is the start cell and must remain unchanged.

\quad - The green square is the goal cell, is walkable, and must remain unchanged.\\

Editing requirements:

1. Preserve the original maze geometry, cell layout, wall positions, start marker, goal marker, and original colors.

2. Overlay thin red grid-boundary lines that divide the maze image into 10 equal parts along the horizontal axis and 10 equal parts along the vertical axis; draw 9 evenly spaced vertical red lines at 10\%, 20\%, ..., 90\% of the image width and 9 evenly spaced horizontal red lines at 10\%, 20\%, ..., 90\% of the image height, forming a 10×10 partition aligned with the maze cells. As a placement reference, the green square should lie exactly inside one grid cell, with red grid boundaries flush with its four sides.

3. Label every cell (black squares, white squares, blue dot and green square) at its center with its coordinate in black text, using the format (r,c). The top-left cell is (0,0). The first row is labeled left-to-right as (0,0), (0,1), (0,2), (0,3), ... until the last cell in that row. The second row is labeled left-to-right as (1,0), (1,1), (1,2), (1,3), ... . In general, row index r increases by 1 when moving downward, and column index c increases by 1 when moving rightward.

4. Do not add any arrows, path lines, legend, explanation text, title, or extra decorations.\\

\textit{[stage2: answer]}\\

\{question\}

\end{tcolorbox}
\caption{prompt for maze under Self-GtA setting.}
\label{tab:maze-Self-GtA}
\end{table*}

\begin{table*}[h!]
\centering
\begin{tcolorbox}[colback=white, colframe=black!50, sharp corners=south, boxrule=0.4pt, width=0.95\linewidth]

\textit{[stage1: generate]}\\

Edit the input Sokoban image to create exactly one auxiliary annotated image.\\

Sokoban semantics:

- Brick walls remain unchanged.

- Sand floor is walkable.

- The player marker, box marker, and green target marker must remain unchanged.\\

Editing requirements:

1. Preserve original colors, geometry, walls, player, box, and target.

2. Overlay thin red grid-boundary lines aligned with the puzzle cells (one line per row/column boundary).

3. Label every non-wall cell at its center with its coordinate in black text using the format (r,c). Top-left cell is (0,0); r increases downward, c increases rightward.

4. Do not add arrows, path lines, captions, titles, legend, or any other decoration.\\

\textit{[stage2: answer]}\\

\{question\}

\end{tcolorbox}
\caption{prompt for sokoban under Self-GtA setting.}
\label{tab:sokoban-Self-GtA}
\end{table*}

\begin{table*}[h!]
\centering
\begin{tcolorbox}[colback=white, colframe=black!50, sharp corners=south, boxrule=0.4pt, width=0.95\linewidth]

\textit{[stage1: generate]}\\

Edit the input parking-exit puzzle image to create exactly one auxiliary annotated image.\\

Puzzle semantics:

- The board is a \{grid\_size\}x\{grid\_size\} grid.

- Each vehicle has fixed orientation and length.

- Legal moves slide one vehicle by one cell along its orientation.

- The red car "R" must exit to the right boundary on row index \{exit\_row - 1\} (0-based).\\

Editing requirements:

1. Preserve the original board geometry, vehicle positions, labels, and colors exactly.

2. Overlay thin red grid-boundary lines aligned to the board cell boundaries.

3. Label every cell at its center with black text in the format (r,c). The top-left cell is (0,0); r increases downward and c increases rightward.

4. Do not add arrows, path lines, legends, captions, titles, or any extra decorations.

\textit{[stage2: answer]}\\

\{question\}

\end{tcolorbox}
\caption{prompt for parking under Self-GtA setting.}
\label{tab:parking-Self-GtA}
\end{table*}

\begin{table*}[h!]
\centering
\begin{tcolorbox}[colback=white, colframe=black!50, sharp corners=south, boxrule=0.4pt, width=0.95\linewidth]

\textit{[stage1: generate]}\\

Edit the input image to create exactly one auxiliary annotated image.\\

Image semantics:

- Black squares are walls and must remain unchanged.

- Light-gray cells are walkable and must remain unchanged.

- The orange triangle is the start cell and must remain unchanged.

- The blue triangle is the destination cell and must remain unchanged.

- The letters A, B, C, D, E mark walkable label points and must remain unchanged.\\

Editing requirements:

1. Preserve the original geometry, cell layout, wall positions, start marker, destination marker, A/B/C/D/E label markers, and original colors.

2. Overlay thin red grid-boundary lines that divide the image into 10 equal parts along the horizontal axis and 10 equal parts along the vertical axis; draw 9 evenly spaced vertical red lines at 10\%, 20\%, ..., 90\% of the image width and 9 evenly spaced horizontal red lines at 10\%, 20\%, ..., 90\% of the image height, forming a 10x10 partition aligned with the cells.

3. Label every cell (black squares, light-gray cells, orange triangle cell, blue triangle cell, A/B/C/D/E label cells) at its center with its coordinate in black text, using the format (r,c). The top-left cell is (0,0). The first row is labeled left-to-right as (0,0), (0,1), (0,2), ..., until the last cell in that row. In general, row index r increases by 1 when moving downward, and column index c increases by 1 when moving rightward.

4. Do not add any arrows, path lines, legend, explanation text, title, or extra decorations.

\textit{[stage2: answer]}\\

\{question\}

\end{tcolorbox}
\caption{prompt for path under Self-GtA setting.}
\label{tab:path-Self-GtA}
\end{table*}

\begin{table*}[h!]
\centering
\begin{tcolorbox}[colback=white, colframe=black!50, sharp corners=south, boxrule=0.4pt, width=0.95\linewidth]

\textit{[stage1: generate]}\\

Edit the input Onet puzzle image to create exactly one auxiliary annotated image.\\

Onet puzzle semantics:

- The board is a 6x6 grid of cells.

- Each cell contains a fruit icon or is empty.

- Identical fruit icons can be removed as a pair if they can be connected by a path with at most 2 turns using only horizontal/vertical segments that do not pass through other fruits.\\

Editing requirements:

1. Preserve the original board geometry, cell layout, fruit icons, and original colors exactly.

2. Overlay thin red grid-boundary lines that divide the image into a 6x6 grid: draw 5 evenly spaced vertical red lines and 5 evenly spaced horizontal red lines, aligned with cell boundaries.

3. Label every cell at its center with its coordinate in black text, using the format (r,c). The top-left cell is (0,0). Row index r increases downward; column index c increases rightward.

4. Do not add any arrows, path lines, legend, explanation text, title, or extra decorations.\\

\textit{[stage2: answer]}\\

\{question\}

\end{tcolorbox}
\caption{prompt for onet under Self-GtA setting.}
\label{tab:onet-Self-GtA}
\end{table*}

\begin{table*}[h!]
\centering
\begin{tcolorbox}[colback=white, colframe=black!50, sharp corners=south, boxrule=0.4pt, width=0.95\linewidth]

\textit{[stage1: generate]}\\

Edit the input Klotski image to create exactly one auxiliary annotated image.\\

Klotski semantics:

- Grid is 6x6.

- Blocks keep shape and orientation; every move slides one block by one cell.

- Target is to move the 2x2 block T so its top-left reaches (4,2).\\

Editing requirements:

1. Preserve original image dimensions, board geometry, block positions, labels, and
   colors exactly. The input image may omit visible cell lines; infer the implied 6x6
   tiling from the layout and keep each block's cell extent unchanged (e.g. 1x2 stays
   1x2; do not shrink, split, or merge blocks).
   
2. Overlay thin red dashed grid-boundary lines aligned to the 6x6 cell boundaries.

3. Label every cell center with coordinate "(r,c)" in black text.

4. Do not add arrows, legends, captions, titles, or extra decorations.\\

\textit{[stage2: answer]}\\

\{question\}

\end{tcolorbox}
\caption{prompt for klotski under Self-GtA setting.}
\label{tab:klotski-Self-GtA}
\end{table*}

\begin{table*}[h!]
\centering
\begin{tcolorbox}[colback=white, colframe=black!50, sharp corners=south, boxrule=0.4pt, width=0.95\linewidth]
You are a precise maze solver.\\

SEMANTICS

- Black squares: walls (impassable)

- White squares: path (walkable)

- Blue dot: start (the agent)

- The green square marks the goal cell. It is passable, not a wall.

- Legal moves: up, down, left, right only. One cell per step.\\

OUTPUT FORMAT (STRICT)\\

1) MULTI-IMAGE MODE : generate a SEQUENCE OF SEPARATE IMAGES, one per move:

\quad - Each output image must depict the maze state AFTER applying exactly one legal move.
   
\quad - Do NOT include the initial (pre-move) state.

\quad - Keep palette/layout/scale identical to the input; only the blue dot moves.

\quad - The number of returned images MUST equal the number of moves in the final answer (see step 2).
   - Absolutely FORBIDDEN: any collage/montage/spritesheet/grid/multi-panel/side-by-side/stacked images; no arrows, captions, or overlays; no GIFs/animations/video.\\

2) After all step images, emit EXACTLY ONE LINE containing ONLY the final move list as a JSON array of lowercase strings, wrapped as:
   \texttt{\detokenize{<ANSWER_JSON>["right","down","left"]</ANSWER_JSON>}}\\

NO EXTRAS

- No tools, no OCR, no explanations, and no text other than the single \texttt{\detokenize{<ANSWER_JSON>}} line.

- Do not restate the instructions or the condition.\\

REMINDERS

- Decide the full path first, then emit the image sequence (one image per move), then the single \texttt{\detokenize{<ANSWER_JSON>}} line.

- One move per image; images must be separate files/parts, not stitched together in any way.

\end{tcolorbox}
\caption{prompt for maze under Self-VCoT setting.}
\label{tab:maze-self-VCoT}
\end{table*}

\begin{table*}[h!]
\centering
\begin{tcolorbox}[colback=white, colframe=black!50, sharp corners=south, boxrule=0.4pt, width=0.95\linewidth]
You are solving a single-box Sokoban puzzle.\\

Legend:

- Brick wall: blocked

- Sand floor: walkable

- Small person: player start

- Wooden crate with X: box

- Green X: target\\

Rules:

1) Allowed moves: up, down, left, right (one cell per step).

2) If the player moves into the box, the box is pushed by one cell in the same direction.

3) A push is valid only if the box destination is not a wall.

4) No diagonal movement.\\

Goal:

Push the box onto the target cell.\\

OUTPUT FORMAT (STRICT)\\

1) MULTI-IMAGE MODE: generate a SEQUENCE OF SEPARATE IMAGES, one per move:

\quad - Each output image must depict the maze state AFTER applying exactly one legal move.

\quad - Do NOT include the initial (pre-move) state.

\quad - Keep palette/layout/scale identical to the input; only the blue dot moves.

\quad - The number of returned images MUST equal the number of moves in the final answer (see step 2).

\quad - Absolutely FORBIDDEN: any collage/montage/spritesheet/grid/multi-panel/side-by-side/stacked images; no arrows, captions, or overlays; no GIFs/animations/video.\\

2) After all step images, emit EXACTLY ONE LINE containing ONLY the final move list as a JSON array of lowercase strings, wrapped as:
   \texttt{\detokenize{<ANSWER_JSON>["right","down","left"]</ANSWER_JSON>}}\\

NO EXTRAS

- No tools, no OCR, no explanations, and no text other than the single \texttt{\detokenize{<ANSWER_JSON></ANSWER_JSON>}} line.

- Do not restate the instructions or the condition.\\

REMINDERS

- Decide the full path first, then emit the image sequence (one image per move), then the single \texttt{\detokenize{<ANSWER_JSON>}} line.

- One move per image; images must be separate files/parts, not stitched together in any way.

\end{tcolorbox}
\caption{prompt for sokoban under Self-VCoT setting.}
\label{tab:sokoban-self-VCoT}
\end{table*}

\begin{table*}[h!]
\centering
\begin{tcolorbox}[colback=white, colframe=black!50, sharp corners=south, boxrule=0.4pt, width=0.95\linewidth]
In the image there are five labeled points: A, B, C, D, E.
From the orange triangle (start) to the blue triangle (destination), determine which labeled points lie on valid simple path without entering black grids.\\

SEMANTICS

- Black squares: walls (impassable)

- Light-gray cells are walkable.

- Orange triangle: start/current position.

- Blue triangle: destination.

- A/B/C/D/E: labeled walkable points.

- Legal moves: up, down, left, right only. One cell per step.

- Follow the valid simple path from start to destination: no backtracking and no revisits.\\

OUTPUT FORMAT (STRICT)

You need to start from the start point, plan the path to the destination step by step, and output the passed labeled points.

1) MULTI-IMAGE MODE: generate a SEQUENCE OF SEPARATE IMAGES, one per move:

\quad - Each output image must depict the state AFTER applying exactly one legal move.

\quad - Do NOT include the initial (pre-move) state.

\quad - Keep palette/layout/scale identical to the input; only the current position marker moves.

\quad - The number of returned images MUST equal the number of moves in the final path.

\quad - Absolutely FORBIDDEN: any collage/montage/spritesheet/grid/multi-panel/side-by-side/stacked images; no arrows, captions, overlays, GIFs, animations, or video.\\

2) STEP DECISION FORMAT

for EACH step decision, output exactly one JSON object:
   \texttt{\detokenize{<STEP_JSON>{"move":"up|down|left|right","hit_labels":["A"]}</STEP_JSON>}}
   
\quad - hit\_labels must be labels newly passed on THIS move only.
\quad - Use uppercase letters among A/B/C/D/E only.

\quad - If no new label is passed this step, output an empty list [].\\

3) FINAL LINE — after all step images/decisions, emit exactly one line:
   \texttt{\detokenize{<answer>point1,point2,...</answer>}}\\

NO EXTRAS

- No extra explanation text.

\end{tcolorbox}
\caption{prompt for path under Self-VCoT setting.}
\label{tab:path-self-VCoT}
\end{table*}

\begin{table*}[h!]
\centering
\begin{tcolorbox}[colback=white, colframe=black!50, sharp corners=south, boxrule=0.4pt, width=0.95\linewidth]
You are solving an Onet puzzle on a 6x6 board.

Coordinate system

- Use 0-based coordinates.

- (r, c) means row r, column c.

- (0,0) is the top-left cell.

- r increases downward, c increases to the right.\\

Goal:

- Decide whether the whole board can be fully cleared.

- In each step, remove exactly one pair of identical fruits if they can be linked.

- A valid link uses only horizontal/vertical segments and can have at most 2 turns.

- The path must not pass through any other fruit.\\

OUTPUT FORMAT (STRICT)\\

1) MULTI-IMAGE MODE: generate a SEQUENCE OF SEPARATE IMAGES, one per move:

\quad - Each output image must depict the board state AFTER applying exactly one pair removal.
   
\quad - Do NOT include the initial (pre-move) state.

\quad - Keep palette/layout/scale identical to the input; empty cells must be clearly indicated (e.g., cleared/blank).

\quad- The number of returned images MUST equal the number of moves in the final answer (see step 2).

\quad - Absolutely FORBIDDEN: any collage/montage/spritesheet/grid/multi-panel/side-by-side/stacked images; no arrows, captions, or overlays; no GIFs/animations/video.\\

2) After all step images, emit EXACTLY ONE LINE containing ONLY the final move list as a JSON array of coordinate pairs, wrapped as:
\texttt{\detokenize{<answer_json>[[[r1,c1],[r2,c2]],[[r3,c3],[r4,c4]],...]</answer_json>}}\\

NO EXTRAS

- No tools, no OCR, no explanations, and no text other than the single \texttt{\detokenize{<ANSWER_JSON>}} line.

- Do not restate the instructions or the condition.\\

REMINDERS

- Determine the full sequence of moves first, then emit the image sequence (one image per move), then the single \texttt{\detokenize{<ANSWER_JSON>}} line.

- One move per image; images must be separate files/parts, not stitched together in any way.

\end{tcolorbox}
\caption{prompt for onet under Self-VCoT setting.}
\label{tab:onet-self-VCoT}
\end{table*}

\begin{table*}[h!]
\centering
\begin{tcolorbox}[colback=white, colframe=black!50, sharp corners=south, boxrule=0.4pt, width=0.95\linewidth]
You are solving a parking-exit puzzle on \{grid\_size\}x\{grid\_size\} board.\\

Rules:

1) Vehicle orientations are fixed (horizontal or vertical).

2) In one move, exactly one vehicle slides by one cell along its orientation.

3) No overlap and no leaving the board.

4) Success is achieved when the rightmost cell of the red car "R" reaches the right boundary on row index {exit\_row - 1} (indices are 0-based).\\

OUTPUT FORMAT (STRICT)\\

1) MULTI-IMAGE MODE: generate a SEQUENCE OF SEPARATE IMAGES, one per move:

\quad - Each output image must depict the board state AFTER applying exactly one legal move.

\quad - Do NOT include the initial (pre-move) state.

\quad - Keep palette/layout/scale identical to the input; only the moved vehicle changes position.

\quad - The number of returned images MUST equal the number of moves in the final answer (see step 2).

\quad - Absolutely FORBIDDEN: any collage/montage/spritesheet/grid/multi-panel/side-by-side/stacked images; no arrows, captions, or overlays; no GIFs/animations/video. \\

2) After all step images, emit EXACTLY ONE LINE containing ONLY the final move list as a JSON array of move objects, wrapped as:
   \texttt{\detokenize{<ANSWER_JSON>[{"car":"A","direction":"up"}, {"car":"R","direction":"right"}]</ANSWER_JSON>}}\\

NO EXTRAS

- No tools, no OCR, no explanations, and no text other than the single \texttt{\detokenize{<ANSWER_JSON>}} line.

- Do not restate the instructions or the condition.\\

REMINDERS

- Decide the full move sequence first, then emit the image sequence (one image per move), then the single \texttt{\detokenize{<ANSWER_JSON>}} line.

- One move per image; images must be separate files/parts, not stitched together in any way.

\end{tcolorbox}
\caption{prompt for parking under Self-VCoT setting.}
\label{tab:parking-self-VCoT}
\end{table*}

\begin{table*}[h!]
\centering
\small
\begin{tcolorbox}[colback=white, colframe=black!50, sharp corners=south, boxrule=0.4pt, width=\linewidth]
You are an expert Klotski puzzle solver.\\

Coordinate system

- Use 0-based coordinates.

- (r, c) means row r, column c.

- (0,0) is the top-left cell.

- r increases downward, c increases to the right.\\

Board

- Grid size: width=6, height=6.

- There are exactly two empty cells at any time.

- Every non-empty cell belongs to exactly one block from the listed block shapes.\\

Block shapes

- One 2x2 target block.

- Multiple 1x2 vertical blocks.

- Multiple 2x1 horizontal blocks.

- Multiple 1x1 single blocks.\\

Block reference in actions

- In the action output, a block is identified by its CURRENT TOP-LEFT coordinate at the time of that move.\\

Legal move definition

1. Each step slides ONE block by exactly ONE cell in {UP, DOWN, LEFT, RIGHT}.

2. No rotation, no diagonal move, no jump.

3. A move is legal only if EVERY destination cell of that block is currently empty.

\quad - Let current\_cells be the cells currently occupied by the moving block.
   
\quad - Let destination\_cells be the cells occupied after translating the block by one cell in the chosen direction.
   
\quad - A move is legal iff every cell in (destination\_cells - current\_cells) is empty before the move.
   
\quad  - For 1x2 / 2x1 blocks, required empty destination cells can be 1 or 2 depending on move direction.
   
\quad  - For 2x2 blocks, required empty destination cells are 2.
   
4. Cells occupied by OTHER blocks are never passable.\\

Goal

- Make the 2x2 target block's TOP-LEFT coordinate reach (4,2).\\

OUTPUT FORMAT (STRICT)\\

1) MULTI-IMAGE MODE: generate a SEQUENCE OF SEPARATE IMAGES, one per move:

\quad - Each output image must depict the board state AFTER applying exactly one legal move.

\quad - Do NOT include the initial (pre-move) state.

\quad - Keep board layout/scale/colors consistent with the input board.

\quad - The number of returned images MUST equal the number of moves in your final answer.

\quad - Absolutely FORBIDDEN: collage/montage/spritesheet/grid/multi-panel/side-by-side/stacked images; no GIF/animation/video.\\

2) For each planning turn, output exactly one move as:

\texttt{\detokenize{<STEP_JSON>{"from":[r,c],"direction":"UP|DOWN|LEFT|RIGHT"}</STEP_JSON>}}\\

3) After reaching the goal, output exactly one final JSON object:

\{

 \quad "solution": [
  
 \quad\quad   {"from": [<r1>, <c1>], "direction": "<DIRECTION>"},
    
 \quad\quad   {"from": [<r2>, <c2>], "direction": "<DIRECTION>"}
    
 \quad ]
  
\}

\end{tcolorbox}
\caption{prompt for klotski under Self-VCoT setting.}
\label{tab:klotski-self-VCoT}
\end{table*}

\begin{table*}[h!]
\centering
\small
\begin{tcolorbox}[colback=white, colframe=black!50, sharp corners=south, boxrule=0.4pt, width=\linewidth]
Edit the input maze image to create exactly one auxiliary annotated image that highlights a feasible solution path.\\

Maze semantics:

- Black squares are walls and must remain unchanged.

- White squares are walkable cells.

- The blue dot is the start cell and must remain unchanged.

- The green square is the goal cell, is walkable, and must remain unchanged.\\

Editing requirements:\\

1. Preserve the original maze geometry, cell layout, wall positions, start marker, goal marker, and original colors.\\

2. Overlay a single continuous RED polyline that traces a legal path from the blue start cell to the green goal cell:

\quad- The polyline must follow walkable (white) cells only and must NEVER cross or touch any black wall cell.
   
\quad- Movement is restricted to up / down / left / right between cell centers; no diagonals, no jumps.

\quad - The polyline starts exactly at the center of the blue start cell and ends exactly at the center of the green goal cell.
   
\quad - Use a clean, solid bright red stroke of moderate thickness (roughly 10-15 percent of a cell width) so the path is clearly visible against the maze background.\\
   
3. Do not draw arrows, dots along the way, grid lines, coordinate labels, legend, captions, titles, or any other decoration. Only the single red polyline should be added.\\

4. Do not modify the colors or positions of the start marker, goal marker, walls, or walkable cells in any other way.

\end{tcolorbox}
\caption{prompt for maze under Self-VR setting.}
\label{tab:maze-self-VR}
\end{table*}

\begin{table*}[h!]
\centering
\small
\begin{tcolorbox}[colback=white, colframe=black!50, sharp corners=south, boxrule=0.4pt, width=\linewidth]
Edit the input Sokoban image to create exactly one auxiliary annotated image that highlights a feasible player path that pushes the box onto the target.\\

Sokoban semantics:

- Brick walls remain unchanged.

- Sand floor is walkable.

- The player, box, and green target must remain unchanged.

- Pushes happen when the player steps into the box; the box moves one cell in the same direction; pushes into walls are illegal.\\

Editing requirements:\\

1. Preserve original colors, geometry, walls, player, box, and target.\\

2. Overlay a single continuous RED polyline that traces the player's trajectory from the player start cell, ending exactly at the cell from which the box gets pushed onto the green target. The polyline:

\quad - Follows non-wall cells only and never crosses a wall.

\quad - Moves only orthogonally (up / down / left / right) between cell centers.

\quad- Uses a clean solid red stroke (~10-15 percent of a cell width) so it is clearly visible.\\

3. Do NOT draw the box's trajectory, arrows, dots, grid lines, coordinate labels, captions, titles, or any other decoration. Only the player's path polyline is added.

\end{tcolorbox}
\caption{prompt for sokoban under Self-VR setting.}
\label{tab:sokoban-self-VR}
\end{table*}

\begin{table*}[h!]
\centering
\small
\begin{tcolorbox}[colback=white, colframe=black!50, sharp corners=south, boxrule=0.4pt, width=\linewidth]
Edit the input image to create exactly one auxiliary annotated image that highlights a feasible solution path.\\

Image semantics:

- Black squares are walls and must remain unchanged.

- Light-gray cells are walkable.

- The orange triangle is the start cell and must remain unchanged.

- The blue triangle is the destination cell and must remain unchanged.

- The letters A, B, C, D, E mark walkable label points and must remain unchanged.\\

Editing requirements:\\

1. Preserve the original geometry, cell layout, wall positions, start marker, destination marker, A/B/C/D/E label markers, and original colors.\\

2. Overlay a single continuous RED polyline that traces a legal simple path from the orange-triangle start cell to the blue-triangle destination cell:
   - The polyline must follow walkable (non-wall) cells only and must NEVER cross or touch any black wall cell.
   
\quad - Movement is restricted to up / down / left / right between cell centers; no diagonals, no jumps.
   
\quad - The polyline starts exactly at the center of the orange-triangle start cell and ends exactly at the center of the blue-triangle destination cell.
   
\quad - The path must be a simple path: no revisits and no backtracking.
   
\quad - Use a clean, solid bright red stroke of moderate thickness (roughly 10-15 percent of a cell width) so the path is clearly visible against the background.\\
   
3. Do not draw arrows, dots along the way, grid lines, coordinate labels, legend, captions, titles, or any other decoration. Only the single red polyline should be added.\\

4. Do not modify the colors or positions of the start marker, destination marker, A/B/C/D/E label markers, walls, or walkable cells in any other way.

\end{tcolorbox}
\caption{prompt for path under Self-VR setting.}
\label{tab:path-self-VR}
\end{table*}

\begin{table*}[h!]
\centering
\small
\begin{tcolorbox}[colback=white, colframe=black!50, sharp corners=south, boxrule=0.4pt, width=\linewidth]
Edit the input parking-exit puzzle image to create exactly one auxiliary annotated image that highlights a feasible move strategy for the red car.\\

Puzzle semantics:\\

- The board is a \{grid\_size\}x\{grid\_size\} grid.

- Each vehicle has fixed orientation and length.

- Legal moves slide one vehicle by one cell along its orientation.

- The red car "R" must exit to the right boundary on row index {exit\_row - 1} (0-based).\\

Editing requirements:

1. Preserve the original board geometry, vehicle positions, labels, and colors exactly.

2. Add visual hints that indicate a feasible strategy (for example, numbered move markers or directional cues near vehicles) while keeping all hints on valid cells only.

3. Ensure the hints correspond to legal single-step moves and do not imply illegal orientation changes or overlaps.

4. Do not alter car locations in the image; only overlay hints. Do not add unrelated decorations.
\end{tcolorbox}
\caption{prompt for parking under Self-VR setting.}
\label{tab:parking-self-VR}
\end{table*}

\begin{table*}[h!]
\centering
\small
\begin{tcolorbox}[colback=white, colframe=black!50, sharp corners=south, boxrule=0.4pt, width=\linewidth]
Edit the input Onet puzzle image to create exactly one auxiliary annotated image that highlights all matching pairs with their connection paths.\\

Onet puzzle semantics:\\

- The board is a 6x6 grid of cells.

- Each cell contains a fruit icon or is empty.

- Identical fruit icons can be removed as a pair if they can be connected by a path with at most 2 turns using only horizontal/vertical segments that do not pass through other fruits.\\

Editing requirements:

1. Preserve the original board geometry, cell layout, fruit icons, and original colors exactly.

2. For each pair of identical fruits that can be validly connected (forming a complete clearing solution), draw a solid colored polyline from the center of one fruit cell to the center of the other, following the valid connection path (horizontal/vertical segments, at most 2 turns, not passing through other fruits). Use a distinct bright color per pair so all pairs are visually distinguishable.\\

3. Use a clean stroke of moderate thickness (roughly 10-15 percent of a cell width) so paths are clearly visible against the background.

4. Do not add grid lines, coordinate labels, arrows, legend, captions, titles, or any other decoration. Only the colored polylines should be added.
5. Do not modify the fruit icons, empty cells, or board layout in any other way.
\end{tcolorbox}
\caption{prompt for onet under Self-VR setting.}
\label{tab:onet-self-VR}
\end{table*}

\begin{table*}[h!]
\centering
\small
\begin{tcolorbox}[colback=white, colframe=black!50, sharp corners=south, boxrule=0.4pt, width=\linewidth]
Edit the input Klotski image to create exactly one auxiliary annotated image that highlights one feasible solving strategy.\\

Klotski semantics:

- Grid is 6x6.

- Blocks keep shape and orientation; every move slides one block by one cell.

- Target is to move the 2x2 block T so its top-left reaches (4,2).\\

Editing requirements:

1. Preserve original image dimensions, board geometry, block positions, labels, and
   colors exactly. The input image may omit visible cell lines; infer the implied 6x6
   tiling from the layout and keep each block's cell extent unchanged (e.g. 1x2 stays
   1x2; do not shrink, split, or merge blocks).
   
2. Add concise visual hints indicating a feasible move strategy (for example, ordered markers and local directional cues near moved blocks).

3. Hints must correspond to legal one-cell slides and must not imply illegal overlaps or rotations.

4. Do not actually move blocks in the edited image; only overlay hints.

5. Do not add unrelated decorations or long text.
\end{tcolorbox}
\caption{prompt for klotski under Self-VR setting.}
\label{tab:klotski-self-VR}
\end{table*}

\end{document}